\pdfoutput=1
\documentclass[11pt]{article}

\usepackage[preprint]{acl}

\usepackage{times}
\usepackage{latexsym}

\usepackage[T1]{fontenc}
\usepackage[utf8]{inputenc}

\usepackage{microtype}

\usepackage{inconsolata}

\usepackage{algorithm}
\usepackage{tabularx} 
\usepackage{graphicx}
\usepackage{soul}
\usepackage[normalem]{ulem}

\usepackage{xcolor}
\usepackage{colortbl}
\usepackage[most]{tcolorbox}
\definecolor{highlight}{HTML}{81CE6D}
\definecolor{mycolor}{HTML}{2650CC}
\definecolor{second}{HTML}{666666}
\definecolor{rightcolor}{HTML}{28a745}
\definecolor{wrongcolor}{HTML}{dc3545}

\definecolor{Gray}{gray}{0.95}
\definecolor{Cyan}{rgb}{0.88,1,1}
\definecolor{LightCyan}{rgb}{0.92,1,1}
\definecolor{DarkCyan}{rgb}{0.82,1,1}
\definecolor{tabhighlight}{HTML}{e5e5e5}

\newcommand{\arxiv}[1]{{\color{black} #1}}

\usepackage{booktabs}
\usepackage{multirow}
\usepackage{multicol}
\usepackage{stfloats} % Improved behavior for double-column floats
\usepackage{amssymb} % for checkmark
\title{Confidence Matters: Revisiting Intrinsic Self-Correction Capabilities of Large Language Models} % A Causal View

\author{First Author \\
  Affiliation / Address line 1 \\
  Affiliation / Address line 2 \\
  Affiliation / Address line 3 \\
  \texttt{email@domain} \\\And
  Second Author \\
  Affiliation / Address line 1 \\
  Affiliation / Address line 2 \\
  Affiliation / Address line 3 \\
  \texttt{email@domain} \\}

\author{Loka Li$^{1}$\thanks{~~Equal Contribution.} \quad Zhenhao Chen$^{1*}$ \quad Guangyi Chen$^{1,2*}$ \quad Yixuan Zhang$^1$  \quad Yusheng Su$^{1}$ \\ \bf Eric Xing$^{1,2}$ \quad Kun Zhang$^{1,2}$ \\
	$^1$ Mohamed bin Zayed University of Artificial Intelligence, Abu Dhabi, UAE \\ $^2$ Carnegie Mellon University, Pittsburgh PA, USA }

\usepackage{afterpage} 
\usepackage{titling}

\begin{document}

\makeatletter
\let\@oldmaketitle\@maketitle
\renewcommand{\@maketitle}{\@oldmaketitle
\begin{minipage}{\textwidth}
\vspace{-0.8cm}
\centering
\includegraphics[width=\textwidth]{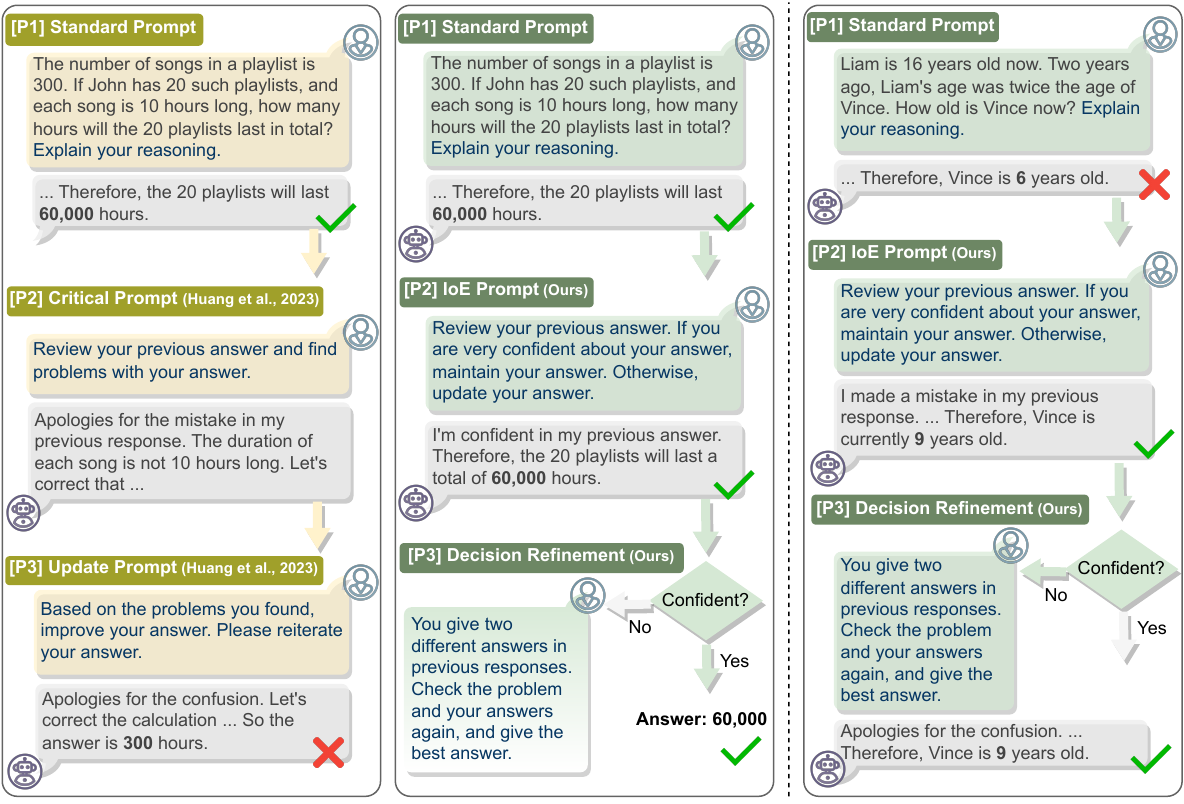} 
\captionof{figure}{\textbf{Comparisons between our IoE-based Prompt and Critical Prompt}. \textbf{Left:} Critical Prompt \cite{huang2023large} as baseline. \textbf{Middle/Right:} the proposed IoE-based Prompt. Regarding our prompts, when the answers of [P1] standard question and [P2] IoE prompt match, the final answer will be directly output, as shown in the \textbf{middle}. Otherwise, the decision prompt for final decision-making will execute, as the example shown in the \textbf{right}. All examples are generated from GSM8K \cite{cobbe2021training} and  evaluated by \texttt{gpt-3.5-turbo-1106} model.}
\label{fig:shortexample}
    \vspace{+6mm}
\end{minipage}}
\makeatother

\maketitle

% \begin{center}
% \includegraphics[width=\textwidth]{figures/gsm8k.drawio.pdf}
% \captionof{figure}{How do humans localize unintentional events such as "A boy fell down while playing skateboard"? We argue that unintentional event localization requires 1) understanding what the boy is doing; 2) learning how to intentionally do this thing; 3) rethinking the cause by comparisons. Green denotes intentional events while orange denotes unintentional one.}
% \label{fig:unintentional_event} % For referencing within the text
% \end{center}

\begin{abstract}

The recent success of Large Language Models (LLMs) has catalyzed an increasing interest in their self-correction capabilities. This paper presents a comprehensive investigation into the intrinsic self-correction of LLMs, attempting to address the ongoing debate about its feasibility. 
Our research has identified an important latent factor - the ``confidence'' of LLMs - during the self-correction process. Overlooking this factor may cause the models to over-criticize themselves, resulting in unreliable conclusions regarding the efficacy of self-correction. 
We have experimentally observed that LLMs possess the capability to understand the ``confidence'' in their own responses. It motivates us to develop an ``If-or-Else'' (IoE) prompting framework, designed to guide LLMs in assessing their own ``confidence'', facilitating intrinsic self-corrections. We conduct extensive experiments and demonstrate that our IoE-based Prompt can achieve a consistent improvement regarding the accuracy of self-corrected responses over the initial answers. Our study not only sheds light on the underlying factors affecting self-correction in LLMs, but also introduces a practical framework that utilizes the IoE prompting principle to efficiently improve self-correction capabilities with ``confidence''. The code is available at \url{https://github.com/MBZUAI-CLeaR/IoE-Prompting.git}.

\end{abstract}

\section{Introduction}
\label{sec1:intro}

\begin{table*}[t]
	\small
	\centering
	\begin{tabular}{lcccc}    
		\toprule
		% Work & Few-shot & Iterative & Agent & Confidence & Application \\
            % & Non-iterative & Task-agnostic & Zero-shot & Confidence-based \\
            & \textbf{One-pass Correction} &  \textbf{Zero-shot}  & \textbf{Task-agnostic} & \textbf{Confidence-based}  \\
		\midrule
  	Self-Refine \citep{madaan2023self} & \textcolor{wrongcolor}{\texttimes} & \textcolor{wrongcolor}{\texttimes} & \textcolor{wrongcolor}{\texttimes} & \textcolor{wrongcolor}{\texttimes}  \\
		IterRefinement \citep{chen2023teaching}& \textcolor{wrongcolor}{\texttimes} & \textcolor{wrongcolor}{\texttimes} & \textcolor{wrongcolor}{\texttimes} & \textcolor{wrongcolor}{\texttimes}  \\
        SelFee \cite{selfee2023} & \textcolor{wrongcolor}{\texttimes} & \textcolor{rightcolor}{\checkmark} & \textcolor{wrongcolor}{\texttimes} & \textcolor{wrongcolor}{\texttimes} \\
        Self-Verification \citep{gero2023self} & \textcolor{rightcolor}{\checkmark} & \textcolor{wrongcolor}{\texttimes} & \textcolor{wrongcolor}{\texttimes} & \textcolor{wrongcolor}{\texttimes} \\
        Self-Defense \cite{helbling2023llm} & \textcolor{rightcolor}{\checkmark} & \textcolor{rightcolor}{\checkmark} & \textcolor{wrongcolor}{\texttimes} & \textcolor{wrongcolor}{\texttimes}  \\
		SelfCheckGPT \citep{manakul2023selfcheckgpt} & \textcolor{rightcolor}{\checkmark} & \textcolor{rightcolor}{\checkmark} & \textcolor{wrongcolor}{\texttimes} & \textcolor{wrongcolor}{\texttimes} \\
	Critical Prompt \cite{huang2023large} & \textcolor{rightcolor}{\checkmark} & \textcolor{rightcolor}{\checkmark} & \textcolor{rightcolor}{\checkmark} & \textcolor{wrongcolor}{\texttimes}  \\
	IoE-based Prompt (Ours) & \textcolor{rightcolor}{\checkmark} & \textcolor{rightcolor}{\checkmark} & \textcolor{rightcolor}{\checkmark} & \textcolor{rightcolor}{\checkmark} \\
		\bottomrule
	\end{tabular}
 \setlength{\abovecaptionskip}{0.1cm}
	\caption{\textbf{The summery of related work on {intrinsic self-correction}.} This table categorizes intrinsic self-correction based on four properties including One-pass Correction (self-correction achieved in a single attempt, regardless of the number of prompts, and no multiple attempts required), Zero-shot (independent of few-shot examples), Task-agnostic (applicable across various tasks), and Confidence-based (reliant on LLM confidence levels). Please refer to Appendix \ref{sec:related-work} for more discussions.
}
	\label{tab:related_works}
 \vspace{-0.4cm}
\end{table*}

 % Related works on \textit{intrinsic self-correction}. The properties of different prompts: One-pass (indicating if the refinement process does not require multiple iterations), Zero-shot (indicating if few-shot in-context examples are not required), Task-agnostic (indicating if the prompt is versatile and can be used in many tasks), and Confidence-based (indicating if the prompt is based on the confidence of LLMs). Refer to Section \ref{sec:related-work} for more discussions.
% Overview of Prompt Properties in Intrinsic Self-Correction Studies: This table categorizes prompts based on key attributes—One-pass (no multiple iterations required), Zero-shot (independent of few-shot examples), Task-agnostic (applicable across various tasks), and Confidence-based (reliant on LLM confidence levels). For detailed discussions, see Section \ref{sec:related-work}.

% \textcolor{rightcolor}{\checkmark}

% \textcolor{wrongcolor}{\texttimes}

% With the rapid development of artificial intelligence \citep{lecun2015deep} nowadays, 
Large language models (LLMs) trained with massive data and expansive parameters showcase promising performance on human instruction understanding \citep{jin2023instance, weld2022survey, wu2022natural} and text generation \citep{fernandes2023bridging, qian2022controllable}.
However, the exploration of their inherent reasoning abilities and iterative refinement capabilities is ongoing.

% One of most promising exploration is the ability of "self-correction".  

% reasoning ability and 
% have attracted a great amount of attention recently, with their surprising performance on language understanding \citep{jin2023instance, weld2022survey, wu2022natural}, generation \citep{fernandes2023bridging, qian2022controllable}, and reasoning \citep{webb2023emergent, wei2022chain}. 
% due to their impressive performance across a wide range of natural language processing benchmarks \citep{chen2023extensive, niklaus2023lextreme, shankar2023privacyglue}. Indeed, LLMs have exhibited surprising abilities of language understanding \citep{jin2023instance, weld2022survey, wu2022natural}, generation \citep{fernandes2023bridging, qian2022controllable}, and reasoning \citep{webb2023emergent, wei2022chain}. 
% However, regarding the mastery of certain capabilities by LLMs, there remains significant controversy. For instance, the question of whether LLMs can self-correct reasoning has sparked intense debate \citep{pan2023automatically}.

Among the most promising areas of exploration is the capability for ``self-correction''~\citep{pan2023automatically,shinn2023reflexion,yao2022react,madaan2023self,bai2022constitutional,ganguli2023capacity,chen2023teaching,kim2023language}, which refers to whether LLMs can refine their responses based on their previous outputs, leveraging either external or internal feedback. Numerous studies have demonstrated effective self-correction performance through the integration of external feedback generated by the learned critic models~\citep{paul2023refiner, akyurek2023rl4f, welleck2022generating,peng2023check,gao2023rarr} or interacting with human or tools~\citep{gou2023critic, chern2023factool, olausson2023self}. Despite effectiveness, training extra model or interacting with real world may result in extra costs. Thus, recent studies \citep{madaan2023self, chen2023iterative, manakul2023selfcheckgpt, huang2023large} are beginning to explore \textit{intrinsic self-correction}, noted for its cost efficiency. Table \ref{tab:related_works} summarizes and compares these related studies. However, the practicality of \textit{intrinsic self-correction} remains a topic of debate. For example, \citet{huang2023large} argues that LLMs are not yet capable of self-correcting reasoning, observing a decrease in model performance when prompted to find problems and revise answers by LLMs themselves. More discussions on the related work are in Appendix \ref{sec:related-work}.

In this paper, we revisit the scenario of \textit{intrinsic self-correction} of LLMs, identifying the critical role of LLMs'  ``confidence'' in this process. The ``confidence'' refers to the degree of LLMs assigns to its own responses, indicating how likely the model believes its output is correct. We use ``confidence'' in quotation marks to denote the overall certainty of the model's global response, rather than individual generated words (unless otherwise noted, confidence in the following parts of the paper pertains to this global context). 
We propose a hypothesis on confidence: LLMs have the inherent ability to understand and assess their confidence, a capability that can be used to enhance self-correction. 
To validate this hypothesis, we conduct a series of experiments, yielding the following insightful observations:
1) In deterministic tasks, LLM-assessed confidence levels align with those derived from multiple inferences for consistency checking, yet require only a single inference. 
2) In open tasks, LLMs show improved confidence assessment compared to traditional consistency checking methods.
3) The impact of self-correction prompts varies across different confidence levels. 
4) Guiding self-correction using assessed confidence levels notably enhances performance.

It motivates us to introduce the If-or-Else (IoE) prompting principle to guide LLMs in assessing their own confidence. 
Ignoring the confidence, directly using critical prompts like ``find your problem'' in ~\cite{huang2023large} (we named this baseline method as Critical Prompt) may lead LLMs to doubt their initially correct responses, resulting in the alteration of these responses to incorrect ones.  
This phenomenon is not unexpected, as strong criticism can undermine the model's confidence in its answers, similar to a child facing challenges from a strict teacher. 
Thus, as illustrated in Figure~\ref{fig:shortexample}, we pose an If-or-Else question to the LLM prior to initiating self-review. If the LLM believes itself confident, the response remains unchanged. Otherwise, if the LLM lacks confidence, it will revise the answer, addressing any identified issues. 
In our experiments, we evaluated the self-correction ability of our IoE-based Prompt on four LLMs, across six benchmark tests. Our findings demonstrate that incorporating confidence consistently and significantly enhances the self-correction capabilities, as compared to \cite{huang2023large}. Additionally, we conducted a detailed investigation into mechanisms behind the effective performance of IoE-based Prompt in facilitating self-correction.

\begin{figure}[t!] %H为当前位置，!htb为忽略美学标准，htbp为浮动图形
    \centering %图片居中
    \includegraphics[width=0.5\textwidth]{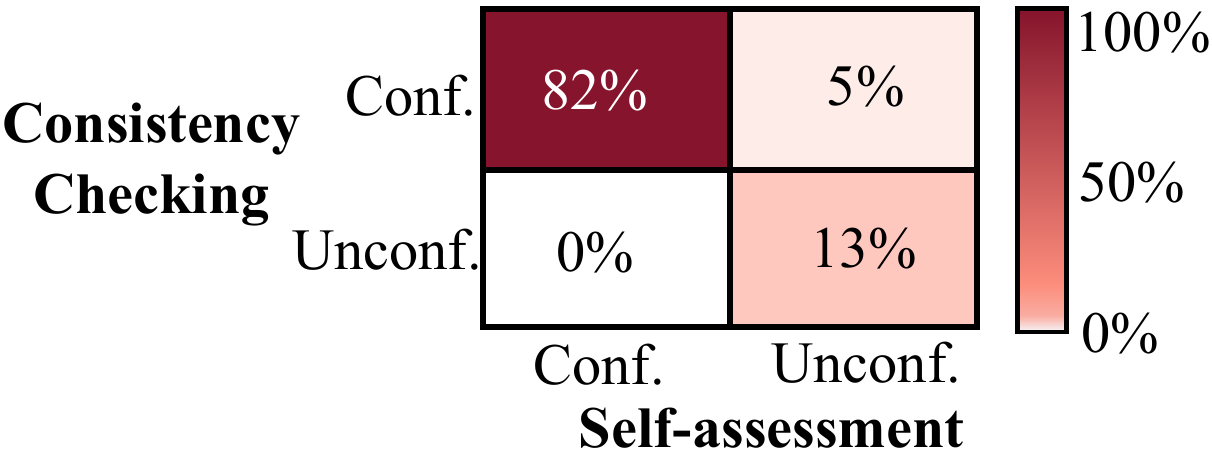}
    \setlength{\abovecaptionskip}{-0.4cm}
    \caption{\textbf{A confusion matrix for the confident and unconfident sets by Consistency Checking method and Self-assessment method}. 
    % In our method, LLM is \textit{confident} to a question if the answers by the standard prompt and the IoE prompt are the same. When running 10 times, it is \textit{confident} if all 10 results are identical. 
}\label{fig:confusion_matrix}
    \vspace{-0.1cm}
\end{figure}

\begin{figure}[t!] %H为当前位置，!htb为忽略美学标准，htbp为浮动图形
    \centering %图片居中
    \includegraphics[width=0.5\textwidth]{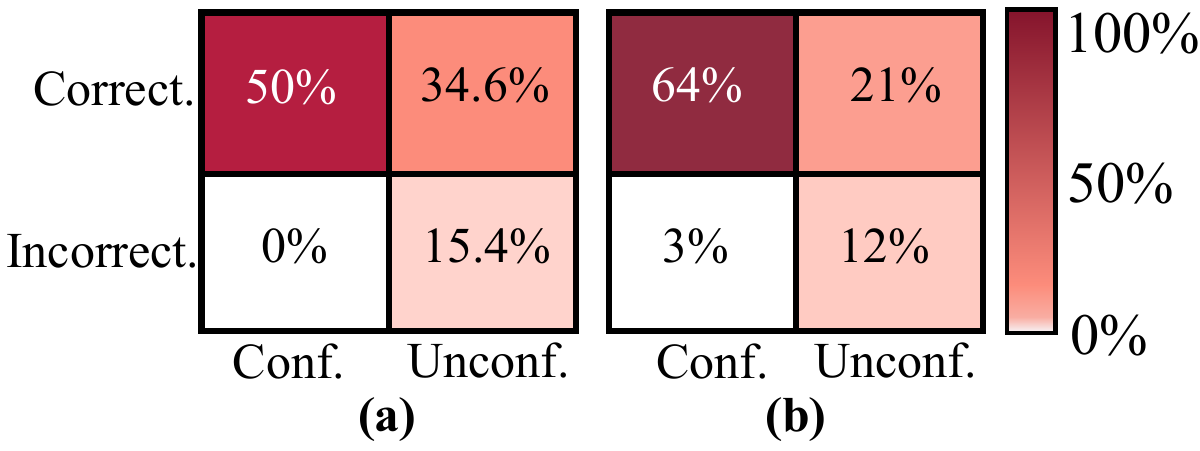}
    \setlength{\abovecaptionskip}{-0.4cm}
    \caption{\textbf{Confusion matrices comparing confident/unconfident and correct/incorrect classifications for (a) Consistency Checking and (b) Self-assessment}.
    % In our method, LLM is \textit{confident} to a question if the answers by the standard prompt and the IoE prompt are the same. When running 10 times, it is \textit{confident} if all 10 results are identical. 
}
    \label{fig:confusion_matrix-2}
    \vspace{-0.3cm}
\end{figure}

% \begin{table}[t!]
% \small
% \centering
% \resizebox{0.46\textwidth}{!}{
% \begin{tabular}{llc} 
% \toprule
% \textbf{Repetition} & \textbf{Prompt} & \textbf{Accuracy} \\
% \midrule
% \multirow{2}{*}{1 time} & Standard Prompt & 94\% \\
% & \ \ + IoE Prompt & \textbf{95\%} \\
% \midrule
% \multirow{2}{*}{10 times} & Standard Prompt & -\% \\
% & \ \ + IoE Correction & \textbf{-\%} \\
% \bottomrule
% \end{tabular}
% }
% \setlength{\abovecaptionskip}{0.1cm}
% \caption{\textbf{Comparison of different confidence assessment methods: single IoE Prompt vs. 10 iterative runs for consistency check.} These experiments were conducted on 100 new questions such as ``Can you provide a leap year between 1000 and 2000?'' using the \texttt{gpt-3.5-turbo-1106} model.}
% \label{tab:compare-1-10-new}
% \vspace{-0.3cm}
% \end{table}

% \section{If-or-Else Prompting}
\section{Observations}
\label{sec2:motivation}
 
In this section, we present observations and analysis from four experiments to explore confidence of LLMs' responses. We attempt to answer two questions: whether LLMs can assess the confidence of their own answers, and how the assessed confidence affects the intrinsic self-correction process. 
% We employ \texttt{gpt-3.5-turbo-1106} model with the temperature set to 0 for all experiments in this section.\footnote{According to OpenAI document, lower values of temperature will make it more focused and deterministic. However, even though we set it to 0, there are still variations.}
% Most experiments are conducted on randomly selected 100 questions (The indices are given in Appendix \ref{appendix:index}) from the GSM8K dataset \cite{cobbe2021training}. Beyond that, we created a dataset with open-ended mathematical questions, such as ``xxx'', to further illustrate the ability of LLMs to assess confidence. Please refer to xxx for the details of this dataset. 

\subsection{Can LLMs Assess Confidence by Themselves?}
% \subsection{Advantage of LLM Assessed Confidence.}
To answer this question, we present the results of confidence assessment by LLMs in both deterministic and open tasks. \arxiv{The deterministic tasks exactly have one unique solution, while the open tasks possibly have multiple solutions.} 
The \textit{Consistency Checking Method} assesses confidence by running the inference multiple times and checking whether they are consistent. \arxiv{The model is \textit{confident} towards a question when its answers remain consistent across multiple runs.} The \textit{Self-assessment Method} assesses the model with prompts such as ``If you are confident about your answer, maintain your answer. Otherwise, update your answer.'' \arxiv{The model is \textit{confident} to a question when the answers before and after self-assessment are consistent.}

\noindent\textbf{Assessing confidence for deterministic tasks.} \ 
We employ \texttt{gpt-3.5-turbo-1106} model with the temperature set to 0.\footnote{According to the OpenAI document, lower values of temperature will make it more focused and deterministic. However, even though we set it to 0, there are still variations.} Experiments are conducted on randomly selected 100 questions (The indices are given in Appendix \ref{appendix:index}) from the GSM8K dataset \cite{cobbe2021training}. We named this sub-dataset as GSM8K-100. As shown in the confusion matrix in Figure \ref{fig:confusion_matrix}, we compared the confident and unconfident question sets assessed by the self-assessment method and the consistency checking method.
The observed significant overlap in the classification of confident and unconfident questions indicates that self-assessment can efficiently evaluate confidence levels for deterministic tasks with a single run.

\begin{table}[t!]
\centering
% \small 
\resizebox{0.47\textwidth}{!}{
\begin{tabular}{lcc} 
\toprule
\textbf{Prompts} & \textbf{Standard} & \textbf{Standard + Critical} \\
\midrule
\textbf{Confident} (87/100) & 82.75\% & 77.35\% \\
\textbf{Unconfident} (13/100) & 27.69\% & 39.23\% \\
\textbf{Total} (100/100) & 75.60\% & 72.40\% \\
\bottomrule
\end{tabular}}
\setlength{\abovecaptionskip}{0.1cm}
\caption{\textbf{Evaluation of critical prompt's impact across different confidence levels.} 
% We randomly select 100 questions from GSM8K \cite{cobbe2021training}, and run 10 times by \texttt{gpt-3.5-turbo-1106}. LLM is \textit{confident} to a question \textit{if the answers by the standard prompt over the 10 runs are the same}. 
For the unconfident questions, LLM achieved better self-correction.}\label{tab:preliminary}
% \vspace{-0.3cm}
\end{table}

\begin{table}[t!]
% \small
\centering
\resizebox{0.46\textwidth}{!}{
\begin{tabular}{llc} 
\toprule
\textbf{Repetition} & \textbf{Prompt} & \textbf{Accuracy} \\
\midrule
\multirow{3}{*}{1 time} & Standard Prompt & 75\% \\
& \ \ + Critical Prompt & 73\% \\
& \ \ + Self-assessment & \textbf{80\%} \\
% & \ \ + IoE-based Prompt + Decision & \textbf{80\%} \\
\midrule
\multirow{2}{*}{10 times} & Standard Prompt & 75.60\% \\
& \ \ + Consistency Checking & \textbf{77.09\%} \\
\bottomrule
\end{tabular}
}
\setlength{\abovecaptionskip}{0.1cm}
\caption{\textbf{Comparison of different confidence assessment methods: one-trial Confidence Self-assessment vs. multi-trial Consistency Checking.} 
% These experiments were conducted on 100 chosen questions from the GSM8K dataset using the \texttt{gpt-3.5-turbo-1106} model.
}
\label{tab:compare-1-10}
\vspace{-0.3cm}
\end{table}

\noindent\textbf{Assessing confidence for open tasks.} \ 
% The results of open-ended tasks are shown in Table~\ref{xxx}. Due to there is no deterministic results, the method assessing the confidence with checking consistency failed.  Interestingly, we observe that LLMs can still assess meaningful confidence level. It demonstrates that LLMs have the inherent ability to understand the confidence of their response.
To test the effectiveness of confidence assessment methods in open tasks, we created a task with multiple correct answers. An example question is: \textit{Can you provide a year that is not a leap year and is divisible by 5?}
The details of datasets can be found in Appendix \ref{data_gen}. We use GPT-4 in this experiment due to its better consistency. 
The results are presented in Figure~\ref{fig:confusion_matrix-2}. 
In these tasks, the non-deterministic answers leads the consistency checking method towards classifying answers as low confidence.
However, it is noteworthy that despite this limitation, LLMs demonstrated the capability to assess meaningful confidence levels in their responses. This observation is particularly significant as it highlights the inherent ability of LLMs to understand and quantify the confidence of their responses, even in the absence of clear and deterministic answers.

% \textbf{Takeaway}: \textit{\ul{LLMs can efficiently and effectively assess confidence of their own response.}}

\textbf{Takeaway}: {\ul{\textit{LLMs can efficiently and effectively assess confidence of their own responses.}}}

% \subsection{How does Confidence Affect Self-Correction?} 
% In this subsection, we further discuss the influence of confidence for the self-correction of LLMs. We first evaluate the impact of self-correction methods at varying levels of confidence. Furthermore, we observe that involving the estimated confidence, LLMs can perform self-correction better. 

\subsection{How does Confidence Affect Intrinsic Self-Correction?}
In this subsection, we delve deeper into the role of confidence in the intrinsic self-correction processes of LLMs. 
Initially, we evaluate the impact of self-correction methods at varying levels of confidence. Subsequently, our observations find that incorporating the estimated confidence enables LLMs to enhance their self-correction capabilities, by retaining the initial answer with high confidence and reviewing more for the ones with low confidence. The experiments are based on GSM8K-100 and \texttt{gpt-3.5-turbo-1106}.

\noindent\textbf{Self-correction at varying levels of confidence.} \ First, we applied the consistency checking method to separate the responses into two subgroups: confident and unconfident ones. Then we compared the results using the standard prompt against those using the further self-corrective Critical Prompt ~\citep{huang2023large}. For simplicity, we combined the second and third stages of Critical Prompt shown in the left part of Figure \ref{fig:shortexample}, phrasing it as \textit{``There are some problems in your previous answer. Find problems with your answer and improve your answer.''}
We report the accuracy of different confidence subgroups given different prompts. 
The accuracy is averaged out of 10 runs.    

The results are presented in Table \ref{tab:preliminary}. According to the results, for confidence level, we found there are 87 confident and 13 unconfident questions out of 100 in total.
We observed that across all 100 questions, LLM failed in intrinsic self-correction, as evidenced by a decrease in accuracy from 75.6\% to 72.4\% after executing the critical prompt. Intriguingly, for the subgroup of 13 unconfident questions, the accuracy increased from 27.69\% to 39.23\%. Despite the relatively low accuracy, LLM successfully performed self-correction in the unconfident set. Additionally, the same self-correction prompt exhibits varying performance depending on different levels of confidence.
These results offer empirical evidence and support that confidence matters in the self-correction process.

% \noindent\textbf{Self-correction using assessed confidence.} \  We evaluate both confidence assessment methods (using LLM to self-assess and checking consistency) by the performance of self-correction. Specifically, we apply the prompt shown in Figure~\ref{fig:shortexample} to guide LLMs to jointly assess confidence and conduct self-correction. 
% When evaluating the method inferencing 10 times and checking consistency, we first assess the condence, and then apply this condience result to decide whether to do self-correction using Cirtical Prompt.  

\noindent\textbf{Self-correction using assessed confidence.} \ We evaluate the effectiveness of two confidence assessment methods for enhancing LLMs' self-correction: assessment by the LLM and assessment through consistency checking. 
When evaluating the LLM assessment, we apply the prompt shown in Figure~\ref{fig:shortexample} to guide LLMs in simultaneously assessing confidence and performing self-correction. 
For Consistency Checking, we first assess the confidence level, then apply it to decide the application of the Critical Prompt for self-correction (only conduct Critical Prompt for unconfident answers).

Table \ref{tab:compare-1-10} details the performance comparison between two confidence assessment methods. The results show that both methods effectively facilitate self-correction, suggesting the importance of confidence in this process. Notably, LLM assessment method not only demonstrates improved efficiency but also outperforms the consistency check method in effectiveness, showing a superior ability to assess confidence.

\textbf{Takeaway}: {\ul{\textit{
Understanding confidence enhances self-correction, making this process more adaptive and preventing over-criticism 
.}}}

\section{Method}
\label{sec3:ioe}

% In this section, we introduce our proposed If-or-Else (IoE) prompting principle in detail, which apply the prompt to guide the LLM to jointly assess and confidence and review the answers.  

In this section, we elaborate on If-or-Else (IoE) prompting principle, which utilizes prompts to guide the LLMs in simultaneously assessing confidence levels and correcting unconfident answers.

% use a single prompt to ask and incorporate the confidence of LLM. 

% As mentioned in the previous section, it is inefficient to run the question multiple times to determine whether LLM is confident or not. Instead, 

% \noindent\textbf{Insights.} \ Inspired by the aforementioned observation, we propose the hypothesis that LLMs have the inherent ability to understand and assess their confidence, a capability that can be used to enhance self-correction. 
% This hypothesis could explain why self-correction results in worse performance in \citep{huang2023large}: The Critical Prompt overlooks the impact of confidence, leading to an over-critique of all answers and consequently causing significant misleading, especially for the initially correct answers.

\noindent\textbf{IoE Prompt.} \
We propose a hypothesis that LLMs have the inherent ability to understand and assess their confidence.
Contrasting with Critical Prompt \cite{huang2023large}, which directly instructs to \textit{``find problems with your answer''}, our IoE-based prompt emphasizes the significance of confidence through the If-or-Else principle. This principle indicates that if the LLM considers itself confident, the response should be left as is. On the other hand, if the LLM thinks itself unconfident, it should double-check the possible issues and revise the response based on the results.  Specifically, as illustrated in Figure~\ref{fig:shortexample}, we implement the prompt as {\textit{"If you are very confident about your answer, maintain it. Otherwise, update your answer."}}
\arxiv{We name this prompt as \textit{IoE Prompt}. Please note that, in the following contents, we use \textit{IoE-based Prompt} to generally represent our proposed prompting method, while \textit{IoE Prompt} refers to the single prompt for self-correction.}
Following the above IoE Prompt, we trigger the model to output the final answer by saying: \textit{"Your final answer should be put between two \#\# symbols, like \#\# ANSWER \#\#, at the end of your response."}.

It's important to note that our single IoE prompt effectively combines the processes of confidence assessment, response reviewing, and potential updates. This method requires only a single inference, making it more efficient than the two-stage update process described in \cite{huang2023large}, where a \textit{critical prompt} is first used to identify problems, and then an \textit{update prompt} follows to revise the answers based on the identified issues.

\noindent\textbf{Decision Refinement.} \ Furthermore, we introduce a decision refinement stage to further enhance results when the answer after IoE Prompt differs from the initial response. This stage shares the insight that answers reflecting low confidence require additional evaluation and refinement. The discrepancy in answers suggests low confidence, indicating the need for further decision refinement. Specifically, the decision refinement is conducted using the prompt: {\textit{"You have provided two different answers in previous responses. Review the problem and your answers again, and provide the best answer."}} This process is illustrated on the right side of Figure \ref{fig:shortexample}.
Consider a scenario where the LLM, uncertain about an initially correct answer, changes its answer to an incorrect one. In such cases, the third prompt becomes crucial, allowing for a reevaluation of the responses to potentially yield a more reliable final answer.

\section{Models and Benchmarks}
\label{sec4:setup}

Throughout our paper, we considered 4 different large models, and one of them can be used as a multi-modal model. We set the temperature to 0 for all models, aiming at more robust evaluations.
% The answers by all those models are expected to be more accurate rather than more creative, therefore we set the temperature to 0 for all models. 
We evaluated those models on 6 benchmarks, whose information is summarized in Table \ref{tab:benchmark}.
Below are more details about the models and benchmarks.

% The benchmark information is s

% We report the accuracy of all questions for each benchmark.

% \footnotetext{In our experiments, we separately run our prompts and baseline prompts, and the accuracy of standard prompt by baseline may differ a bit from that by ours. For fair comparison in Table \ref{tab:main_results}, we normalize the baseline results so that the accuracy of standard prompts by baseline and ours are equal.}

% In our experiments, we consider 4 different large models, including GPT-4, GPT-3.5-Turbo-1106, GPT-3.5-Turbo-0613, and Mixtral-8x7B-Medium. We evaluate those models on 5 different benchmarks, including GSM8K, SVAMP, HotpotQA, Sports understanding, and Last Letter Concatenation (LLC). Below are more details.

\subsection{Models}

\textbf{GPT-4} \cite{achiam2023gpt} is one of state-of-the-art LLMs known for its advanced reasoning abilities. We utilize the \texttt{gpt-4-0613} version of this model. GPT-4 also supports multi-modal reasoning, thus for those tasks, we employ GPT-4V, specifically the \texttt{gpt-4-vision-preview} version. For ease of reference, we will refer to these models as GPT-4 and GPT-4V, respectively.

% , presenting significant improvement over other GPT models. In our following experiments

% is one of the state-of-the-art LLMs with advanced reasoning capabilities. Here we use \texttt{gpt-4-0613} in our following experiments. Besides, it also enables the multi-modal reasoning. We also use GPT-4V for self-correction on multi-modal scenarios, referred to as \texttt{gpt-4-vision-preview}. For simplicity, we call them GPT-4 and GPT-4V, respectively.
% stands as an expansive multi-modal model, capable of processing both text and image inputs while producing textual outputs. It excels in solving challenging problems with advanced reasoning capabilities, presenting higher accuracy than other GPT models. Here we opt for \texttt{gpt-4-0613}. We also use GPT-4V for self-correction on multi-modal scenarios, referred to as \texttt{gpt-4-vision-preview}. For simplicity, we call them GPT-4 and GPT-4V, respectively.

\begin{table}[t!]
% \small
\centering
\resizebox{0.48\textwidth}{!}{
\begin{tabular}{llc} 
\toprule
\textbf{Benchmark} & \textbf{Reasoning Type} & \textbf{\# Questions}\\
\midrule
GSM8K & Arithmetic Reasoning & 1319 \\
SVAMP & Arithmetic Reasoning & 1000  \\
HotpotQA & Multi-hop Reasoning & 100 \\
Sports & Commonsense Reasoning & 1000 \\
LLC & Symbolic Reasoning &  150  \\
Domestic Robot & Multi-modal Reasoning & 100 \\
% Art \& Design & Multi-modal Reasoning & 110 \\
\bottomrule
\end{tabular}
}
\setlength{\abovecaptionskip}{0.1cm}
\caption{\textbf{Statistics of various benchmarks: types and number of questions.}}
\label{tab:benchmark}
\vspace{-0.3cm}
\end{table}

\noindent\textbf{GPT-3.5} exhibits proficiency in generating natural language and code. The standout model within the GPT-3.5 family is \texttt{gpt-3.5-turbo}, renowned for its exceptional capabilities and cost-effectiveness. Here we opt for two models, i.e., \texttt{gpt-3.5-turbo-0613} and \texttt{gpt-3.5-turbo-1106}. For simplicity, we call them GPT-3.5-0613 and GPT-3.5-1106, respectively.

% \noindent\textbf{GPT-3.5} can understand and generate natural language or code. The most capable and cost effective model in the GPT-3.5 family is \texttt{gpt-3.5-turbo} which has been optimized for chat using the Chat Completions API but works well for traditional completions tasks as well. 

\noindent\textbf{Mixtral} \cite{jiang2024mixtral} is the first open-source mixture-of-experts LLM to reach the state-of-the-art performance. The small model, named \texttt{mistral-small}, refers to the standard Mixtral 8x7B model. Here we apply for \texttt{Mistral-Medium}, which has a larger structure and better performance than the small one.

\subsection{Benchmarks}

\textbf{GSM8K} \cite{cobbe2021training} is a dataset of high-quality linguistically diverse grade school math word problems created by human problem writers. The solutions primarily involve performing a sequence of elementary calculations. We use the test set with 1319 problems.

\noindent\textbf{SVAMP} \cite{patel2021nlp} is another dataset for elementary-level math word problems. Each question consists of a short natural language narrative that describes a state of the world and poses a question about some unknown quantities. There are 1000 questions in total.

\noindent\textbf{HotpotQA} \cite{yang2018hotpotqa} is a question-answering dataset featuring multi-hop reasoning, with strong supervision for supporting facts to enable more explainable question-answering systems. We use the same dataset as \cite{shinn2023reflexion}, consisting of 100 questions with context.

\begin{table*}[t]
	\small
	\centering
\newcolumntype{g}{>{\columncolor{Gray}}c}
 \renewcommand{\arraystretch}{1.0}
 \setlength{\tabcolsep}{3.3pt}
	\begin{tabular}{llcccccg}
		\toprule
		&              & \textbf{GSM8K} & \textbf{SVAMP} & \textbf{HotpotQA} & \textbf{Sports} & \textbf{LLC}  & \textbf{Average}\\\midrule
		\multirow{4}{*}{GPT-3.5-0613} & Standard Prompt  & 74.9  & 82.2  & 51.0  & 75.6 & 68.0   &   70.3\\
		& \ \ + Critical Prompt \citep{huang2023large} &   74.1    &   80.0    &   47.0   & 53.6 & 76.0 & 66.1\\
            & \ \ + IoE Prompt (Ours)   & 77.1  & 81.9  &  \textbf{55.0}     & \textbf{77.1}   & 74.0    &  73.0\\
		& \ \ + IoE Prompt + Decision (Ours)   & \textbf{78.5}  & \textbf{83.3}  &  53.0   & 76.5  & \textbf{77.3}    & \textbf{73.7} \\\midrule
		\multirow{4}{*}{GPT-3.5-1106} & Standard Prompt     & 80.1    & 82.9  & 61.0  & 74.1 & 41.3     & 67.9 \\
		& \ \ + Critical Prompt \citep{huang2023large} &   77.3    &   81.5    &    54.0    &  68.4  &  40.7 & 64.4 \\
            & \ \ + IoE Prompt (Ours)   & 80.9  & 83.2  & 62.0  &  \textbf{75.7} & 38.7 & 68.1\\
		& \ \ + IoE Prompt + Decision (Ours)  & \textbf{82.3}  & \textbf{84.2}    & \textbf{63.0}     & 74.7  & \textbf{44.7} &  \textbf{69.8} \\\midrule
		\multirow{4}{*}{GPT-4}        & Standard Prompt     & 92.5  & 92.8  & 68.0     & 80.7  & 91.3 & 85.1\\
		& \ \ + Critical Prompt \citep{huang2023large}&   88.4    &   89.5    &   62.0     &  82.9  & 89.9 & 82.5\\
		& \ \ + IoE Prompt (Ours)   & 93.4  & \textbf{93.2}  & \textbf{70.0}     & 83.1  & 93.3   & 86.6 \\
            & \ \ + IoE Prompt + Decision (Ours)   & \textbf{93.6}  & 93.1  & \textbf{70.0}     & \textbf{83.3}  & \textbf{94.7}   &  \textbf{86.9} \\\midrule
            \multirow{4}{*}{Mistral-Medium}        & Standard Prompt    & 84.8  & 85.7  & 67.0   & 75.6 & 60.7  &  74.8 \\
		& \ \ + Critical Prompt \citep{huang2023large}   &   62.5    &   74.5     & 65.0  & 51.0  & 35.4  &57.7 \\
		& \ \ + IoE Prompt (Ours)   & 85.4  & 85.7     & \textbf{68.0}   &  75.6  & \textbf{61.3} &  75.2\\
            & \ \ + IoE Prompt + Decision (Ours)   & \textbf{85.6}  & \textbf{85.8}  &  \textbf{68.0}  & \textbf{75.9} & \textbf{61.3}  &  \textbf{75.3}\\\bottomrule
	\end{tabular}
        \setlength{\abovecaptionskip}{0.1cm}
		\caption{\textbf{The accuracy comparisons between our IoE-based Prompt and the baseline Critical Prompt}. The results (\%) are evaluated on 5 different benchmarks by 4 large models.  IoE Prompt $+$ Decision denotes further using the decision refinement stage. Our IoE Prompt achieves consistent improvement on all settings over standard prompt and the Critical Prompt baseline.
}
\label{tab:main_results}
\vspace{-0.3cm}
\end{table*}

\noindent\textbf{Sports} is from BIG-Bench \cite{srivastava2023beyond}. The primary purpose of this task is to test a general understanding of sports. The answer format is either Yes or No. This dataset contains 1000 questions.

\noindent\textbf{Last Letter Concatenation (LLC)} was initially designed by \cite{wei2022chain}. This task asks the model to concatenate the last letters of words in a name (e.g., "Taylor Swift" $\rightarrow$ "rt"). It is more challenging than first letter concatenation. This dataset contains 150 names.

\noindent\textbf{Domestic Robot} simulates a housekeeper robot performing tasks
within a household setting. This dataset has 100 questions and is introduced by BenchLMM \cite{cai2023benchlmm}. It's designed to assess the LMM's ability to understand and execute complex instructions in a multi-modal scenario. 

% \noindent\textbf{Art \& Design} is one of the six common disciplines from MMMU \cite{yue2023mmmu}. MMMU is designed to evaluate LMMs on massive multi-discipline tasks demanding college-level subject knowledge and deliberate reasoning. We use the validation set of the art \& design discipline here, which has 110 questions in total.

% \noindent\textbf{WinoGrande} is a large-scale dataset of 44k problems, inspired by the original WSC design, but adjusted to improve both the scale and the hardness of the dataset. The key steps of the dataset construction consist of (1) a carefully designed crowdsourcing procedure, followed by (2) systematic bias reduction using a novel AfLite algorithm that generalizes human-detectable word associations to machine-detectable embedding associations.

\section{Experiments}

In this section, we evaluate the effectiveness of IoE-based Prompt for intrinsic self-correction. Initially, we conduct a thorough comparison between our IoE-based Prompt and the baseline Critical Prompt \cite{huang2023large} (Section~\ref{sec5-1}). Subsequently, we experimentally 
% understand the reasons behind the effectiveness of IoE Prompt(Section~\ref{sec5-2}) and 
explore the underlying mechanics in it (Section~\ref{sec5-3}). More implementation details are demonstrated in Appendix \ref{appendix:implement_details}.

\subsection{Comparison with Critical Prompt} 
\label{sec5-1}

The primary baseline method in this study is the Critical Prompt \citep{huang2023large}. By comparing our approach with this method, we can directly observe the significance of confidence in LLMs.
% To conduct a comprehensive evaluation of the effectiveness of our IoE Prompt, 
We performed comprehensive evaluations using four different large-scale models across six benchmark datasets. These evaluations spanned five types of tasks, covering both language-based reasoning and multi-modal reasoning.

% To comprehensively evaluate the effectiveness of our IoE Prompt, we conducted evaluations using four distinct large-scale models across six benchmark datasets, covering five types of tasks, across the language understanding and reasoning, and multimodal reasoning. 

% The main results related to language-based reasoning are summarized in Table \ref{tab:main_results}. Additionally, the multi-modal results, specifically for the Domestic Robot scenario using GPT-4V, are detailed in Table \ref{tab:gpt-4v}.

% The main results on language-based reasoning are summarized in Table \ref{tab:main_results}, and the multi-modal results on Domestic Robot by GPT-4V are shown in Table \ref{tab:gpt-4v}.
% To thoroughly assess the effectiveness of our If-or-Else (IoE) Prompt, we conducted evaluations using four distinct large-scale models across six benchmark datasets, covering five types of reasoning tasks. The principal findings related to language-based reasoning are summarized in Table \ref{tab:main_results}. Additionally, the multi-modal reasoning results, specifically for the Domestic Robot scenario using GPT-4V, are detailed in Table \ref{tab:gpt-4v}.

% One interesting result is that the critical prompt can achieve self-correction on LLC, showing accuracy improvement from 68\% to 76\%. 
% The results present a comprehensive overview of accuracy. 

\noindent\textbf{Language Understanding and Reasoning.} \ The main results related to language-based understanding and reasoning are summarized in Table \ref{tab:main_results}. Specifically, we evaluated four models: GPT-3.5-0613, GPT-3.5-1106, GPT-4, and Mistral-Medium. The tasks cover a range of areas, including mathematical reasoning ( GSM8K and SVAMP), closed-book question answering (HotpotQA), sports understanding (Sports), and last letter concatenation (LLC). For all models, the proposed IoE-based Prompt consistently demonstrates improvements across all benchmarks compared to the standard prompt (don't conduct self-correction), and Critical Prompt. For a straightforward comparison, we also provided the average performance across all five datasets. Using the GPT-3.5-0613 model as an example, our IoE-based Prompt obtains an average accuracy of 73.7$\%$, representing a $+3.4\%$ improvement over the standard prompt and a $+7.6\%$ increase compared to the Critical Prompt.
Across different tasks, we found that IoE-based Prompt works better on mathematical reasoning and LLC tasks.
Besides, we observed a trend showing that the magnitude of improvement from our IoE-based Prompt decreases as the capability of the model increases.
On average, we found that the additional decision refinement stage consistently enhances performance across all models. This consistent improvement supports our insight that confidence matters in intrinsic self-correction, and responses with low confidence require more examination.

\begin{table}[t!]
% \small 
\centering
\resizebox{0.34\textwidth}{!}{
\begin{tabular}{lc} 
\toprule
\textbf{Prompt} & \textbf{Accuracy}   \\
\midrule
 Standard Prompt & 40\%  \\
 \midrule
 \ \ + Critical Prompt & 34\%  \\
 \ \ + IoE Prompt & \textbf{42\%}  \\
 \ \ + IoE Prompt + Decision & \textbf{42\%}  \\
\bottomrule
\end{tabular}
}
\setlength{\abovecaptionskip}{0.1cm}
\caption{\textbf{Comparisons between IoE-based Prompt and Critical Prompt on Domestic Robot by GPT-4V.}}
\label{tab:gpt-4v}
\vspace{-0.4cm}
\end{table}

% Across different models, we observed that the improvement decrease when the model capility is stronger. \
% For GPT-3.5-0613, our proposed prompts consistently demonstrate improvements across most benchmarks compared to the standard prompt and the critical prompt. Notably, the result with the decision prompt yields the highest accuracy in GSM8K, SVAMP, and LLC, showcasing its effectiveness in enhancing model performance. Similar trends are observed for GPT-3.5-1106, where our prompts outperform the standard prompt and the critical prompt in most cases. The one with the decision prompt achieves the highest accuracy in the four benchmarks, highlighting its efficacy in refining responses. 
% GPT-4 exhibits superior accuracy across the board, with our prompts further enhancing its performance. The decision prompt consistently produces the highest accuracy, underscoring its effectiveness in self-correcting the answers.
% Mistral-Medium, while generally exhibiting lower accuracy, still benefits from our prompts, particularly the decision prompt, which consistently outperforms the other prompt types. 

\noindent\textbf{Multi-Modal Reasoning.} \ For the evaluation of our IoE-based Prompt in multi-modal reasoning tasks, we utilized GPT-4V as the base model, conducting benchmarks on the Domestic Robot dataset. The results, detailed in Table \ref{tab:gpt-4v}, demonstrate that our method outperforms the standard prompt and the Critical Prompt baseline by +2\% and +8\% in accuracy, respectively. This suggests the effectiveness of the IoE principle in improving self-correction within multi-modal scenarios.

Some successful and failed examples with the complete prompts and responses are given in Appendix \ref{appendix:success} and Appendix \ref{appendix:failed}, respectively.

% \noindent\textbf{Multi-Modal reasoning.} To evaluate our IoE Prompt with the multi-modal reasoning task, we apply GPT-4V as the base model and benchmark models on the Domestic Robot dataset. The results are detailed in Table \ref{tab:gpt-4v}. Specifically, our outperform the standard prompt and the critical prompt with $+2\%$ and $+8\%$ accuracy, respectively. This suggests that incorporating IoE principle into the prompt can also improve the self-correction in multi-modal scenarios. 

% the multi-modal reasoning results, specifically for the Domestic Robot scenario using GPT-4V, are detailed in Table \ref{tab:gpt-4v}.

% Regarding the result for GPT-4V on the Domestic Robot benchmark, our IoE prompts outperform the standard prompt and the critical prompt. This suggests that incorporating IoE principle into the prompt can also improve the model's accuracy in multi-modal scenarios. 

% According to those results, we may conclude that our IoE-based prompts generally achieve intrinsic self-correction on various reasoning tasks. 

% In summary, GPT-4 generally outperforms GPT-3.5-0613, GPT-3.5-1106, and Mistral-Medium across benchmarks, demonstrating the advancements in model capabilities.
% Across all models, our proposed prompts consistently lead to accuracy improvements over the standard prompt and the critical prompt, emphasizing the efficacy of incorporating IoE and decision-making. 
% The inclusion of decision-making in the prompt consistently refines model responses, achieving the highest accuracy in several benchmarks across all models.

\begin{figure*}[t!] %H为当前位置，!htb为忽略美学标准，htbp为浮动图形
    \centering %图片居中
    \includegraphics[width=1\textwidth]{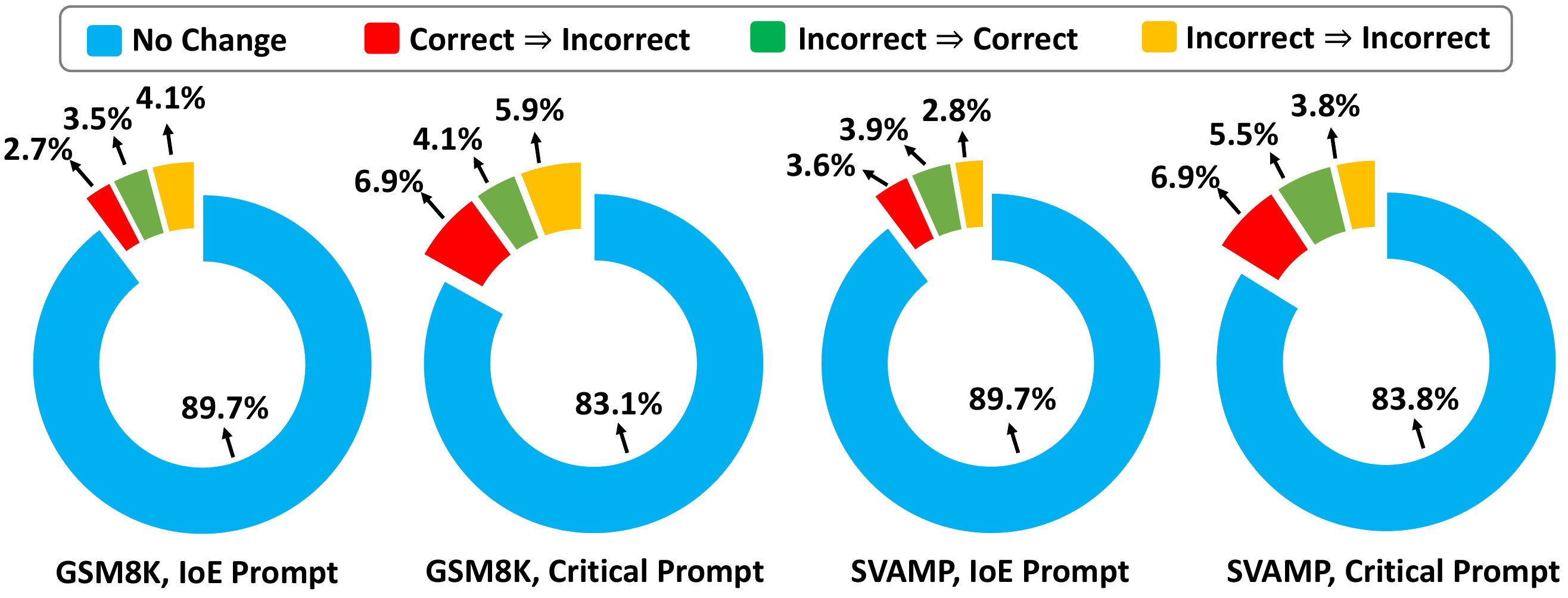}
    \setlength{\abovecaptionskip}{-0.4cm}
    \caption{\textbf{Evaluation on the changes of answers after self-correction.}  We compare the results of our IoE-based Prompt and Critical Prompt \cite{huang2023large} on GSM8K and SVAMP by \texttt{gpt-3.5-turbo-1106} model. \textit{No Change}: The answer remains unchanged. \textit{Correct} $\Rightarrow$ \textit{Incorrect}: A correct answer is altered to an incorrect answer. \textit{Incorrect} $\Rightarrow$ \textit{Correct}: An incorrect answer is changed to a correct answer. \textit{Incorrect} $\Rightarrow$ \textit{Incorrect}: An incorrect answer is changed to another incorrect answer.}
    \label{fig5}
    \vspace{-0.3cm}
\end{figure*}

\subsection{How IoE-based Prompt works}
\label{sec5-3}

In this section, we delve into the mechanics underlying our IoE-based Prompt through a series of ablation studies: 
1) We visualize the changes in LLMs' responses during the self-correction process
2) We evaluate the self-correction effectiveness by contrasting one-stage prompts with two-stage prompts.
3) We examine the impact of tone and politeness in the phrasing of prompts on their effectiveness.
4) We investigate how our IoE-based Prompt might incorporate with existing prompting techniques, such as the Chain-of-Thought (CoT) prompt \cite{wei2022chain}, and the Rephrase-and-Respond (RaR) prompt \cite{deng2023rephrase}.

% The complete prompts for the experiments above are provided in Appendix \ref{appendix:prompts}. 

% Here, we mainly evaluate using GSM8K \cite{cobbe2021training} by \texttt{gpt-3.5-turbo-0613} model.

\subsubsection{Correction {v.s.} Misleading}

\noindent\textbf{Setup.} For a detailed analysis, we compared the answer changes when using the Critical Prompt and our IoE-based Prompt. The GPT-3.5-1106 model was employed, with GSM8K and SVAMP serving as benchmark datasets. We statistically analyzed the responses, categorizing them based on whether they remained unchanged, changed from correct to incorrect, changed from incorrect to correct, or changed from one incorrect answer to another.

\noindent\textbf{Results and Analysis.} The results are illustrated in Figure \ref{fig5}. A key observation when comparing the IoE-based Prompt with the Critical Prompt is a notable decrease in the correct-to-incorrect transitions using our method. This suggests that our approach effectively reduces the risk of being misled by excessive criticism. Furthermore, self-correction is successful when the correct-to-incorrect probability is lower than the incorrect-to-correct probability. This observation provides insights into why the Critical Prompt baseline fails in self-correction, whereas our method achieves success.

\subsubsection{One-Stage {v.s.} Two-Stage}

\noindent\textbf{Setup.} 
% The Critical Prompt operates through a two-stage process for self-correction, whereas our IoE prompt streamlines the process by utilizing a single prompt for feedback and updates.
% To make the inference more efficient, we apply a single prompt to jointly perform feedback and updates, which is different from the two-stage process using in Critical Prompt. 
% This section aims to analyze and compare the efficacy of one-stage versus two-stage prompts for self-correction. 
To enhance inference efficiency, we employ a single prompt that simultaneously handles feedback and updates. It contrasts with the two-stage process (one prompt for finding problems, another for updating answers) utilized in Critical Prompt. The purpose of this section is to analyze and compare the effectiveness of one-stage prompts against two-stage prompts for self-correction.
The complete prompts are provided in Appendix \ref{appendix:tab7}. 

% \noindent\textbf{Results and Analysis.} Table \ref{tab:steps} presents a comparison between one-stage and two-stage strategies on both IoE Prompt and Critical Prompt. We can observe that they are almost compariable in both methods, where the two-stage ones is slightly better. However, after balancing the improvement and extra inference cost, we apply the one-stage strategy in our IoE Prompt for other experiments.

\noindent\textbf{Results and Analysis.} Table \ref{tab:steps} presents a comparative analysis of one-stage and two-stage strategies applied to both IoE-based Prompt and Critical Prompt. The results indicate that their performance is comparable across both methods, with the two-stage strategy demonstrating a slight advantage. Nonetheless, considering the trade-off between incremental improvement and additional inference overhead, we have adopted the one-stage strategy in our IoE-based Prompt for other experiments.

\begin{table}[t!]
\small 
\centering
% \resizebox{0.36\textwidth}{!}{
\begin{tabular}{lc} 
\toprule
\textbf{Prompt} & \textbf{Accuracy}   \\
\midrule
 Standard Prompt & 74.9\%  \\
 \midrule
 \ \ + Critical Prompt (One-Stage) & {73.7\%}  \\
 \ \ + Critical Prompt (Two-Stage) & {74.1\%} \\
 \midrule 
  \ \ + IoE Prompt (One-Stage) & {77.1\%}  \\
 \ \ + IoE Prompt (Two-Stage) & {77.5\%} \\
\bottomrule
\end{tabular}
% }
\setlength{\abovecaptionskip}{0.1cm}
\caption{\textbf{Comparison between one-stage and two-stage prompts} on GSM8K by \texttt{gpt-3.5-turbo-0613}.}
\label{tab:steps}
\vspace{-0.15cm}
\end{table}

\begin{table}[t!]
\small 
\centering
% \resizebox{0.49\textwidth}{!}{
\begin{tabular}{lc} 
\toprule
\textbf{Prompt} & \textbf{Accuracy}   \\
\midrule
 Standard Prompt & 74.9\%  \\
 \midrule
 \ \ + IoE Prompt \textit{w.} `very' & {77.1\%}  \\
 \ \ + IoE Prompt \textit{w.o.} `very' & {76.4\%} \\
 \midrule 
 \ \ + IoE Prompt \textit{w.} `please' & {77.9\%}  \\
 \ \ + IoE Prompt \textit{w.o.} `please' & {77.1\%} \\
 \midrule
  \ \ + IoE Prompt \textit{w.} ``find your problems'' & {75.9\%}  \\
 \ \ + IoE Prompt \textit{w.o.} ``find your problems'' & {77.1\%} \\
\bottomrule
\end{tabular}
% }
\setlength{\abovecaptionskip}{0.1cm}
\caption{\textbf{Ablation study of IoE-based Prompt on tone and politeness} on GSM8K by \texttt{gpt-3.5-turbo-0613}.}
\label{tab:ablation2}
\vspace{-0.45cm}
\end{table}

\subsubsection{Impact of Tone and Politeness}
% Furthermore, we examined the impact of tone and politeness in our prompts. 
% Specifically, we evaluated whether the adverb of degree ``very'', the mild tone with ``update your answer'', and the politeness with the word ``please'' help the self correction. 

\noindent\textbf{Setup.} We investigated the influence of tone and politeness in our prompts. Specifically, we assessed the role of the degree adverb ``very'', the gentle tone conveyed by ``update your answer'', and the addition of politeness through the use of ``please'' in facilitating self-correction. The complete prompts are provided in Appendix \ref{appendix:tab8}. 

\noindent\textbf{Results and Analysis.} The experiment result is given in Table \ref{tab:ablation2}. 
First, we can find that the expressions ``confident'' and ``very confident'' lead to similar accuracy, which indicates the robustness of the confidence assessment. Given that the use of ``very confident'' resulted in higher accuracy, we employ it in our IoE-based Prompt.
% Firstly, we observed that the use of ``very confident'' in our prompts resulted in higher accuracy than the use of ``confident'', suggesting that greater emphasis on confidence levels correlates with improved performance or reliability in the model's responses.
Secondly, the use of ``please'' before verbs such as ``maintain'' and ``update'' achieves a slight improvement in accuracy, indicating that politeness could be helpful for self-correction. Please note that we did not apply the ``please'' trick in other experiments to highlight the unique effect of confidence.
% does not improve the performance at all, indicating that employing a polite tone here is not helpful in enhancing the efficacy of self-correction. 
Thirdly, upon adding the phrase ``find your problems'' before ``update your answer'', a notable decrease in accuracy was observed. This indicates that specific terms such as ``problem'' or ``error'' might instigate doubt within the LLMs regarding its initial response, potentially leading to less reliable outcomes. Conversely, the more neutral term ``update your answer'' seems to be able to avoid this, consequently benefiting self-correction processes.

\begin{table}[t!]
\small 
\centering
% \resizebox{0.36\textwidth}{!}{
\begin{tabular}{lc} 
\toprule
\textbf{Prompt} & \textbf{Accuracy}   \\
\midrule
 Standard Prompt \textit{w.} CoT & 74.9\%  \\
 \midrule
 % \ \ + Critical Prompt & 34\%  \\
 \ \ + IoE Prompt \textit{w.} CoT & {76.6\%}  \\
\ \ + IoE Prompt \textit{w.o.} CoT & {77.1\%}  \\
\midrule 
% \cmidrule(lr){1}
 \ \ + IoE Prompt \textit{w.} RaR & 68.4\% \\
 \ \ + IoE Prompt \textit{w.o.} RaR & {77.1\%}  \\
 % \ \ \ \ \ + Decision Prompt & \textbf{78.5\%}  \\
 % \ \ \ \ \ + Decision Prompt & \textbf{78.5\%}  \\
 \midrule
 Standard Prompt \textit{w.o.} CoT & 73.4\%  \\
 \midrule
 % \ \ + Critical Prompt & 34\%  \\
\ \ + IoE Prompt \textit{w.} CoT & {77.1\%} \\
\ \ + IoE Prompt \textit{w.o.} CoT & {75.5\%}   \\
\midrule
% \cmidrule(lr){1}\cmidrule(lr){2}
 \ \ + IoE Prompt \textit{w.} RaR & 65.1\% \\
 \ \ + IoE Prompt \textit{w.o.} RaR & {75.5\%}   \\
 % \ \ \ \ \ + Decision Prompt & \textbf{78.5\%}  \\
 % \ \ \ \ \ + Decision Prompt & \textbf{78.5\%}  \\
\bottomrule
\end{tabular}
% }
\setlength{\abovecaptionskip}{0.1cm}
\caption{\textbf{Results when integrating CoT~\citep{wei2022chain} and RaR~\cite{deng2023rephrase} into our IoE-based Prompt} on GSM8K by \texttt{gpt-3.5-turbo-0613}.}
\label{tab:ablation1}
\vspace{-0.3cm}
\end{table}

\subsubsection{Integration with CoT and RaR}

\noindent\textbf{Setup.} We explored the possible integration of our IoE-based Prompt with established prompting methods for self-correction, specifically for CoT and RaR. The CoT method prompts the model to process information step-by-step,  yielding promising performance in many reasoning tasks. Meanwhile, RaR encourages LLMs to ask themselves better questions by rephrasing and responding. The complete prompts are provided in Appendix \ref{appendix:tab9}. 

% Specifically, we attempt to integrate CoT or RaR into the standard prompt and IoE prompt, respectively. 
% To add CoT, we \cgy{add ``think it step by step''  xxxx }
% For RaR  \cgy{add ``think it step by step''  xxxx }

% how our IoE Prompt might incorporate with existing prompting techniques, We also try to connect our proposed IoE prompting principle with some other popular principles, such as Chain-of-Thought (CoT) and Rephrase-and-Respond (RaR). CoT requires the model to think step-by-step, and has shown promising results in various reasoning tasks. RaR allows LLMs to rephrase and expand questions posed by humans and provide responses in a single prompt, presenting effective performance in accuracy.

\noindent\textbf{Results and Analysis.}  The results are summarized in Table~\ref{tab:ablation1}. Using CoT in the standard prompt resulted in a higher accuracy rate of 74.9\%, compared to 73.4\% without CoT. Intriguingly, once the standard prompt is with CoT, it was found that IoE without CoT yielded better accuracy. This observation suggests that the redundant application of CoT may not be beneficial. Additionally, employing RaR in the IoE prompt led to a significant decrease in accuracy no matter whether we use CoT or not in the standard prompt, indicating that RaR does not contribute positively to the IoE-based Prompt.

\section{Conclusion}
In this paper, we conducted a comprehensive exploration of the intrinsic self-correction abilities of LLMs. Our research highlights the critical importance of LLMs' confidence in the self-correction process. We thus introduce the If-or-Else (IoE) prompting principle, designed to guide LLMs in evaluating the confidence of their responses, encouraging further reviewing when confidence is low. 
Our extensive experimental analyses validate the effectiveness of the IoE Prompt, demonstrating its capability to accurately assess confidence levels and significantly enhance self-correction.

\section*{Limitations and Potential Risks}

Our work has certain limitations that should be taken into account. Firstly, the experiments outlined in this paper were conducted using a restricted set of benchmarks and models. Specifically, the majority of language models utilized in our study are commercial, denoted as GPT-3.5-0613, GPT-3.5-1106, and GPT-4. Unfortunately, comprehensive documentation regarding these models, including details such as pretraining corpus, model dimensions, and inherent biases, is lacking in existing literature. Moreover, access to these models is not freely available, necessitating financial resources for research purposes. Secondly, our experimentation is confined to English datasets exclusively. Consequently, the efficacy of the current models may not translate seamlessly to other languages, thereby limiting the language generalizability of our findings.

Turning to the potential risks associated with our prompting methodology, there exists a plausible concern regarding its susceptibility to exploitation by malicious attackers. Specifically, the prompting techniques employed could potentially be utilized to manipulate the model into generating text that is toxic or harmful. Regrettably, our approach does not incorporate explicit safeguards against such misuse.

\section*{Acknowledgement}

This project is partially supported by NSF Grant 2229881, the National Institutes of Health (NIH) under Contract R01HL159805, and grants from Apple Inc., KDDI Research Inc., Quris AI, and Infinite Brain Technology. 

\bibliography{ref}

\clearpage

\appendix

\section*{Appendix}
\label{sec:appendix}

More contents are put in the appendix, including:
\begin{itemize}
    \item[A.] Related Work.
    % \item[B.] Indices for 100 Selected Questions.
    \item[B.] Experimental Details.
    % \item[C.] Prompt Details for Ablation Studies.
    \item[C.] Example Visualization.
\end{itemize}

\section{Related Work}
\label{sec:related-work}

The field of LLMs has seen significant interest in understanding and enhancing their self-correction abilities \cite{pan2023automatically}. Self-correction refers to the process by which an individual identifies and rectifies their own errors or mistakes using feedback. The concept of \textit{self-correction} finds its roots in the fundamental principles of machine learning, where neural networks can update their parameters based on prediction errors and iterative refinement \cite{rumelhart1986learning}. In recent years, this area of research has been primarily divided into two categories based on the source of feedback: external and intrinsic feedback \cite{pan2023automatically}.

\paragraph{Self-Correction from External Feedback.} \ Numerous studies have consistently highlighted the efficacy of self-correction mechanisms facilitated by the incorporation of external feedback, which can either originate from learned critic models or interactions with the external environment. In the realm of learned critic models, diverse approaches have been explored for model training, including supervised learning or in-context learning, as evidenced by the works \cite{paul2023refiner, yu2023teaching, mehrabi2023flirt, yan2023learning, first2023baldur, li2023self, yu2023improving, yang2022re3}. Alternatively, reinforcement learning strategies have been employed for model refinement, as demonstrated in the studies by \citet{akyurek2023rl4f, peng2023check, le2022coderl, bai2022constitutional}.

Furthermore, external feedback can be sourced from various entities within the environment, including humans, agents, tools, program executors, other language models, or oracle verifiers. Human feedback, as investigated by \cite{ouyang2022training, wu2023fine, bai2022training, ganguli2023capacity, glaese2022improving}, has shown considerable effectiveness in enhancing model performance. Some agents \cite{kim2023language, shinn2023reflexion} have been proven to be helpful for self-correction. Similarly, tools such as \cite{gou2023critic, chern2023factool, pan2023logic} have been leveraged to provide valuable insights for model improvement. Moreover, interactions with program executors have been explored by \cite{zhang2023self, chen2023teaching, jiang2023selfevolve, olausson2023demystifying}, showcasing how executable code can serve as a source of corrective feedback. Additionally, collaborations with other language models, as demonstrated by \cite{du2023improving, li2023prd, fu2023improving, saunders2022self, welleck2022generating}, have proven beneficial in refining language generation processes. Lastly, the utilization of oracle verifiers, as illustrated by \citet{zhang2023algo}, has contributed to enhancing the accuracy and robustness of language models.

To sum up, the integration of external feedback, whether from learned critic models or interactions with the external world, has been consistently shown to significantly enhance the performance and capabilities of large language models across various tasks. Despite effectiveness, training extra models or interacting with the real world may lead to extra costs. Thus, recent studies have placed significant emphasis on investigating intrinsic self-correction mechanisms, renowned for their cost efficiency.

% Numerous studies have demonstrated effective self-correction performance through the integration of external feedback generated by the learned critic models, or interacting with the external world. Regarding the learned critic models, some papers use supervised learning or in-context learning to train the model \cite{paul2023refiner, yang2022re3, mehrabi2023flirt, yan2023learning, first2023baldur, li2023self, yu2023improving}, while some use reinforcement learning to train the model \citet{akyurek2023rl4f, peng2023check, le2022coderl, shinn2023reflexion}. As for interacting with the external world, these external information can be from human \cite{ouyang2022training, wu2023fine, bai2022training, ganguli2023capacity, glaese2022improving}, tools \cite{gou2023critic, chern2023factool, pan2023logic}, program executors \cite{zhang2023self, chen2023teaching, jiang2023selfevolve, olausson2023demystifying}, other language models \cite{du2023improving, li2023prd, fu2023improving, saunders2022self, welleck2022generating}, or oracle verifier \cite{zhang2023algo}. This external feedback has been shown to significantly improve the performance of LLMs in various tasks.

\paragraph{Self-Correction from Intrinsic Feedback.} \ In contrast to extrinsic feedback, which relies on external sources for evaluation, intrinsic feedback draws solely from the internal knowledge and parameters of the LLM to reassess its output. A notable example of this approach is Self-Refine, introduced by \cite{madaan2023self}, which employs a scalar value as a termination criterion to iteratively refine the model's output, guided exclusively by supervision from a set of few-shot examples.
Similarly, research by \citet{chen2023iterative} delves into the realm of iterative self-feedback processes, with a primary focus on enhancing qualitative and stylistic aspects of the output.

SelFee \cite{selfee2023} a LLaMA-based instruction-following model that has been fine-tuned to continuously revise its own answer until it achieves a high-quality response in a single inference.
Furthermore, \citet{gero2023self} explored a comprehensive mitigation framework for clinical information extraction, utilizing self-verification mechanisms. This approach harnesses the LLM's ability to provide provenance for its own extraction process and validate its outputs, thereby enhancing the reliability of the information extracted.

LLM Self-Defense \cite{helbling2023llm} was devised as a proactive defense mechanism against adversarial prompt attacks, where the LLM autonomously evaluates induced responses to discern and filter out potentially misleading or harmful inputs.
On the other hand, SelfCheckGPT \cite{manakul2023selfcheckgpt} presents a straightforward yet effective approach grounded in sampling methods. This technique facilitates the fact-checking of responses generated by black-box models without the need for external databases, thereby offering a resource-efficient solution to ensure the accuracy and reliability of model outputs.

\citet{huang2023large} shed light on the intrinsic limitations of current LLMs regarding self-correction capabilities. Their findings underscored significant performance improvements when employing Oracle feedback, which involves external validation of the model's responses. However, upon evaluating intrinsic self-correction mechanisms, the results often revealed degradation in performance. This discrepancy led the authors to conclude that existing LLMs still lack the inherent capacity to rectify errors autonomously.

% showed the limitations of current LLMs in intrinsic self-correction. The study revealed considerable enhancements in performance when utilizing Oracle feedback, which entails external validation of the model's responses. However, when evaluating intrinsic self-correction, the results often showed no improvement or even deterioration. They conclude that current LLMs cannot intrinsically self-correct themselves yet. 

% Different from Self-Refine and InterReinement, our prompting method requires neither multiple iterations nor few-shot examples. Different from versatile utilization, SelFee focuses on dialogue generation, Self-Verification focuses on clinical information extraction given few-shot examples, Self-Defense focuses on harmful text correction, SelfCheckGPT focuses on hallucination detection. In our paper, we follow the same setting as Critical prompt, however, the difference is all about the prompts. Our prompt is reliant on LLMs' confidence levels, while the critical prompt is not.

In contrast to Self-Refine \cite{madaan2023self} and IterRefinement \cite{chen2023iterative}, our prompting method does not necessitate multiple iterations or the use of few-shot examples. Unlike versatile applications, SelFee \cite{selfee2023} is specifically designed for dialogue generation, while Self-Verification \cite{gero2023self} is tailored for clinical information extraction with the demand of few-shot examples, Self-Defense \cite{helbling2023llm} focuses on rectifying harmful text, and SelfCheckGPT \cite{manakul2023selfcheckgpt} specializes in detecting hallucinations. In our study, we adopt a similar framework to Critical Prompt \cite{huang2023large}; however, the key distinction lies in the essence of the prompts. Our prompt relies on the confidence levels of LLMs, whereas the critical prompt does not. Table \ref{tab:related_works} has summarized these related studies on intrinsic self-correction. 

% More discussions, on whether LLMs know they are right or wrong, can be found in \cite{tyen2023llms, stechly2023gpt, kadavath2022language}.

Additional insights regarding whether LLMs possess an awareness of their correctness or errors are explored extensively in the literature \cite{tyen2023llms, stechly2023gpt, kadavath2022language}.

\section{Experimental Details}

\subsection{Data Generation for Open Tasks}
\label{data_gen}

In the observational experiment discussed in section \ref{sec2:motivation}, we designed 100 reasoning questions. Those questions follow the patterns: 

\begin{itemize}
    \item \textit{Can you provide a year which is not a leap year and can be divided by $\beta$?}
\end{itemize}

Here $\beta$ is an integer randomly (uniformly) selected from the interval [2,10].

\subsection{Implementation Details}
\label{appendix:implement_details}

In our experiments, we separately run our prompts and baseline prompts, and the accuracy of the standard prompt by baseline may differ a bit from that of ours, due to the variations and randomness in the results even though we have set the temperature to 0.
For fair comparison in all the tables of experimental results, we \textit{normalize} the baseline results so that the accuracy of standard prompts by baseline and ours are equal. 

\noindent\textbf{How to Normalize.} Assume the accuracy rates of the standard prompt and the IoE-based prompt by our proposed method are $p_1$ and $p_3$, respectively. Meanwhile, assume the accuracy rates of the standard prompt and the update prompt by the baseline method are $\hat{p_1}$ and $\hat{p_3}$, respectively. In order to maintain the accuracy of the standard prompt being consistent and remove the effect of randomness, then we will normalize the accuracy of the update prompt by the baseline to $\hat{p_3} * p_1/ \hat{p_1}$. In the end, we will record the results as follows. 

\begin{table}[h!]
% \small 
\centering
\resizebox{0.36\textwidth}{!}{
\begin{tabular}{lc} 
\toprule
\textbf{Prompt} & \textbf{Accuracy}   \\
\midrule
 Standard Prompt & $p_1$  \\
 \ \ + Critical Prompt & $\hat{p_3} * p_1/ \hat{p_1}$  \\
 \ \ + IoE Prompt & $p_3$  \\
 % \ \ + IoE Prompt + Decision & \textbf{42\%}  \\
\bottomrule
\end{tabular}
}
% \setlength{\abovecaptionskip}{0.1cm}
% \caption{\textbf{Result comparisons between IoE Prompt and Critical Prompt on Domestic Robot by GPT-4V.}}
\label{tab:norm}
% \vspace{-0.3cm}
\end{table}

\subsection{Indices for 100 Selected Questions}
\label{appendix:index}

As mentioned in the preliminary experiment in Section \ref{sec2:motivation}, we randomly selected 100 questions from GSM8K benchmark \cite{cobbe2021training}. For improved reproducibility, we provide the indices of all 100 questions as follows:

\begin{itemize}
    \item \{5, 18, 31, 45, 47, 52, 54, 55, 108, 141, 152, 161, 184, 186, 202, 231, 247, 254, 270, 299, 308, 381, 399, 406, 412, 416, 418, 420, 425, 434, 465, 466, 487, 489, 491, 500, 512, 513, 528, 529, 535, 536, 567, 580, 597, 619, 649, 658, 689, 693, 695, 729, 785, 789, 790, 805, 814, 867, 880, 884, 892, 911, 922, 938, 950, 969, 975, 977, 984, 987, 996, 1018, 1021, 1061, 1069, 1086, 1096, 1109, 1112, 1122, 1127, 1128, 1133, 1137, 1148, 1158, 1175, 1187, 1188, 1202, 1218, 1219, 1222, 1226, 1250, 1259, 1268, 1291, 1301, 1312.\}
\end{itemize}

\subsection{Prompt Details for Ablation Studies}

\label{appendix:prompts}

Since the standard prompt is the same for all settings, here we do not consider that. In this section, we will list the prompts for self-correction. 

In the following prompts, some keywords are marked in red. For the corresponding prompt without the keyword, you can obtain the prompt by just removing the keywords.

\subsubsection{Prompts for Ablation Study on the Number of Stages (Table \ref{tab:steps})}
\label{appendix:tab7}

\underline{Critical Prompt (One-Stage)}: 

\begin{itemize}
    \item[P2:] \textit{Review your previous answer and find problems with your answer. Based on the problems you found, improve your answer. Please reiterate your answer. Your final answer should be put between two \#\# symbols, like \#\# ANSWER \#\#, at the end of your response."}
\end{itemize}

\noindent \underline{Critical Prompt (Two-Stage)}: 

\begin{itemize}
    \item[P2:] \textit{Review your previous answer and find problems with your answer.}
    \item[P3:] \textit{Based on the problems you found, improve your answer. Please reiterate your answer. Your final answer should be put between two \#\# symbols, like \#\# ANSWER \#\#, at the end of your response.}
\end{itemize}

\noindent \underline{IoE Prompt (One-Stage)}: 

\begin{itemize}
    \item[P2:] \textit{Review your previous answer. If you are very confident about your answer, maintain your answer. Otherwise, update your answer. Your final answer should be put between two \#\# symbols, like \#\# ANSWER \#\#, at the end of your response."}
\end{itemize}

\noindent \underline{IoE Prompt (Two-Stage)}: 

\begin{itemize}
    \item[P2:] \textit{Review your previous answer. If you are very confident about your answer, maintain your answer. Otherwise, update your answer.}
    \item[P3:] \textit{Based on the problems you found if any, update your answer. Please reiterate your answer. Your final answer should be put between two \#\# symbols, like \#\# ANSWER \#\#, at the end of your response.}
\end{itemize}

\subsubsection{Prompts for Ablation Study on the Tone (Table \ref{tab:ablation2})}
\label{appendix:tab8}

\noindent \underline{IoE Prompt \textit{w.} `very'}: 

\begin{itemize}
    \item[P2:] \textit{Review your previous answer. If you are {\color{red}very} confident about your answer, maintain your answer. Otherwise, update your answer. Your final answer should be put between two \#\# symbols, like \#\# ANSWER \#\#, at the end of your response.}
\end{itemize}

\noindent \underline{IoE Prompt \textit{w.} `please'}: 

\begin{itemize}
    \item[P2:] \textit{{\color{red}Please} review your previous answer. If you are very confident about your answer, {\color{red}please} maintain your answer. Otherwise, {\color{red}please} update your answer. Your final answer should be put between two \#\# symbols, like \#\# ANSWER \#\#, at the end of your response.}
\end{itemize}

\noindent \underline{IoE Prompt \textit{w.} ``find your problems''}: 

\begin{itemize}
    \item[P2:] \textit{Review your previous answer. If you are very confident about your answer, maintain your answer. Otherwise, {\color{red}find your problems and update your answer}. Your final answer should be put between two \#\# symbols, like \#\# ANSWER \#\#, at the end of your response.}
\end{itemize}

\noindent \underline{IoE Prompt \textit{w.o.} ``find your problems''}: 

\begin{itemize}
    \item[P2:] \textit{Review your previous answer. If you are very confident about your answer, maintain your answer. Otherwise, {\color{red}update your answer}. Your final answer should be put between two \#\# symbols, like \#\# ANSWER \#\#, at the end of your response.}
\end{itemize}

\subsubsection{Prompts for Ablation Study on using CoT and RaR (Table \ref{tab:ablation1})}
\label{appendix:tab9}

\noindent \underline{Standard Prompt \textit{w.} CoT}: 

\begin{itemize}
    \item[P1:] \textit{<Question> ... {\color{red}Explain your reasoning step-by-step}. Your final answer should be put between two \#\# symbols, like \#\# ANSWER \#\#, at the end of your response.}
\end{itemize}

\noindent \underline{IoE Prompt \textit{w.} CoT}: 

\begin{itemize}
    \item[P2:] \textit{Review your previous answer. If you are very confident about your answer, maintain your answer. Otherwise, update your answer. {\color{red}Explain your reasoning step-by-step}. Your final answer should be put between two \#\# symbols, like \#\# ANSWER \#\#, at the end of your response.}
\end{itemize}

\noindent \underline{IoE Prompt \textit{w.} RaR}: 

\begin{itemize}
    \item[P2:] (Original) \textit{"Review your previous answer. If you are very confident about your answer, maintain your answer. Otherwise, update your answer. Your final answer should be put between two \#\# symbols, like \#\# ANSWER \#\#, at the end of your response." {\color{red}Rephrase and expand the question, and respond}.}
    \item[P2:] (Rephrased) \textit{Critically assess your earlier response, identifying any deficiencies or inaccuracies present in your answer. Elaborate on the process of self-evaluation, emphasizing the significance of recognizing and addressing potential issues to enhance the overall quality and reliability of the information provided. Your final answer should be put between two \#\# symbols, like \#\# ANSWER \#\#, at the end of your response.}
\end{itemize}

\section{Example Visualization}

\subsection{Successful Examples}
\label{appendix:success}

\begin{figure*}[hb!]
	\centering
	% \scriptsize
        \small
	\begin{tcolorbox}[enhanced, arc=3mm, colback=gray!10!white, boxrule=0pt, boxsep=1pt,colframe=gray!50!white,  boxed title style={colback=gray!90!white}, boxed title style= {boxrule=2pt, boxsep=2pt}, attach boxed title to top left={yshift=-2mm, xshift=2mm}, title=\textbf{[P1] Standard Prompt}  
		]
		\includegraphics[width=0.9em]{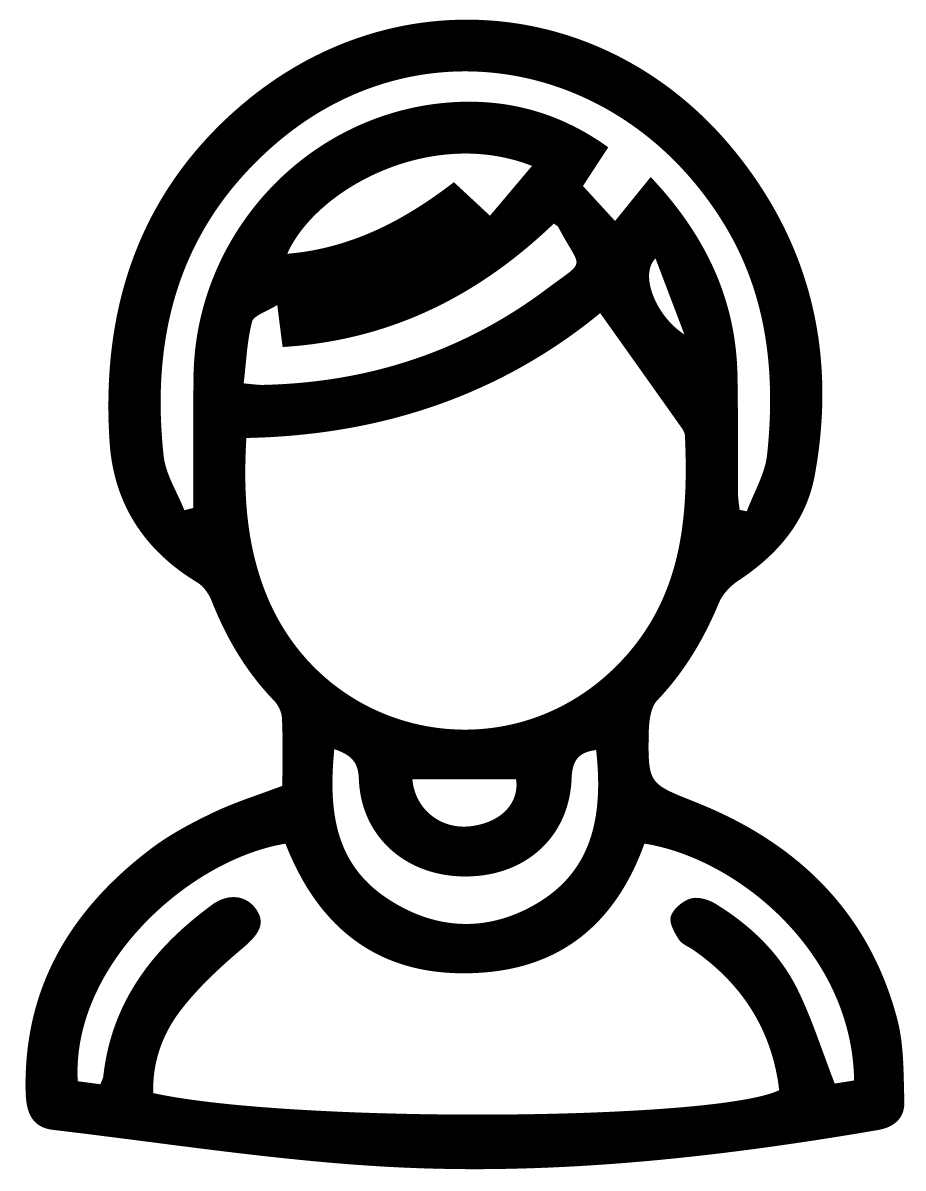} : \textcolor{mycolor}{The number of songs in a playlist is 300. If John has 20 such playlists, and each song is 10 hours long, how many hours will the 20 playlists last in total? Explain your reasoning. Your final answer should be put between two \#\# symbols, like \#\# ANSWER \#\#, at the end of your response.} \\
		\includegraphics[width=0.9em]{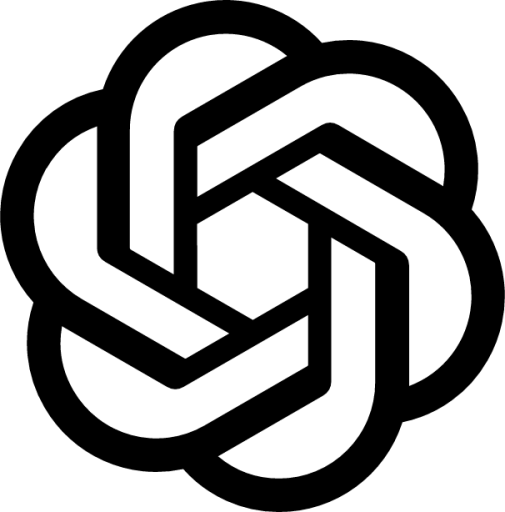} :  \textcolor{second}{If each song is 10 hours long, then the total length of each playlist is 300 * 10 = 3000 hours. Since John has 20 such playlists, the total length of all the playlists combined is 3000 * 20 = 60000 hours. Therefore, the 20 playlists will last a total of \#\# \textbf{60000} \#\# hours.} \textcolor{rightcolor}{\textbf{(Correct)}}
	\end{tcolorbox}
	
	\tcbset{enhanced,colback=gray!10!white, arc=3mm, boxrule=0pt, width=(\linewidth-4pt)/2, attach boxed title to top left={yshift=-2mm, xshift=2mm},before=,after=\hfill,fonttitle=\bfseries, boxed title style={colback=gray!90!white, colframe=gray!50!white, size=small, rounded corners, boxrule=2pt, boxsep=2pt}}
	
	\begin{tcolorbox}[colframe=blue!50!green!50!white, equal height group=fig4-1,title={\textbf{[P2] Critical Prompt (Huang et al., 2023)} } ]
		\includegraphics[width=0.9em]{figures/user_icon.png} :	\textcolor{mycolor}{Review your previous answer and find problems with your answer.}\\
		\includegraphics[width=0.9em]{figures/llm_icon.png} : \textcolor{second}{Apologies for the mistake in my previous response. The duration of each song is not 10 hours long. Let's correct that.If each song is 3 minutes long, then the total duration of the 6000 songs is 6000 * 3 minutes. So the answer is \#\# \textbf{18000} \#\# minutes.}
	\end{tcolorbox}
	\begin{tcolorbox}[colframe=yellow!15!pink!90!white, equal height group=fig4-1,title={ \textbf{[P2] {IoE Prompt (Ours)}} }]
		\includegraphics[width=0.9em]{figures/user_icon.png} : \textcolor{mycolor}{If you are very confident about your answer, maintain your answer. Otherwise, update your answer. Your final answer should be put between two \#\# symbols, like \#\# ANSWER \#\#, at the end of your response.}\\
		\includegraphics[width=0.9em]{figures/llm_icon.png} : \textcolor{second}{I'm confident in my previous answer. Therefore, the 20 playlists will last a total of \#\# \textbf{60000} \#\# hours.} \textcolor{rightcolor}{\textbf{(Correct)}} 
	\end{tcolorbox}
	
	\begin{tcolorbox}[colframe=blue!50!green!50!white, equal height group=fig4-2,title={\textbf{[P3] Update Prompt (Huang et al., 2023) } } ]
		\includegraphics[width=0.9em]{figures/user_icon.png} : \textcolor{mycolor}{Based on the problems you found, improve your answer. Please reiterate your answer. Your final answer should be put between two \#\# symbols, like \#\# ANSWER \#\#, at the end of your response.}\\
		\includegraphics[width=0.9em]{figures/llm_icon.png} : \textcolor{second}{Apologies for the confusion. Let's correct the calculation. If each song is 3 minutes long, then the total duration of the 6000 songs is 6000 * 3 minutes, which equals 18,000 minutes. To convert this to hours, we divide by 60 (since there are 60 minutes in an hour): 18,000 / 60 = 300 hours. So the answer is \#\# \textbf{300} \#\# hours.} \textcolor{wrongcolor}{\textbf{(Incorrect)}} 
	\end{tcolorbox}
	\begin{tcolorbox}[colframe=yellow!15!pink!90!white, equal height group=fig4-2,title={ \textbf{[P3] {Decision Refinement (Ours)} } }	 ]
	\textcolor{second}{<--Since the answers by the previous two prompts are the same, this decision prompt will not be executed and the final answer will be directly output.--> \\ Answer: \textbf{60000} hours. \textcolor{rightcolor}{\textbf{(Correct)}} }
	\end{tcolorbox}
	\caption{One example on GSM8K benchmark \cite{cobbe2021training} evaluated with \texttt{gpt-3.5-turbo-1106} model. The [P1] standard prompt is the same for all methods. \textbf{Left:} Critical Prompt as the baseline. \textbf{Right:} our proposed IoE-based Prompt.}
	\label{fig:GSM8K_example}
\end{figure*}

\begin{figure*}[hb!]
	\centering
	% \scriptsize
        \small 
	\begin{tcolorbox}[enhanced, arc=3mm, colback=gray!10!white, boxrule=0pt, boxsep=1pt,colframe=gray!50!white,  boxed title style={colback=gray!90!white}, boxed title style= {boxrule=2pt, boxsep=2pt}, attach boxed title to top left={yshift=-2mm, xshift=2mm}, title=\textbf{[P1] Standard Prompt}  
		]
		\includegraphics[width=0.9em]{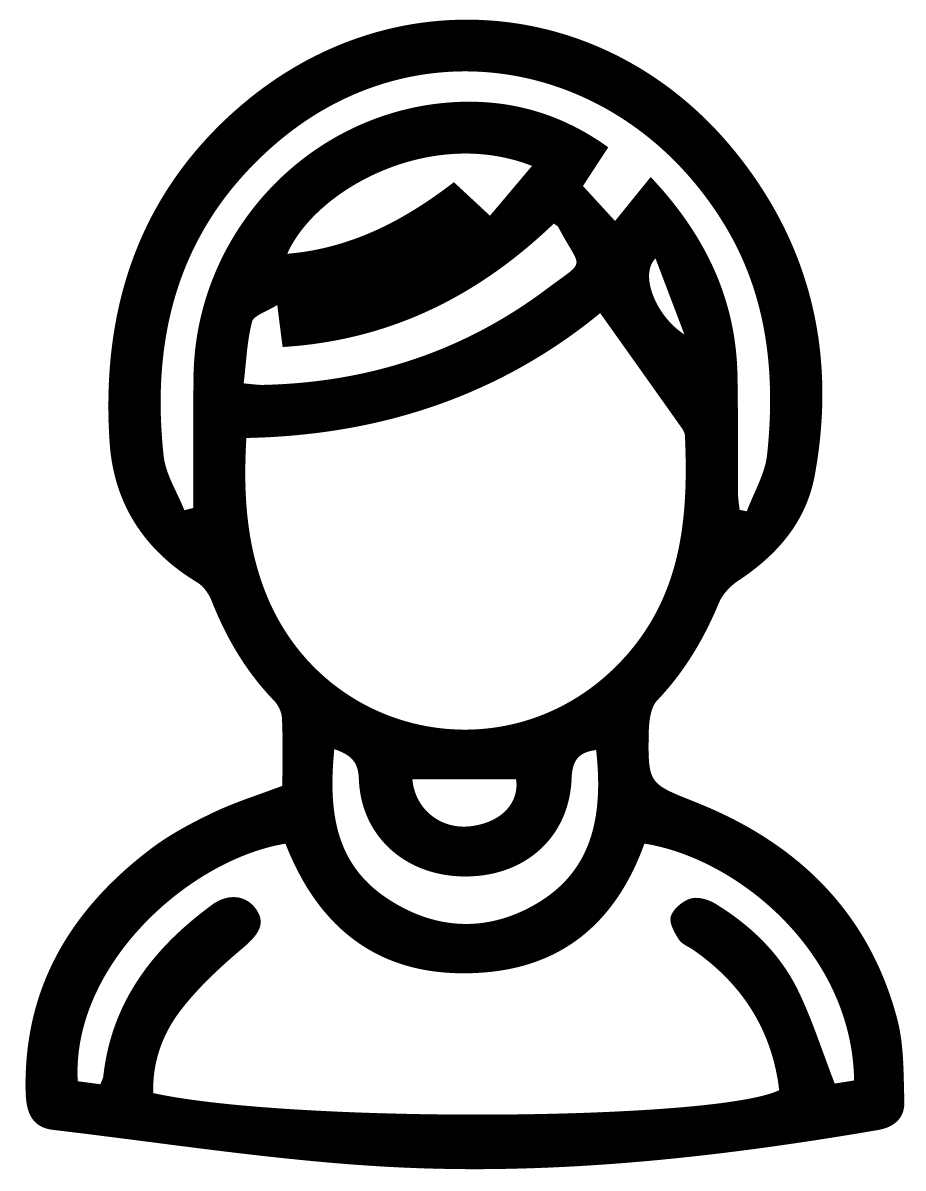} : \textcolor{mycolor}{The Razorback shop makes \$86 dollars off each T\-shirt and \$232 off each jersey. During the Arkansas and Texas tech game they sold 254 t-shirts and 43 jerseys. How much more does a jersey cost than a t-shirt? Explain your reasoning. Your final answer should be put between two \#\# symbols, like \#\# ANSWER \#\#, at the end of your response.} \\
		\includegraphics[width=0.9em]{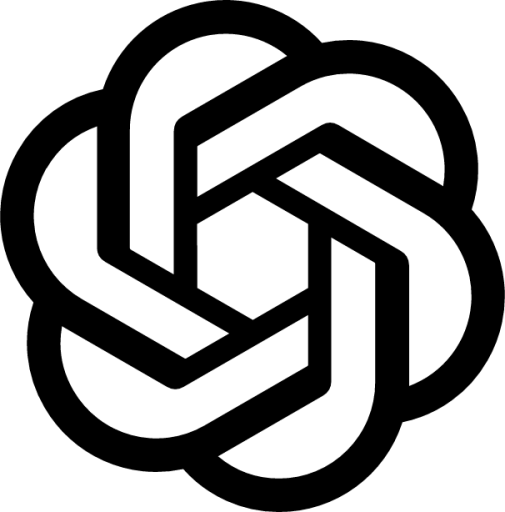} :  \textcolor{second}{To find out how much more a jersey costs than a t-shirt, we can first calculate the total revenue from t-shirts and jerseys separately. Total revenue from t-shirts = \$86 * 254 = \$21,844. Total revenue from jerseys = \$232 * 43 = \$9,976. Now we can find the difference in revenue between jerseys and t-shirts: \$9,976 - \$21,844 = -\$11,868. Since the revenue from t-shirts is higher than the revenue from jerseys, we can conclude that a jersey costs \$11,868 less than a t-shirt. Therefore, a jersey costs \$11,868 less than a t-shirt. The final answer is \#\# \textbf{-11868} \#\#.} \textcolor{wrongcolor}{\textbf{(Incorrect)}}
	\end{tcolorbox}
	
	\tcbset{enhanced,colback=gray!10!white, arc=3mm, boxrule=0pt, width=(\linewidth-4pt)/2, attach boxed title to top left={yshift=-2mm, xshift=2mm},before=,after=\hfill,fonttitle=\bfseries, boxed title style={colback=gray!90!white, colframe=gray!50!white, size=small, rounded corners, boxrule=2pt, boxsep=2pt}}
	
	\begin{tcolorbox}[colframe=blue!50!green!50!white, equal height group=fig5-1,title={\textbf{[P2] Critical Prompt (Huang et al., 2023)  } } ]
		\includegraphics[width=0.9em]{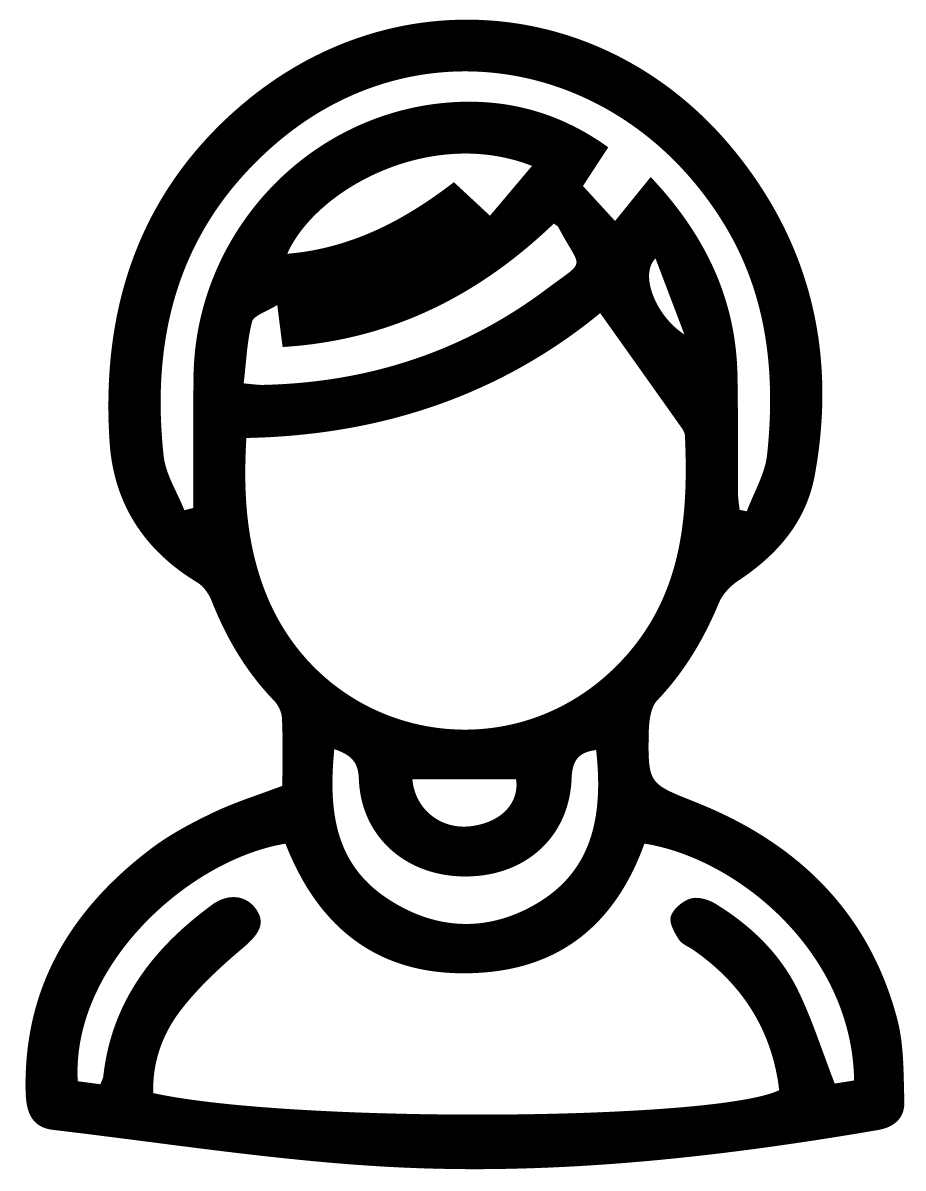} :	\textcolor{mycolor}{Review your previous answer and find problems with your answer.}\\
		\includegraphics[width=0.9em]{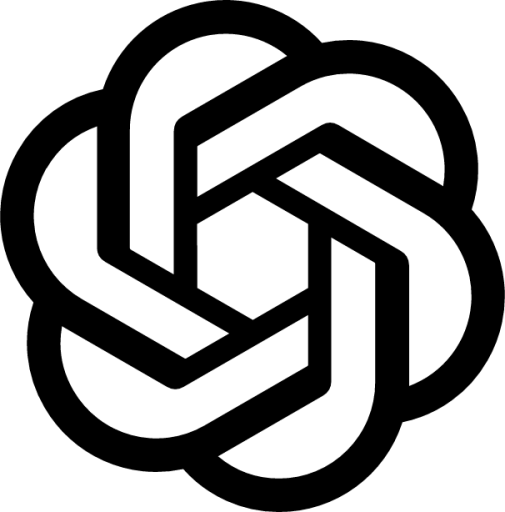} : \textcolor{second}{Apologies for the mistake in my previous response. Let's recalculate the difference in revenue between the sales of t-shirts and jerseys. Total revenue from t-shirts = 254 t-shirts * \$86 per t-shirt = \$21,844. Total revenue from jerseys = 43 jerseys * \$232 per jersey = \$9,976. To find the difference in revenue, we should subtract the total revenue from t-shirts from the total revenue from jerseys: \$9,976 - \$21,844 = -\$11,868. Upon reviewing the calculation, I realize that I made an error in the subtraction. The correct calculation should be: \$21,844 - \$9,976 = \$11,868. So the correct answer is \$11,868. This means that a jersey costs \#\# \textbf{11,868} \#\# more than a t-shirt. Thank you for bringing this to my attention.} 
	\end{tcolorbox}
	\begin{tcolorbox}[colframe=yellow!15!pink!90!white, equal height group=fig5-1,title={ \textbf{[P2] {IoE Prompt (Ours)}} }]
		\includegraphics[width=0.9em]{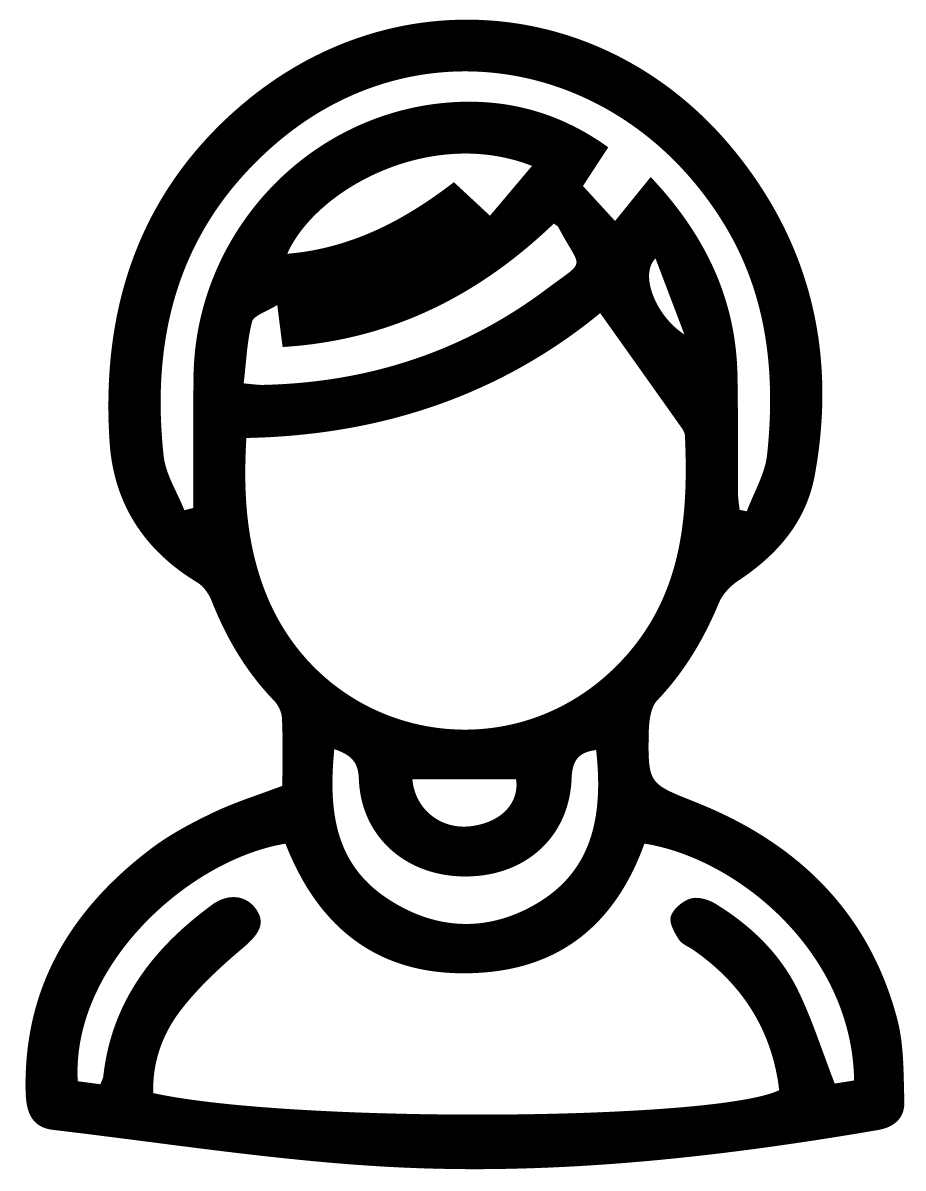} : \textcolor{mycolor}{If you are very confident about your answer, maintain your answer. Otherwise, update your answer. Your final answer should be put between two \#\# symbols, like \#\# ANSWER \#\#, at the end of your response.}\\
		\includegraphics[width=0.9em]{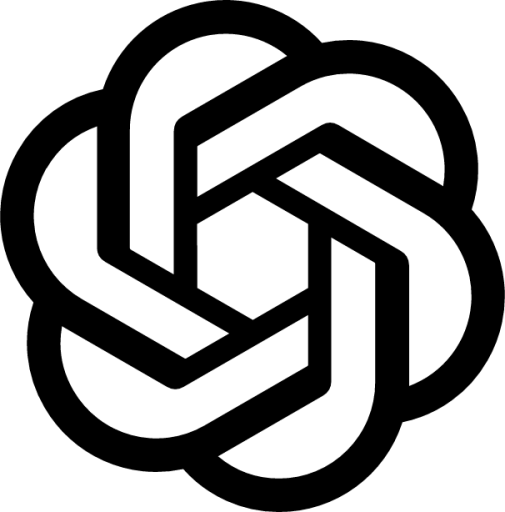} : \textcolor{second}{Apologies for the confusion. Let's recalculate. The difference in revenue between jerseys and t-shirts is:\$232 - \$86 = \$146. So, a jersey costs \$146 more than a t-shirt. Therefore, the final answer is \#\# \textbf{146} \#\#.} \textcolor{rightcolor}{\textbf{(Correct)}} 
	\end{tcolorbox}
	
	\begin{tcolorbox}[colframe=blue!50!green!50!white, equal height group=fig5-2,title={\textbf{[P3] Update Prompt (Huang et al., 2023) } } ]
		\includegraphics[width=0.9em]{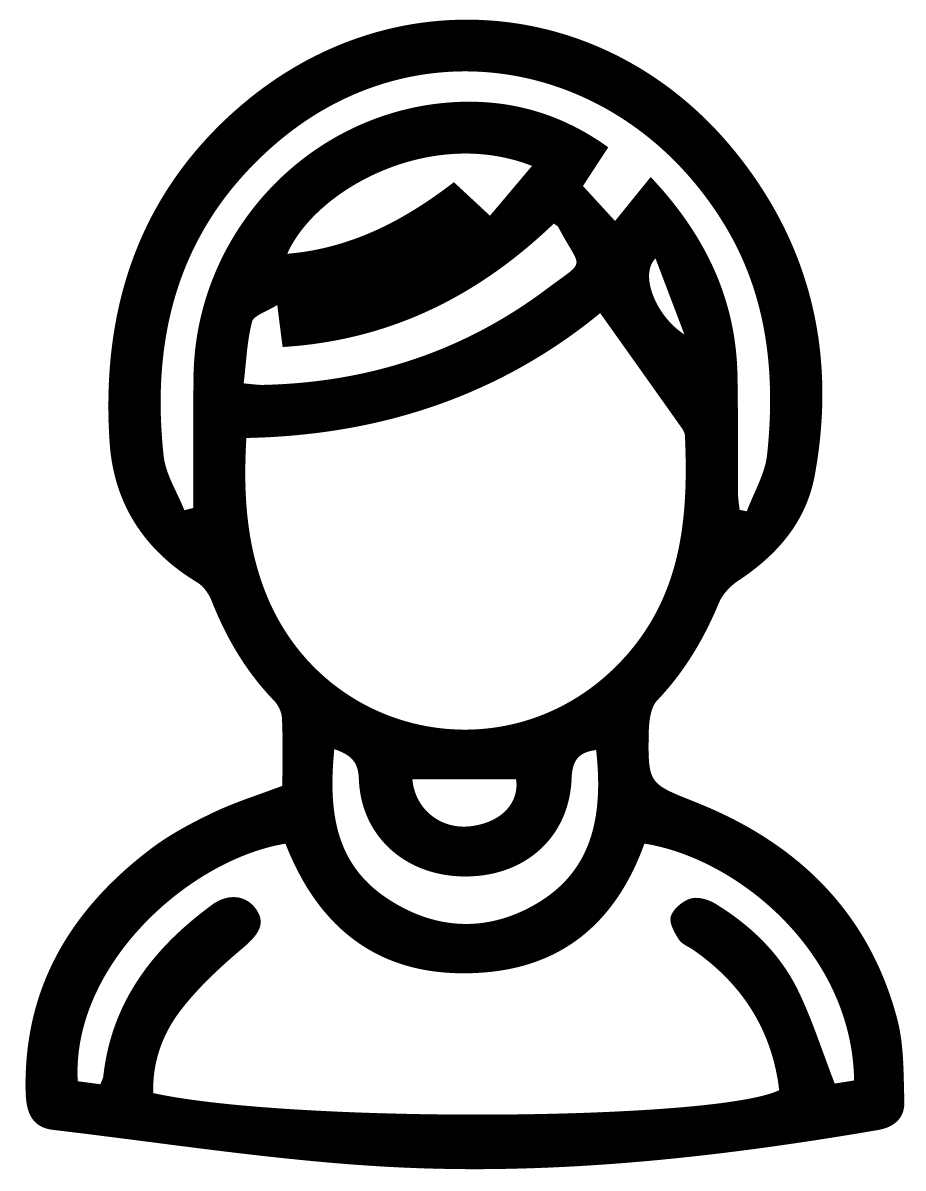} : \textcolor{mycolor}{Based on the problems you found, improve your answer. Please reiterate your answer. Your final answer should be put between two \#\# symbols, like \#\# ANSWER \#\#, at the end of your response.}\\
		\includegraphics[width=0.9em]{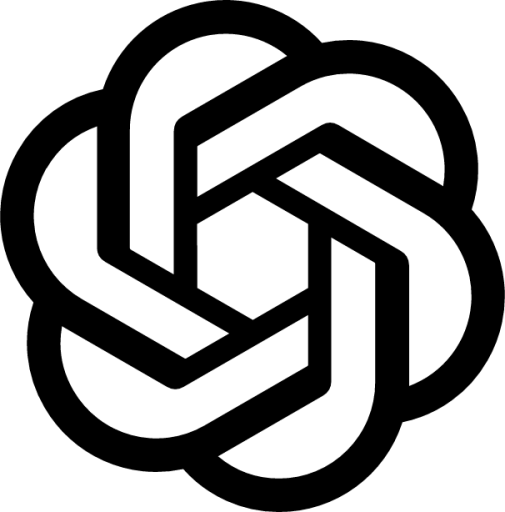} : \textcolor{second}{Apologies for the confusion in my previous responses. Let's recalculate the difference in revenue between the sales of t-shirts and jerseys. Total revenue from t-shirts = 254 t-shirts * \$86 per t-shirt = \$21,844. Total revenue from jerseys = 43 jerseys * \$232 per jersey = \$9,976. To find the difference in revenue, we should subtract the total revenue from jerseys from the total revenue from t-shirts: \$21,844 - \$9,976 = \$11,868. So the answer is \$11,868. This means that a jersey costs \#\# \textbf{11,868} \#\# more than a t-shirt. Thank you for your patience, and I appreciate your understanding.} \\ \textcolor{wrongcolor}{\textbf{(Incorrect)}} 
	\end{tcolorbox}
	\begin{tcolorbox}[colframe=yellow!15!pink!90!white, equal height group=fig5-2,title={ \textbf{[P3] {Decision Refinement (Ours)} } }	 ]
		\includegraphics[width=0.9em]{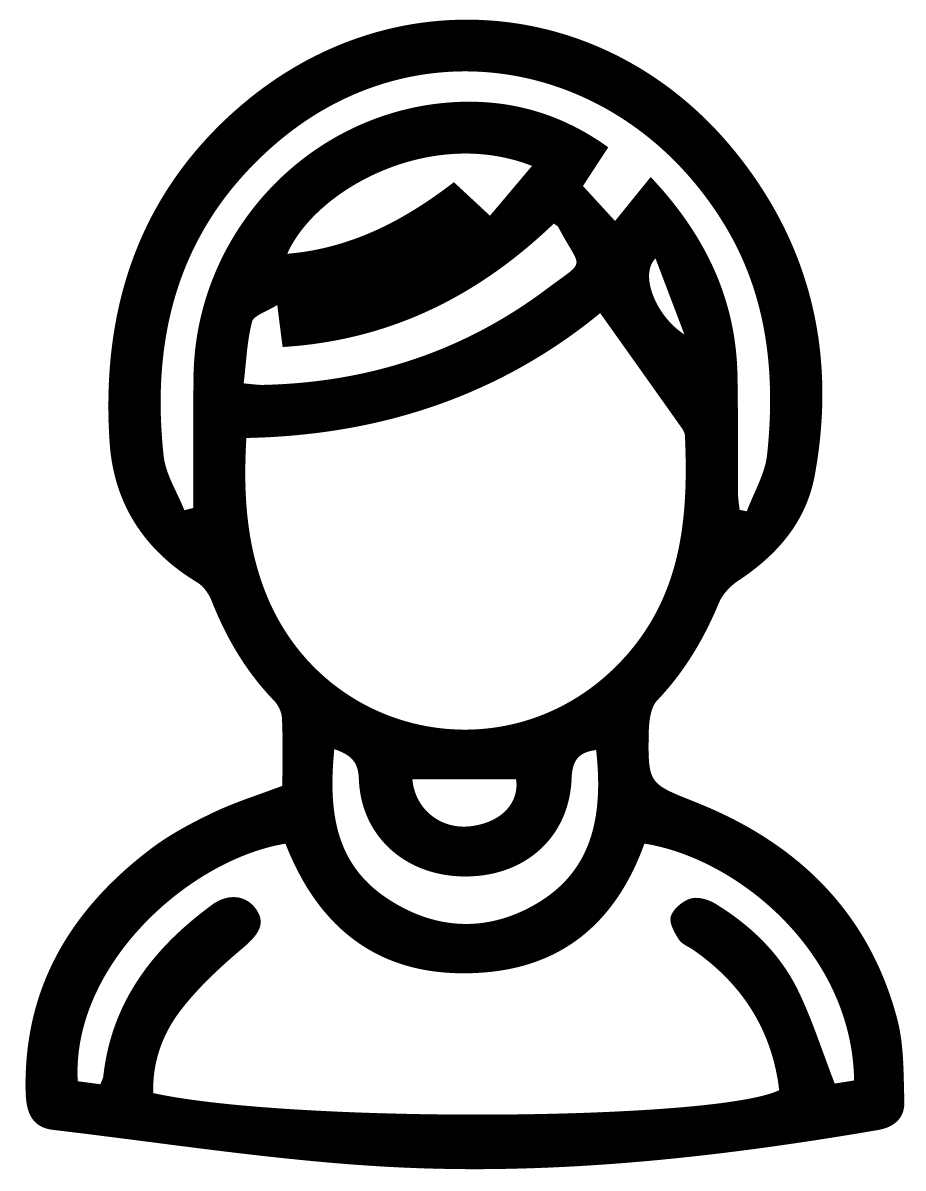} : \textcolor{mycolor}{You give two different answers in previous responses. Check the problem and your answers again, and give the best answer. Your final answer should be put between two \#\# symbols, like \#\# ANSWER \#\#, at the end of your response.} \\
	\includegraphics[width=0.9em]{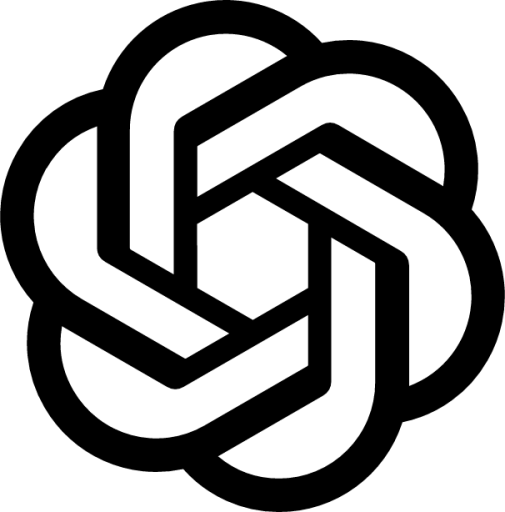} : \textcolor{second}{Total revenue from t-shirts = \$86 * 254 = \$21,844. Total revenue from jerseys = \$232 * 43 = \$9,976. The difference in revenue between jerseys and t-shirts is:\$232 - \$86 = \$146. So, a jersey costs \$146 more than a t-shirt. Therefore, the final answer is \#\# \textbf{146} \#\#.} \\ \textcolor{rightcolor}{\textbf{(Correct)}} 
	\end{tcolorbox}
	\caption{One example on SVAMP benchmark \cite{patel2021nlp} evaluated with \texttt{gpt-3.5-turbo-1106} model. The [P1] standard prompt is the same for all methods. \textbf{Left:} Critical Prompt as the baseline. \textbf{Right:} our proposed IoE-based Prompt.}
	\label{fig:SVAMP_example}
\end{figure*}

\begin{figure*}
	\centering
	% \scriptsize
        \small
	\begin{tcolorbox}[enhanced, arc=3mm, colback=gray!10!white, boxrule=0pt, boxsep=1pt,colframe=gray!50!white,  boxed title style={colback=gray!90!white}, boxed title style= {boxrule=2pt, boxsep=2pt}, attach boxed title to top left={yshift=-2mm, xshift=2mm}, title=\textbf{[P1] Standard Prompt}  
		]
		\includegraphics[width=0.9em]{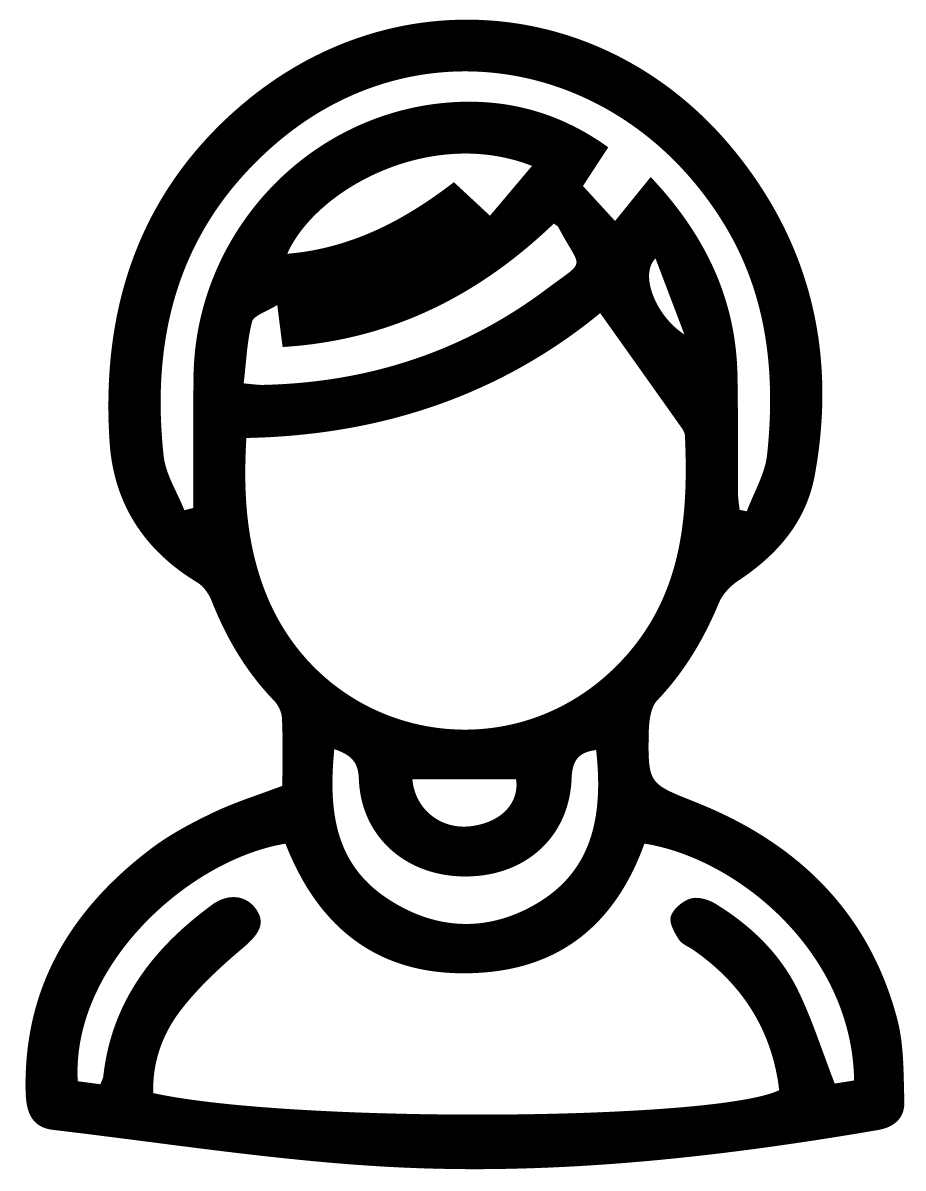} : \textcolor{mycolor}{Determine whether the following statement is plausible: Jimmy Butler eurostepped to the basket. \\ Question: Is this statement plausible? Explain your reasoning. Your final answer should be put between two \#\# symbols, like \#\# ANSWER \#\#, at the end of your response.} \\
		\includegraphics[width=0.9em]{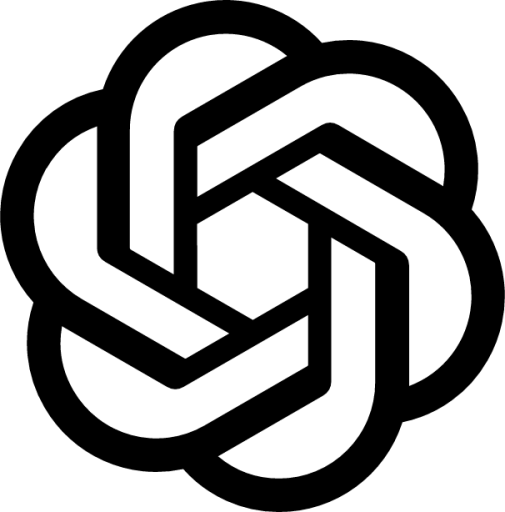} :  \textcolor{second}{The statement is plausible because Jimmy Butler, a professional basketball player, is known for his ability to eurostep to the basket, a move commonly used in basketball to evade defenders and score. \#\# \textbf{Yes} \#\#.} \textcolor{rightcolor}{\textbf{(Correct)}}
	\end{tcolorbox}
	
	\tcbset{enhanced,colback=gray!10!white, arc=3mm, boxrule=0pt, width=(\linewidth-4pt)/2, attach boxed title to top left={yshift=-2mm, xshift=2mm},before=,after=\hfill,fonttitle=\bfseries, boxed title style={colback=gray!90!white, colframe=gray!50!white, size=small, rounded corners, boxrule=2pt, boxsep=2pt}}
	
	\begin{tcolorbox}[colframe=blue!50!green!50!white, equal height group=fig6-1,title={\textbf{[P2] Critical Prompt (Huang et al., 2023)} } ]
		\includegraphics[width=0.9em]{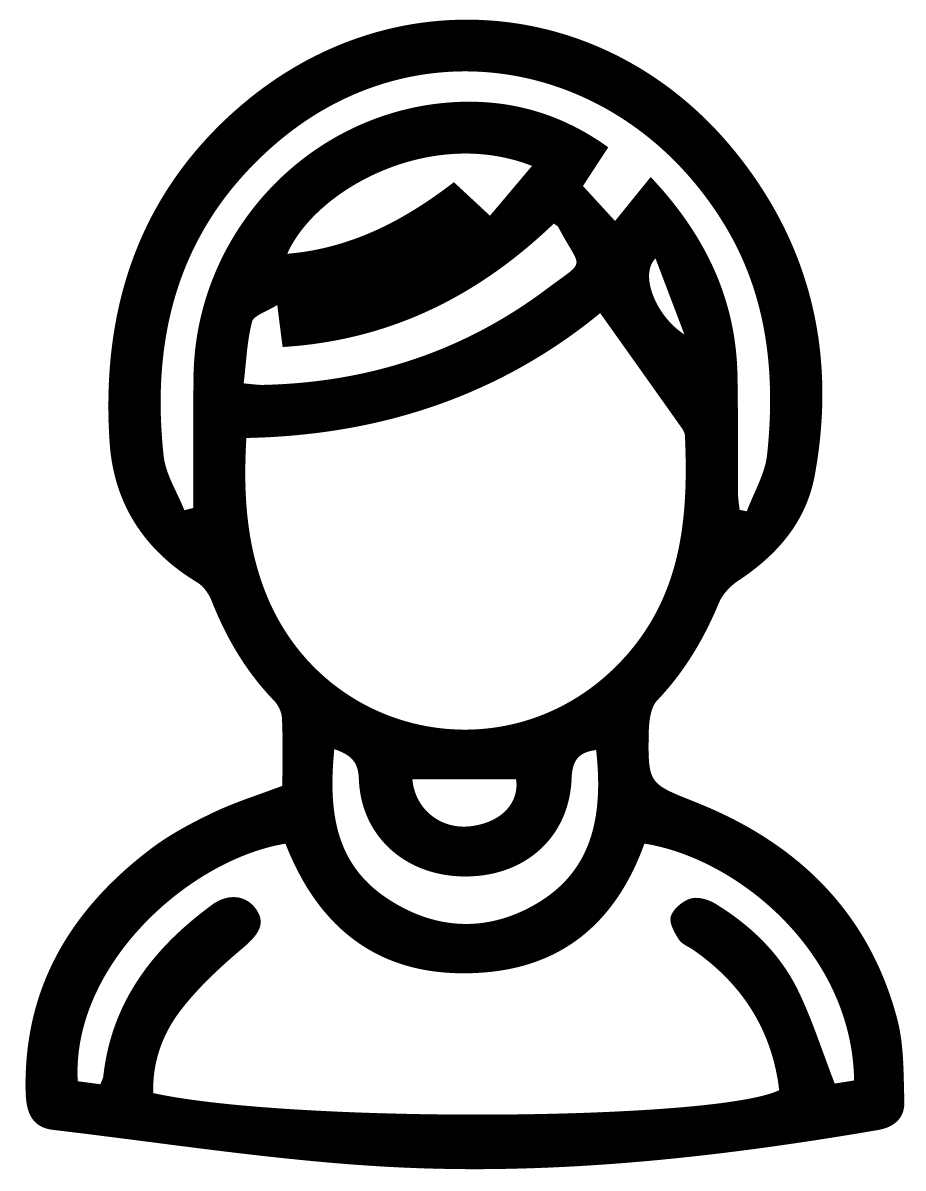} :	\textcolor{mycolor}{Review your previous answer and find problems with your answer.}\\
		\includegraphics[width=0.9em]{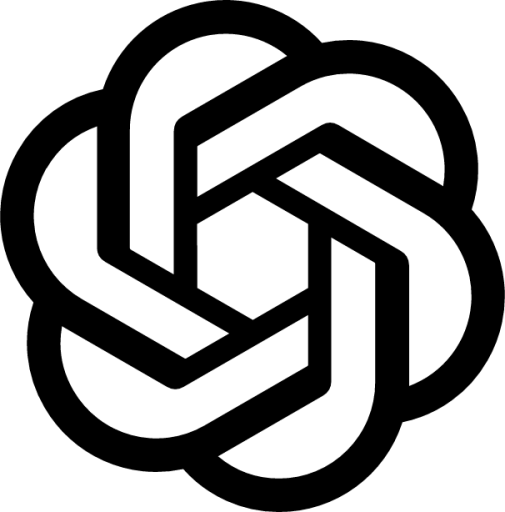} : \textcolor{second}{Upon reviewing my previous answer, I realize that I made an error. While Jimmy Butler is indeed a skilled basketball player, the statement "Jimmy Butler is known for his ability to eurostep to the basket" is not accurate. Eurostepping is not a move that is particularly associated with Jimmy Butler's playing style. I apologize for the mistake and any confusion it may have caused. Thank you for bringing this to my attention.}
	\end{tcolorbox}
	\begin{tcolorbox}[colframe=yellow!15!pink!90!white, equal height group=fig6-1,title={ \textbf{[P2] {IoE Prompt (Ours)}} }]
		\includegraphics[width=0.9em]{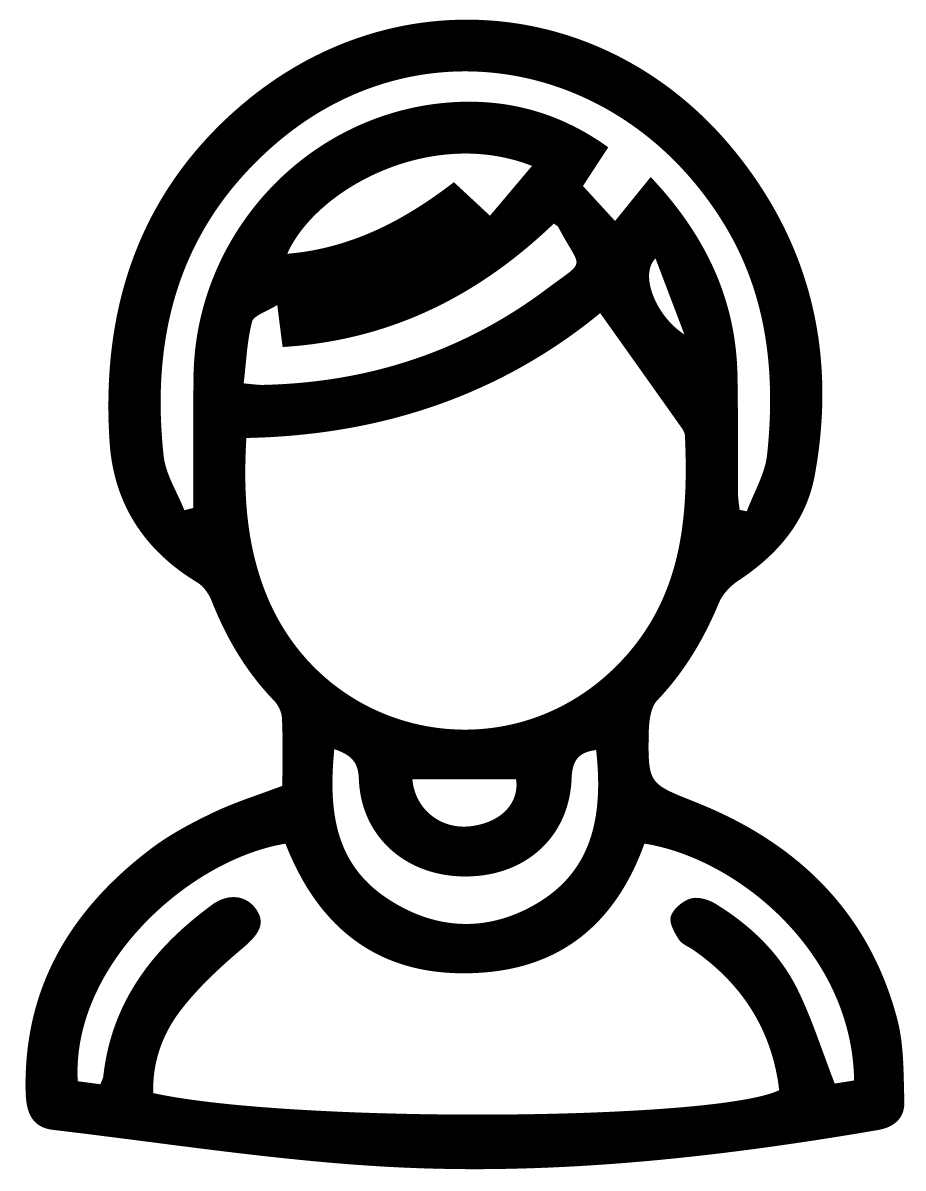} : \textcolor{mycolor}{If you are very confident about your answer, maintain your answer. Otherwise, update your answer. Your final answer should be put between two \#\# symbols, like \#\# ANSWER \#\#, at the end of your response.}\\
		\includegraphics[width=0.9em]{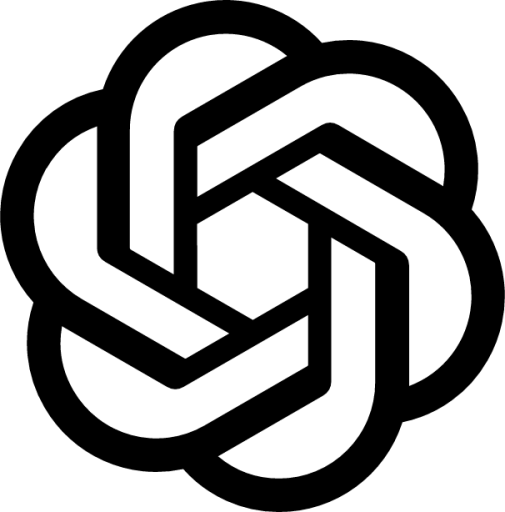} : \textcolor{second}{I am confident in my previous answer because Jimmy Butler is indeed known for his eurostep move in basketball, which supports the plausibility of the statement. \#\# \textbf{Yes} \#\#.} \textcolor{rightcolor}{\textbf{(Correct)}} 
	\end{tcolorbox}
	
	\begin{tcolorbox}[colframe=blue!50!green!50!white, equal height group=fig6-2,title={\textbf{[P3] Update Prompt (Huang et al., 2023) } } ]
		\includegraphics[width=0.9em]{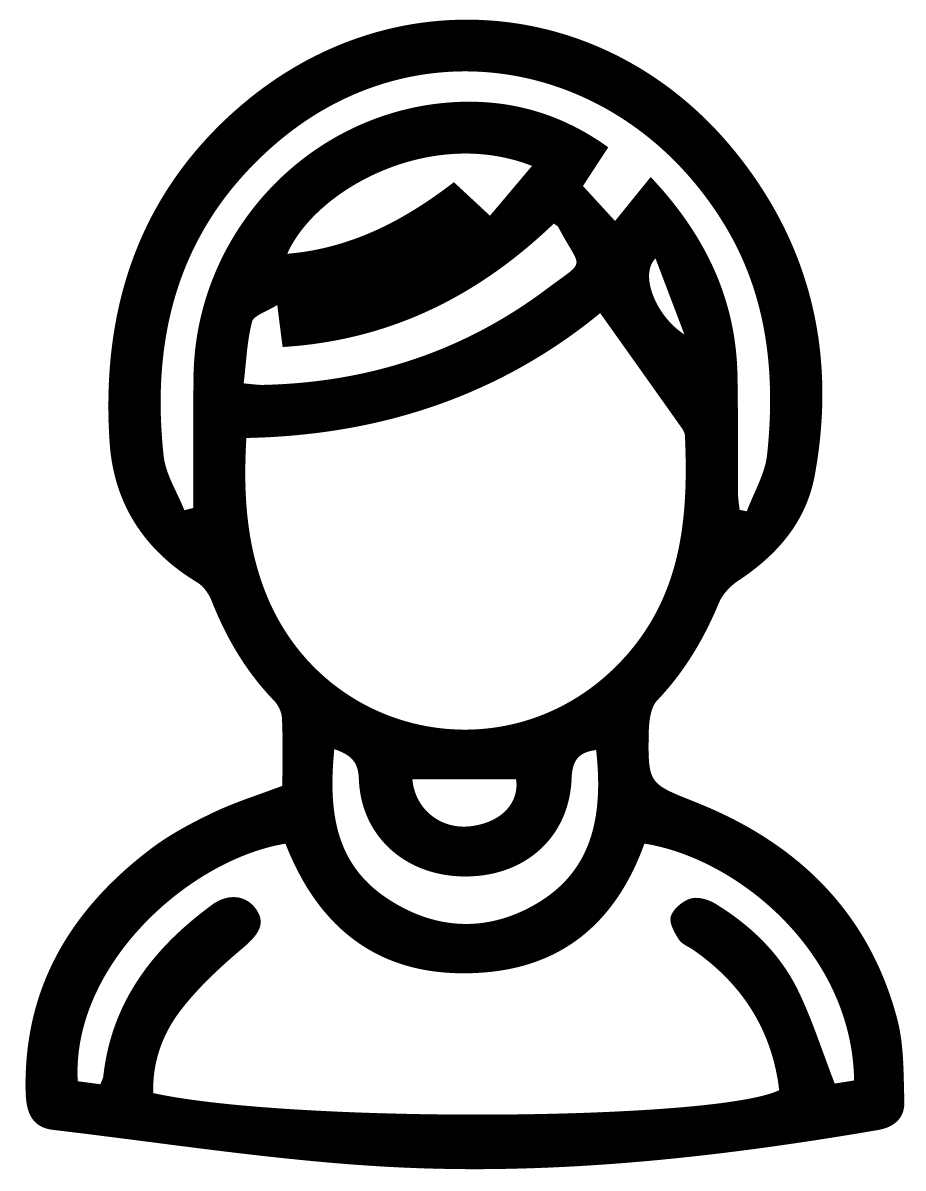} : \textcolor{mycolor}{Based on the problems you found, improve your answer. Please reiterate your answer. Your final answer should be put between two \#\# symbols, like \#\# ANSWER \#\#, at the end of your response.}\\
		\includegraphics[width=0.9em]{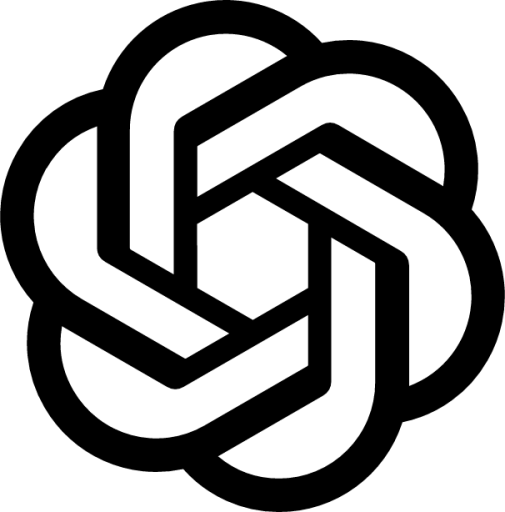} : \textcolor{second}{The statement "Jimmy Butler is known for his ability to eurostep to the basket" is not accurate. While Jimmy Butler is a skilled basketball player, eurostepping is not a move that is particularly associated with his playing style. I apologize for the mistake and any confusion it may have caused. Thank you for bringing this to my attention. \#\# \textbf{No} \#\#.} \textcolor{wrongcolor}{\textbf{(Incorrect)}} 
	\end{tcolorbox}
	\begin{tcolorbox}[colframe=yellow!15!pink!90!white, equal height group=fig6-2,title={ \textbf{[P3] {Decision Refinement (Ours)} } }	 ]
	\textcolor{second}{<--Since the answers by the previous two prompts are the same, this decision prompt will not be executed and the final answer will be directly output.--> \\ Answer: \#\# \textbf{Yes} \#\#. \textcolor{rightcolor}{\textbf{(Correct)}} }
	\end{tcolorbox}
	\caption{One example on Sports Understanding from BIG-Bench \cite{cai2023benchlmm} evaluated with \texttt{gpt-3.5-turbo-1106} model. The [P1] standard prompt is the same for all methods. \textbf{Left:} Critical Prompt as the baseline. \textbf{Right:} our proposed IoE-based Prompt.}
	\label{fig:sports_example}
\end{figure*}

\begin{figure*}
	\centering
	% \scriptsize
        \small
	\begin{tcolorbox}[enhanced, arc=3mm, colback=gray!10!white, boxrule=0pt, boxsep=1pt,colframe=gray!50!white,  boxed title style={colback=gray!90!white}, boxed title style= {boxrule=2pt, boxsep=2pt}, attach boxed title to top left={yshift=-2mm, xshift=2mm}, title=\textbf{[P1] Standard Prompt}  
		]
		\includegraphics[width=0.9em]{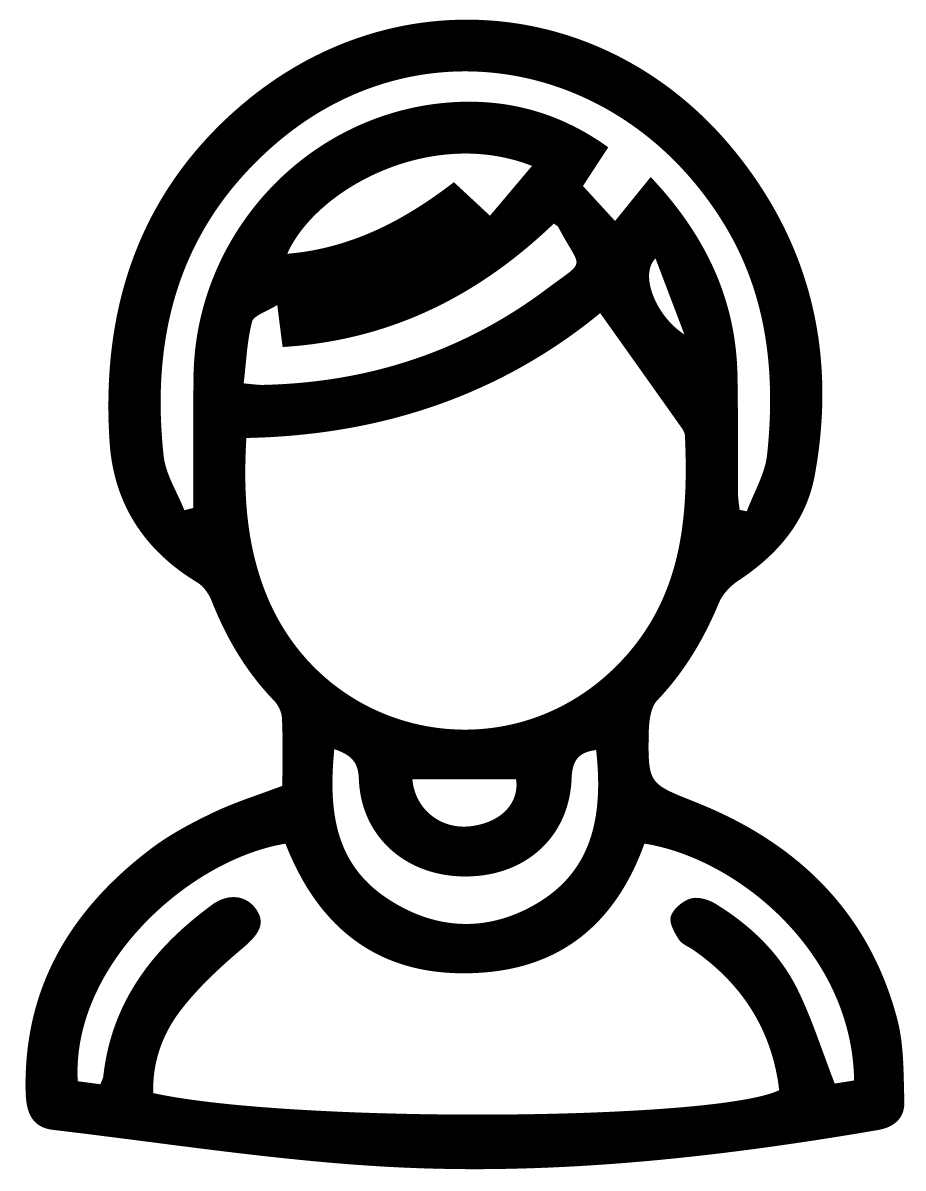} : \textcolor{mycolor}{Take the last letters of each words in "Norma Timothy Willie Mathew" and concatenate them. Explain your reasoning. Your final answer should be put between two \#\# symbols, like \#\# ANSWER \#\#, at the end of your response.} \\
		\includegraphics[width=0.9em]{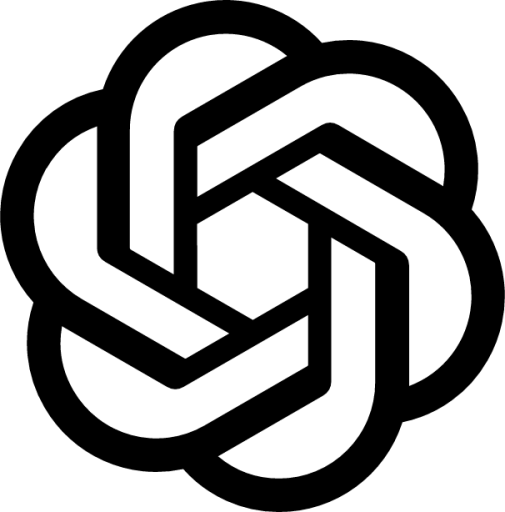} :  \textcolor{second}{The last letters of each word are: a, y, e, w. Concatenating them gives us "ayew". \#\# \textbf{ayew} \#\#.} \textcolor{rightcolor}{\textbf{(Correct)}}
	\end{tcolorbox}
	
	\tcbset{enhanced,colback=gray!10!white, arc=3mm, boxrule=0pt, width=(\linewidth-4pt)/2, attach boxed title to top left={yshift=-2mm, xshift=2mm},before=,after=\hfill,fonttitle=\bfseries, boxed title style={colback=gray!90!white, colframe=gray!50!white, size=small, rounded corners, boxrule=2pt, boxsep=2pt}}
	
	\begin{tcolorbox}[colframe=blue!50!green!50!white, equal height group=row4,title={\textbf{[P2] Critical Prompt (Huang et al., 2023)} } ]
		\includegraphics[width=0.9em]{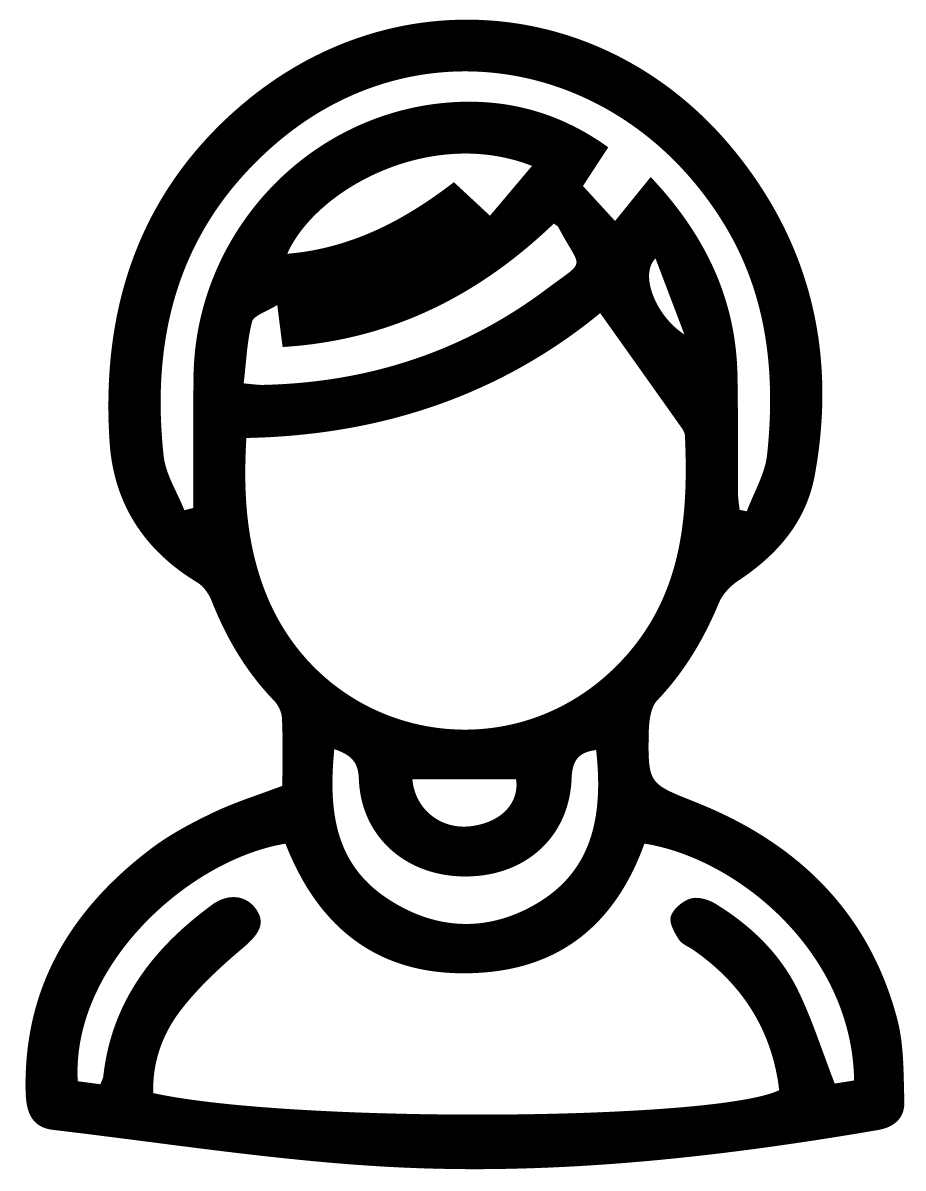} :	\textcolor{mycolor}{Review your previous answer and find problems with your answer.}\\
		\includegraphics[width=0.9em]{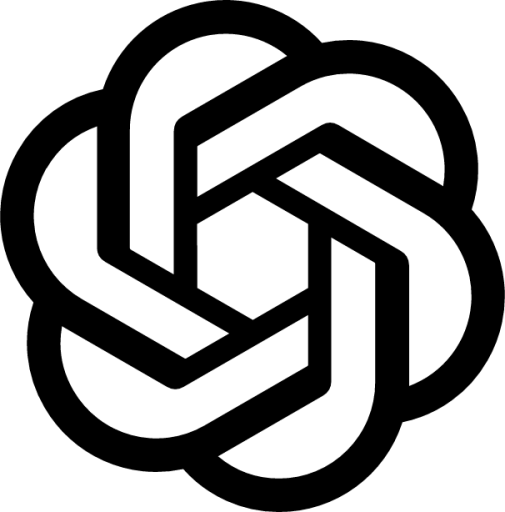} : \textcolor{second}{Upon reviewing my previous answer, I realize that I made a mistake. The last letters of each word are actually: e, y, e, d. Concatenating them gives us "eyed". I apologize for the error in my previous response. Thank you for bringing it to my attention.}
	\end{tcolorbox}
	\begin{tcolorbox}[colframe=yellow!15!pink!90!white, equal height group=row4,title={ \textbf{[P2] {IoE Prompt (Ours)}} }]
		\includegraphics[width=0.9em]{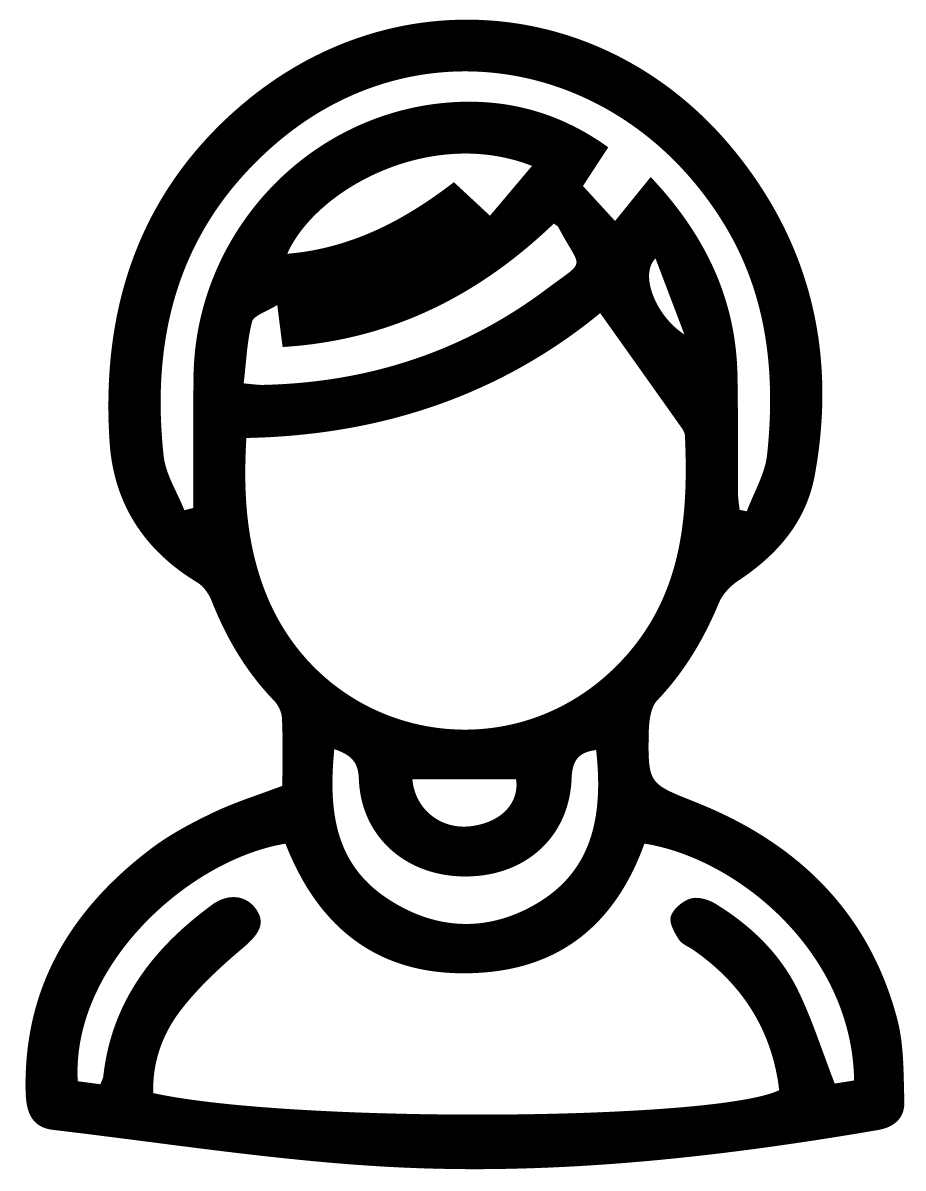} : \textcolor{mycolor}{If you are very confident about your answer, maintain your answer. Otherwise, update your answer. Your final answer should be put between two \#\# symbols, like \#\# ANSWER \#\#, at the end of your response.}\\
		\includegraphics[width=0.9em]{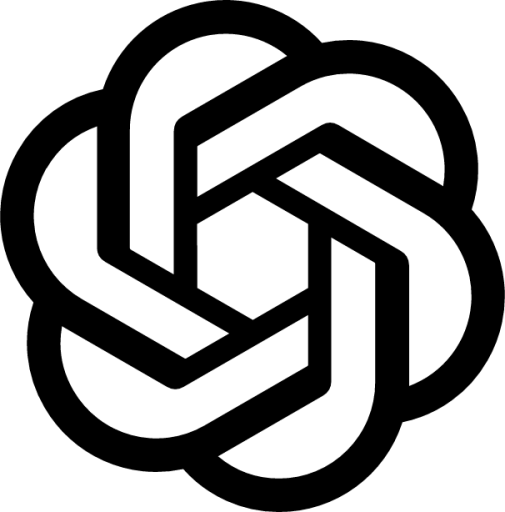} : \textcolor{second}{Upon reviewing my previous answer, I realize that I made a mistake. The last letters of each word in "Norma Timothy Willie Mathew" are a, y, e, w, not a, y, e, w. Therefore, the correct concatenation of the last letters is "ayew". \#\# \textbf{ayew} \#\#.} \textcolor{rightcolor}{\textbf{(Correct)}} 
	\end{tcolorbox}
	
	\begin{tcolorbox}[colframe=blue!50!green!50!white, equal height group=row4,title={\textbf{[P3] Update Prompt (Huang et al., 2023) } } ]
		\includegraphics[width=0.9em]{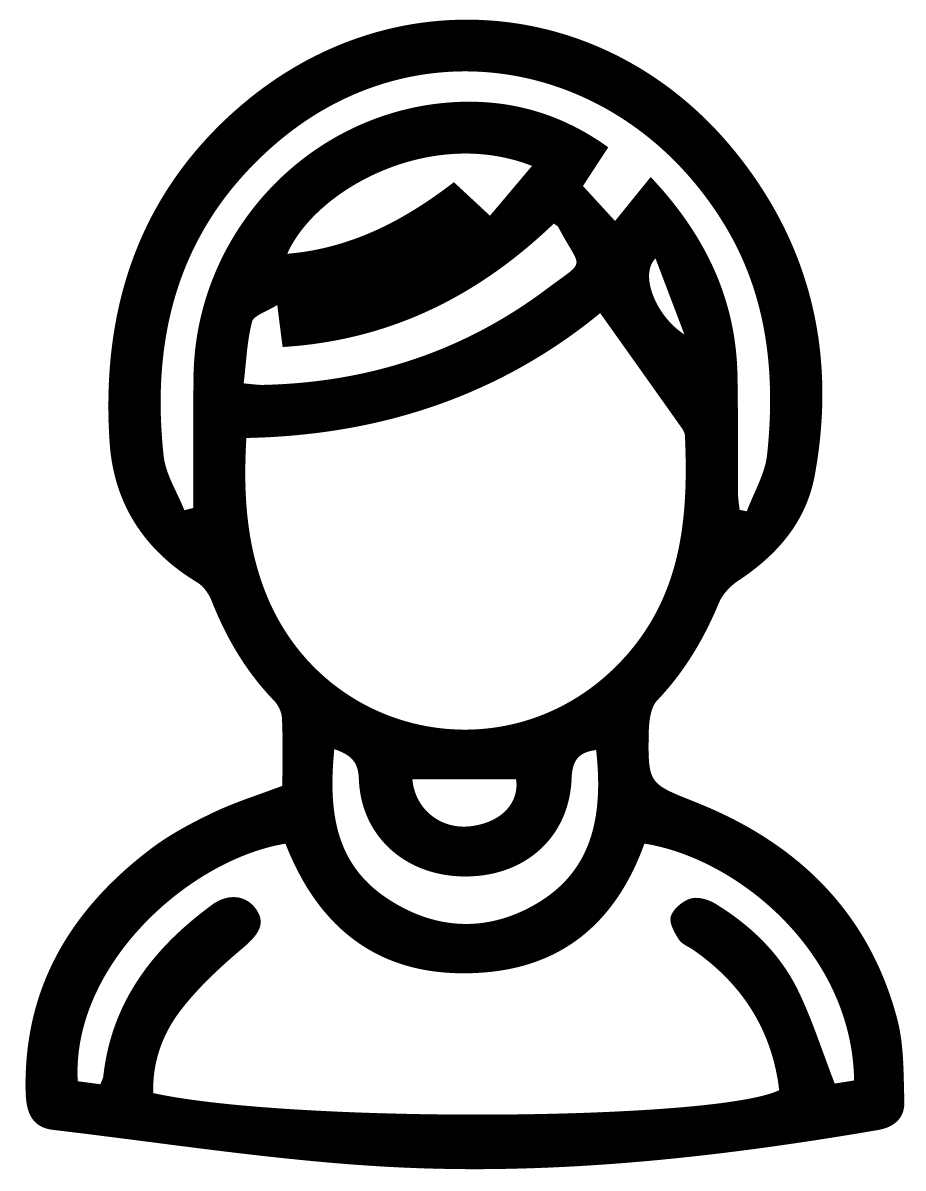} : \textcolor{mycolor}{Based on the problems you found, improve your answer. Please reiterate your answer. Your final answer should be put between two \#\# symbols, like \#\# ANSWER \#\#, at the end of your response.}\\
		\includegraphics[width=0.9em]{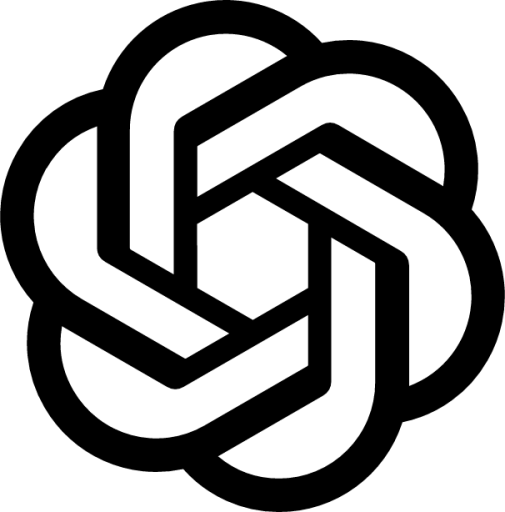} : \textcolor{second}{The last letters of each word are: e, y, e, d. Concatenating them gives us "eyed". \#\# \textbf{eyed} \#\#.} \textcolor{wrongcolor}{\textbf{(Incorrect)}} 
	\end{tcolorbox}
	\begin{tcolorbox}[colframe=yellow!15!pink!90!white, equal height group=row4,title={ \textbf{[P3] {Decision Refinement (Ours)} } }	 ]
	\textcolor{second}{<--Since the answers by the previous two prompts are the same, this decision prompt will not be executed and the final answer will be directly output.--> \\ Answer: \#\# \textbf{ayew} \#\#. \textcolor{rightcolor}{\textbf{(Correct)}} }
	\end{tcolorbox}
	\caption{One example on Last Letter Concatenation (LLC) benchmark \cite{wei2022chain} evaluated with \texttt{gpt-3.5-turbo-1106} model. The [P1] standard prompt is the same for all methods. \textbf{Left:} Critical Prompt as the baseline. \textbf{Right:} our proposed IoE-based Prompt.}
	\label{fig:llc_example}
\end{figure*}

\begin{figure*}
	\centering
	\scriptsize
        % \small 
	\begin{tcolorbox}[enhanced, arc=3mm, colback=gray!10!white, boxrule=0pt, boxsep=1pt,colframe=gray!50!white,  boxed title style={colback=gray!90!white}, boxed title style= {boxrule=2pt, boxsep=2pt}, attach boxed title to top left={yshift=-2mm, xshift=2mm}, title=\textbf{[P1] Standard Prompt}  
		]
		\includegraphics[width=0.9em]{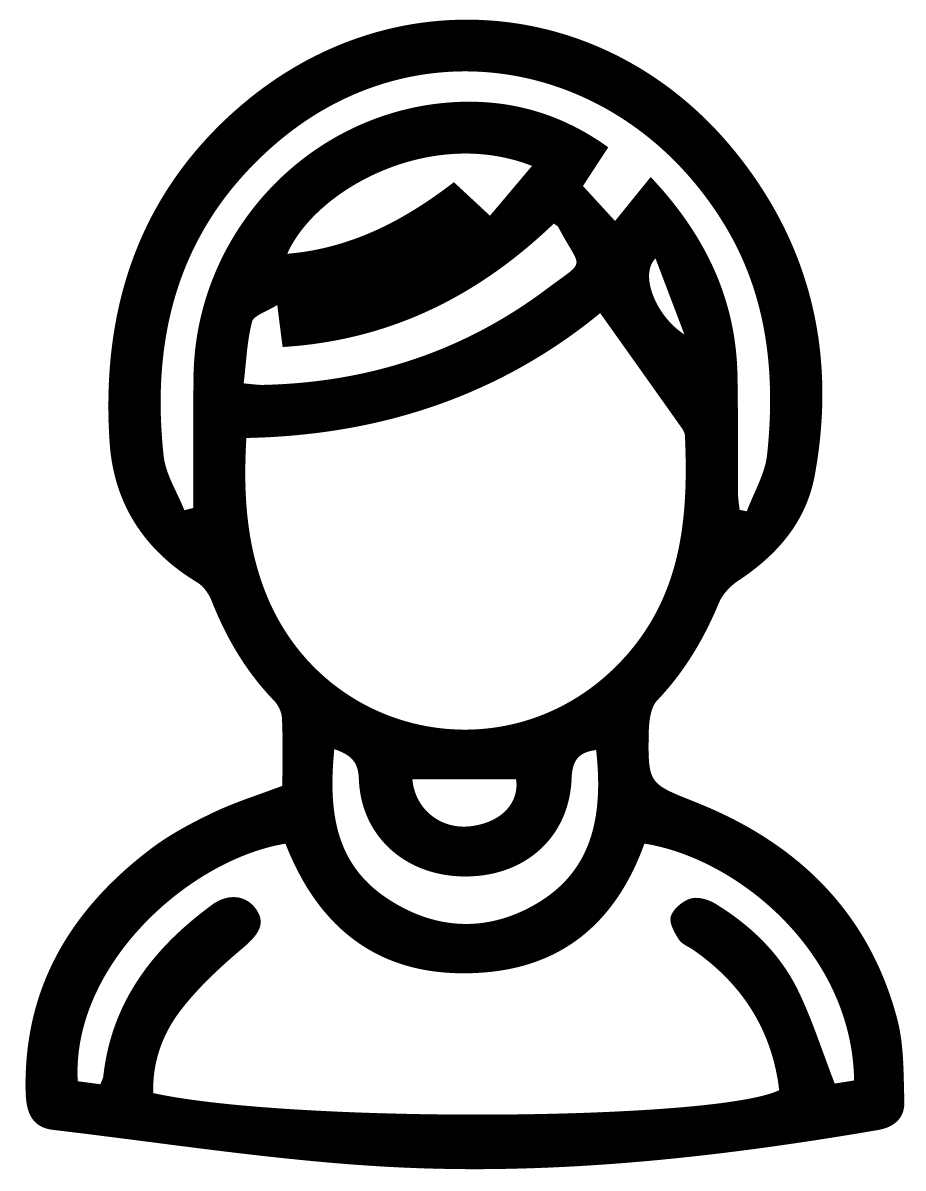} : \textcolor{mycolor}{\textbf{Context:} Peppes Pizza is a Norwegian pizza chain that serves American style and Italian style pizza.  Peppes is the largest pizza chain in Scandinavia.  The restaurant was founded by two Americans, Louis Jordan and his wife Anne from Hartford, Connecticut.  The restaurant chain is part of Umoe Catering As which consists of restaurants such as Burger King, TGI Fridays, La Baguette and Cafe Opus.  Peppes Pizza is one of the first restaurants that brought foreign food to Norway.  9 million pizzas are served by Peppes each year with deliveries in 11 cities in Norway.  Their menu was first put online in March 1995.  The servings have been described as enough for two people and that the pizza chain is a cut above the rest. Gino's East is a Chicago-based restaurant chain, notable for its deep-dish pizza (sometimes called Chicago-style pizza), and for its interior walls, which patrons have covered in graffiti and etchings.  The restaurant features deep-dish pizza baked in cast-iron pans, as well as sandwiches, soups and salads. Big Mama's \& Papa's Pizzeria is a pizza restaurant chain primarily located in Southern California.  The chain is notable for its extremely large Giant Sicilian pizza, which is claimed to be the largest deliverable pizza in the world.  Additionally, the chain gained notoriety when, during the 2014 Academy Awards, host Ellen Degeneres had Big Mama's pizzas delivered onstage. Papa John's Pizza is an American restaurant franchise company.  It runs the third largest take-out and pizza delivery restaurant chain in the United States, with headquarters in Jeffersontown, Kentucky, a suburb of Louisville. Pizza 73 is a Canadian restaurant chain that offers a number of different styles of pizza, along with chicken wings.  It has been operated by Pizza Pizza since 2007.  Toronto-based Pizza Pizza had acquired the restaurant for a total of \$CAN70.2 million.  There are 89 locations throughout Western Canada, which include the provinces of British Columbia, Alberta, and Saskatchewan.  The restaurant's name originates from its original phone number: 473 .  Founded by David Tougas and Guy Goodwin in 1985, Pizza 73 is headquartered in Edmonton, Alberta, Canada. Papa Gino's, Inc. is a restaurant chain based in Dedham, Massachusetts specializing in American-style pizza along with pasta, subs, salads, and a variety of appetizers.  There are over 150 Papa Gino's locations in Connecticut, Maine, Massachusetts, New Hampshire, and Rhode Island. Pie Five Pizza Co. is a fast casual restaurant chain specializing in handcrafted personal pizza made in less than 5 minutes.  The brand is owned by Rave Restaurant Group, which also owns Pizza Inn.  As of December 2016, Pie Five has 98 restaurants in the following locations: Alabama, Arkansas, Florida, Georgia, Illinois, Indiana, Kansas, Maryland, Minnesota, Missouri, Nebraska, North Carolina, Oklahoma, Tennessee, Texas, Utah, Kentucky, Virginia and Washington, D.C. with more than 400 additional company-owned and franchise units anticipated. Gino's Pizza and Spaghetti is a restaurant chain with 40 locations, most of them within the U.S. state of West Virginia.  The company was founded by Kenney Grant in 1961.  Many locations are shared with Tudor's Biscuit World although the Gino's brand is exclusive to West Virginia.  There is one located in Ohio, while there are stand alone Tudor's locations in eastern Kentucky, southern Ohio and southwest Virginia.  Gino's serves pizza, spaghetti, sandwiches, and more.  Company headquarters are located in Huntington, West Virginia and Nitro, West Virginia. Pizza Inn is an American restaurant chain and international food franchise, specializing in American-style pan pizza and side dishes.  The company is based in the Dallas suburb of The Colony, Texas. Tudor's Biscuit World is a restaurant chain based in Huntington, West Virginia, most commonly found in West Virginia.  Many West Virginia locations share a building with Gino's Pizza and Spaghetti, although the chain is more extensive than Gino's (which is exclusive to West Virginia), having locations in southern Ohio, eastern Kentucky, and southwestern Virginia.  In 2016 a franchise was opened in Panama City, Florida.  Tudor's serves biscuits, biscuit sandwiches, homestyle breakfasts and dinners, muffins, and several side dishes.  The chain was originally based in Charleston, West Virginia and many of the biscuit sandwiches are named for sports teams of interest in that area, including teams at Marshall University, West Virginia University, and The University of Charleston.\\ \textbf{Question:} Which restaurant chain's headquarters is further north, Pizza Inn or Papa Gino's? Explain your reasoning. Your final answer should be put between two \#\# symbols, like \#\# ANSWER \#\#, at the end of your response.} \\
		\includegraphics[width=0.9em]{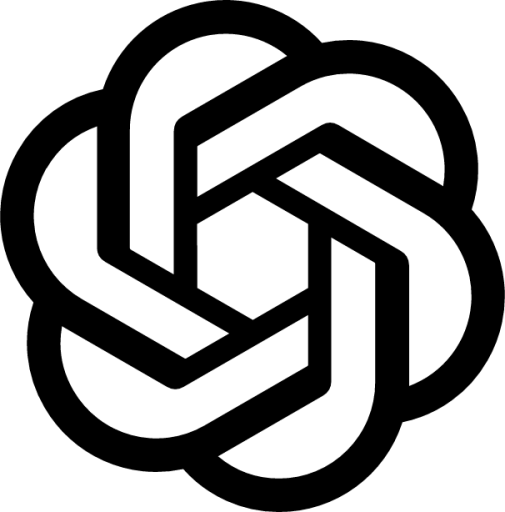} :  \textcolor{second}{Pizza Inn's headquarters is further north than Papa Gino's. Pizza Inn is based in The Colony, Texas, which is further north than Dedham, Massachusetts, where Papa Gino's is headquartered. The Colony, Texas is located at a latitude of approximately 33.0801 N, while Dedham, Massachusetts is located at a latitude of approximately 42.2418 N. Therefore, Pizza Inn's headquarters is further north. \#\# \textbf{Pizza Inn} \#\#.} \textcolor{wrongcolor}{\textbf{(Incorrect)}}
	\end{tcolorbox}
	
	\tcbset{enhanced,colback=gray!10!white, arc=3mm, boxrule=0pt, width=(\linewidth-4pt)/2, attach boxed title to top left={yshift=-2mm, xshift=2mm},before=,after=\hfill,fonttitle=\bfseries, boxed title style={colback=gray!90!white, colframe=gray!50!white, size=small, rounded corners, boxrule=2pt, boxsep=2pt}}
	
	\begin{tcolorbox}[colframe=blue!50!green!50!white, equal height group=row7,title={\textbf{[P2] Critical Prompt (Huang et al., 2023)  } } ]
		\includegraphics[width=0.9em]{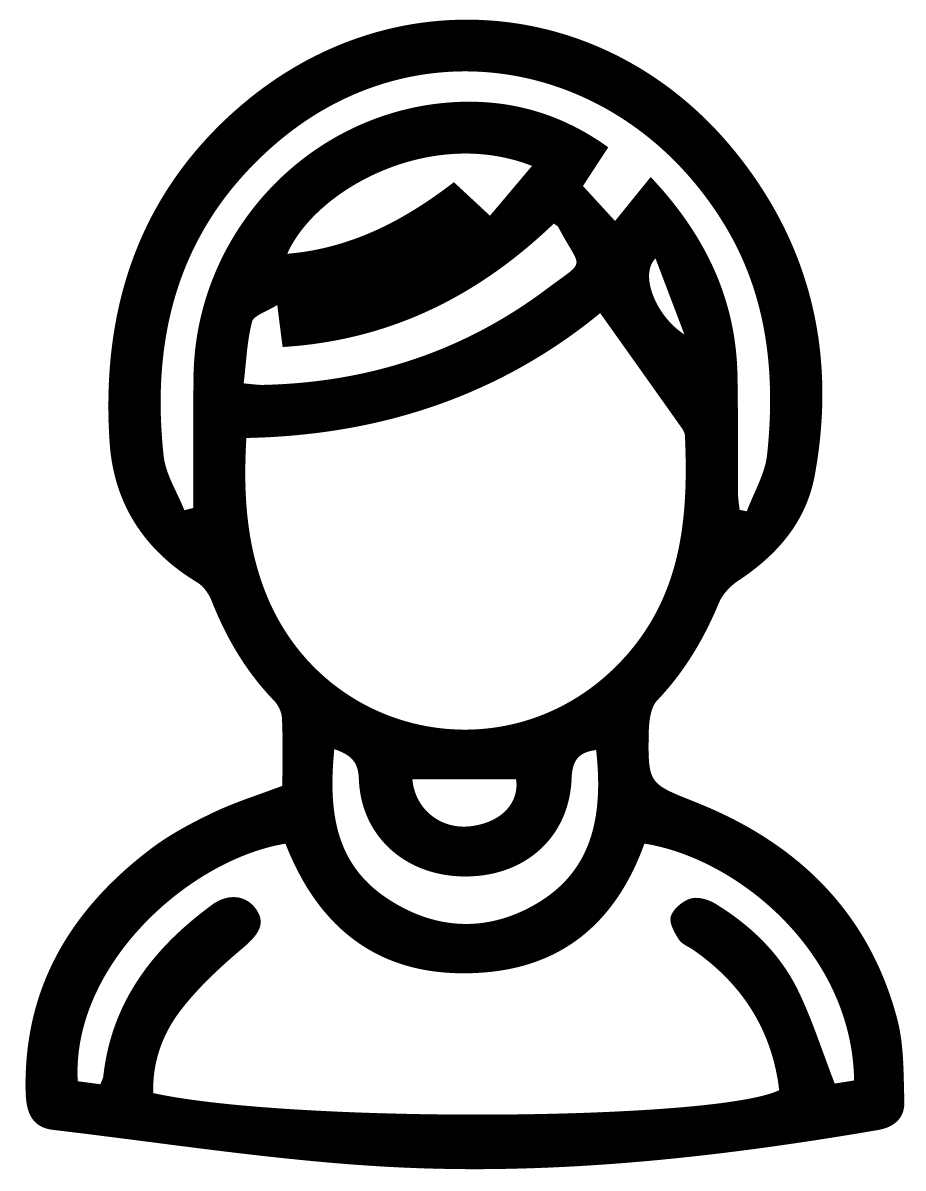} :	\textcolor{mycolor}{Review your previous answer and find problems with your answer.}\\
		\includegraphics[width=0.9em]{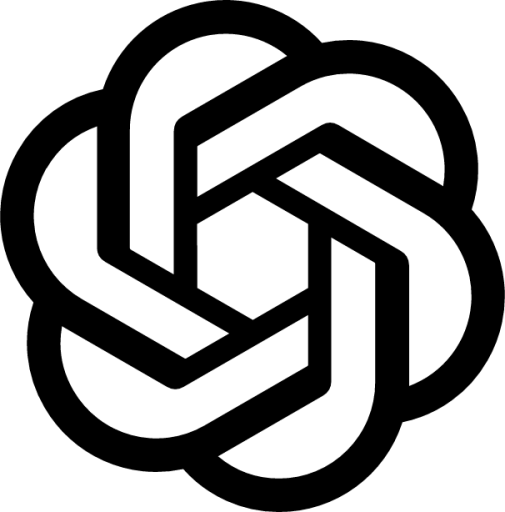} : \textcolor{second}{Apologies for the confusion. Upon reviewing the information, I realize that I made an error in my previous response. The correct answer is that Papa Gino's headquarters in Dedham, Massachusetts is further north than Pizza Inn's headquarters in The Colony, Texas. Therefore, the correct answer is Papa Gino's. Thank you for bringing this to my attention.} 
  % \textcolor{rightcolor}{\textbf{(Correct)}}
	\end{tcolorbox}
	\begin{tcolorbox}[colframe=yellow!15!pink!90!white, equal height group=row7,title={ \textbf{[P2] {IoE Prompt (Ours)}} }]
		\includegraphics[width=0.9em]{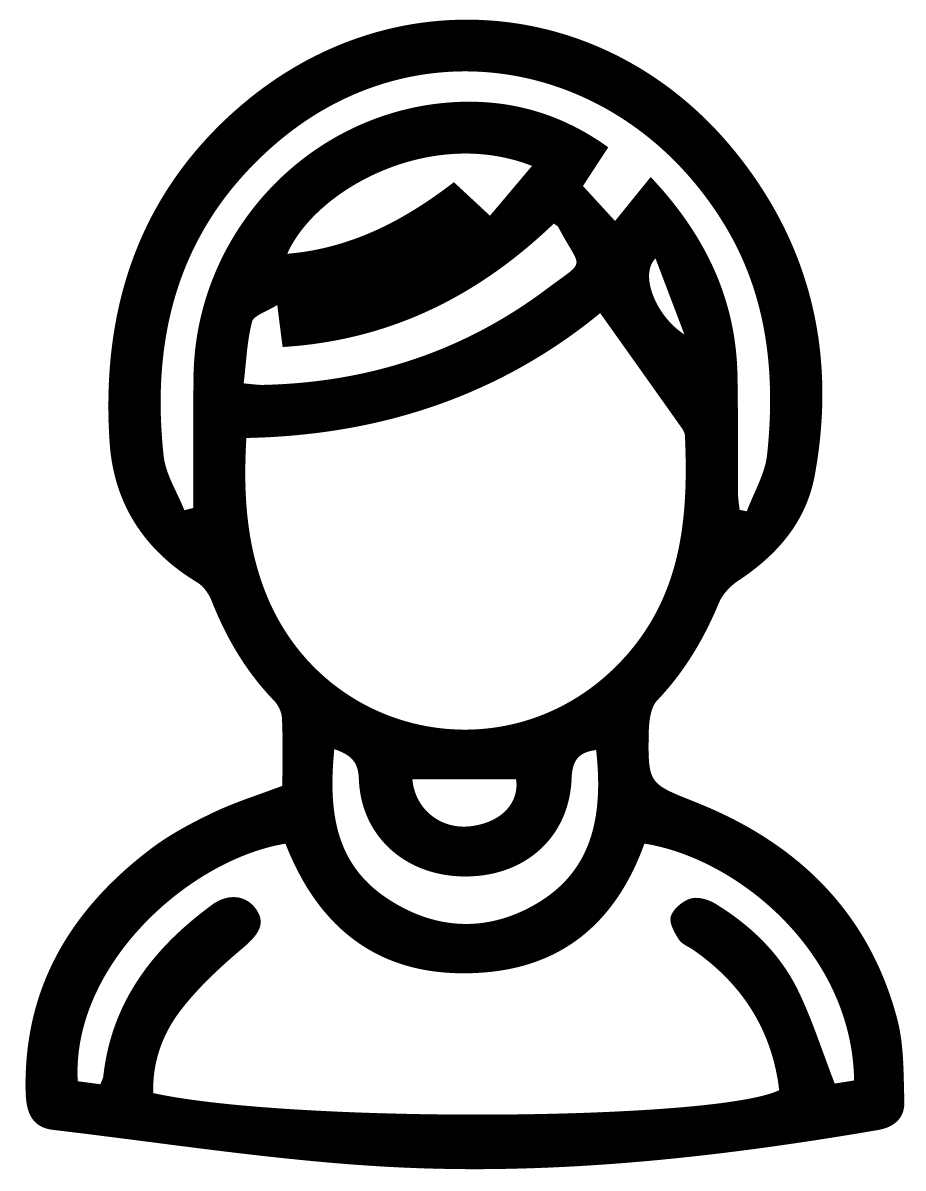} : \textcolor{mycolor}{If you are very confident about your answer, maintain your answer. Otherwise, update your answer. Your final answer should be put between two \#\# symbols, like \#\# ANSWER \#\#, at the end of your response.}\\
		\includegraphics[width=0.9em]{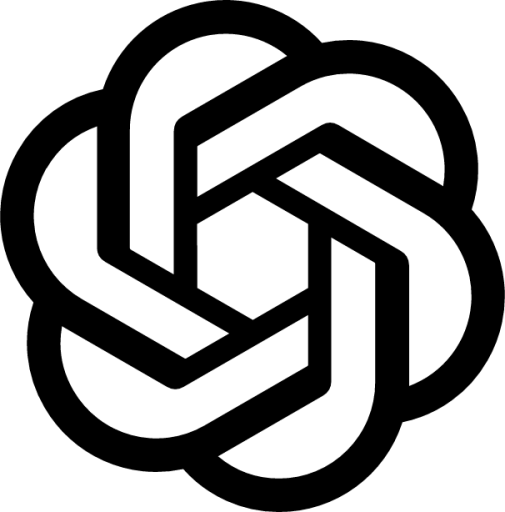} : \textcolor{second}{Upon reviewing my previous answer, I realize that I made an error. Dedham, Massachusetts, where Papa Gino's is headquartered, is further north than The Colony, Texas, where Pizza Inn is based. Dedham, Massachusetts is located at a latitude of approximately 42.2418 N, while The Colony, Texas is located at a latitude of approximately 33.0801 N. Therefore, Papa Gino's headquarters is further north. \#\# \textbf{Papa Gino's} \#\#.} \textcolor{rightcolor}{\textbf{(Correct)}} 
	\end{tcolorbox}
	
	\begin{tcolorbox}[colframe=blue!50!green!50!white,equal height group=row8, title={\textbf{[P3] Update Prompt (Huang et al., 2023) } } ]
		\includegraphics[width=0.9em]{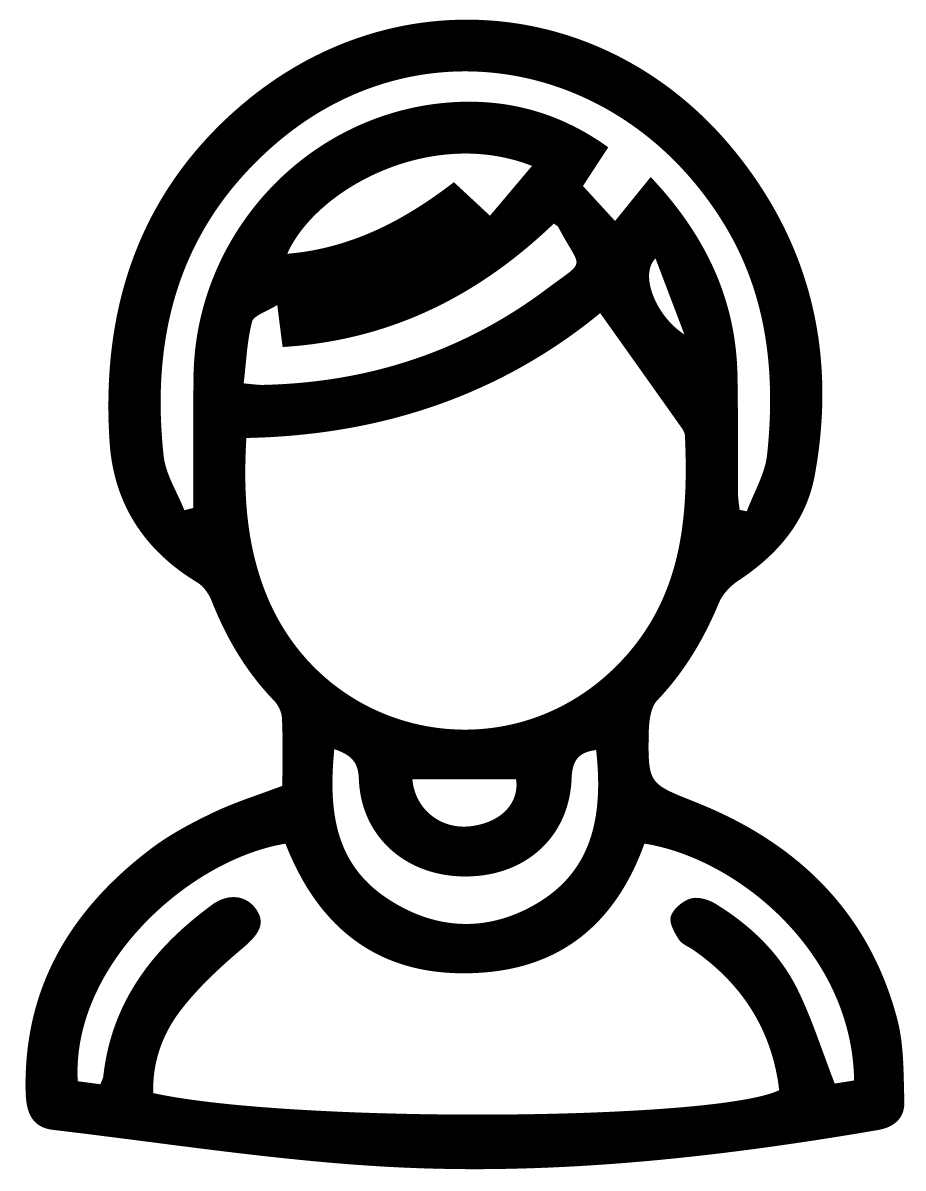} : \textcolor{mycolor}{Based on the problems you found, improve your answer. Please reiterate your answer. Your final answer should be put between two \#\# symbols, like \#\# ANSWER \#\#, at the end of your response.}\\
		\includegraphics[width=0.9em]{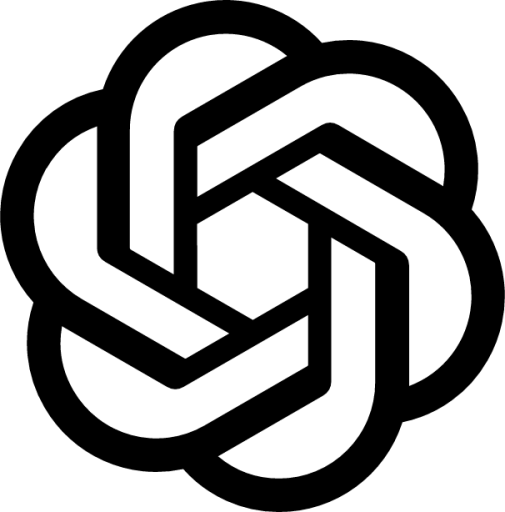} : \textcolor{second}{Upon reviewing the information, I realize that I made an error in my previous response. The correct answer is that Papa Gino's headquarters in Dedham, Massachusetts is further north than Pizza Inn's headquarters in The Colony, Texas. Therefore, the correct answer is \#\# \textbf{Papa Gino's} \#\#.} \textcolor{rightcolor}{\textbf{(Correct)}} 
	\end{tcolorbox}
	\begin{tcolorbox}[colframe=yellow!15!pink!90!white,equal height group=row8, title={ \textbf{[P3] {Decision Refinement (Ours)} } }	 ]
	\includegraphics[width=0.9em]{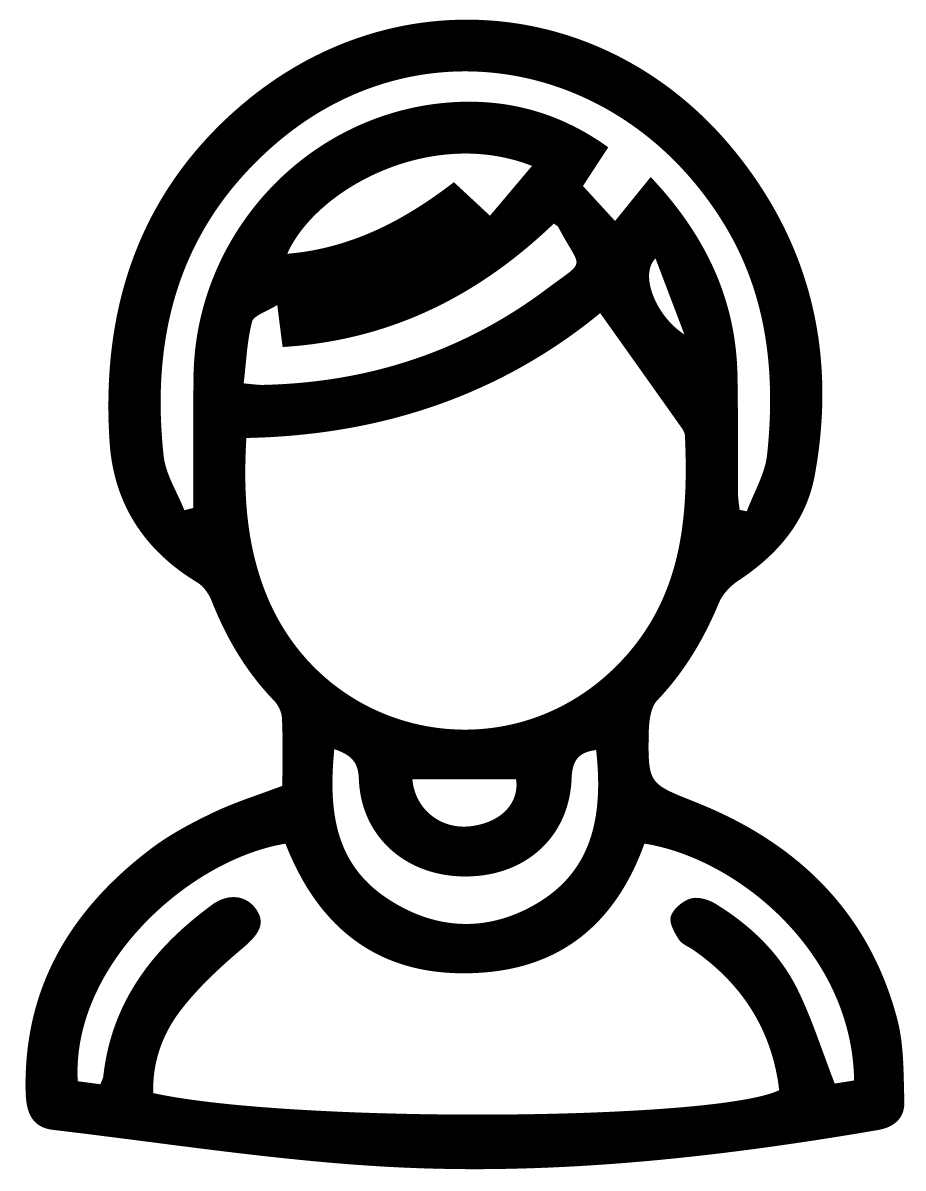} : \textcolor{mycolor}{{You give two different answers in previous responses. Check the problem and your answers again, and give the best answer. Your final answer should be put between two \#\# symbols, like \#\# ANSWER \#\#, at the end of your response.}} \\
	\includegraphics[width=0.9em]{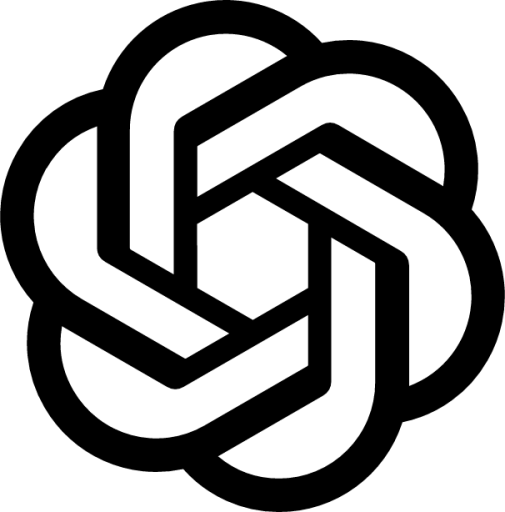} : \textcolor{second}{Upon re-evaluating the locations of the headquarters for Pizza Inn and Papa Gino's, I have found that The Colony, Texas, where Pizza Inn is based, is further south than Dedham, Massachusetts, where Papa Gino's is headquartered. The Colony, Texas is located at a latitude of approximately 33.0801 N, while Dedham, Massachusetts is located at a latitude of approximately 42.2418 N. Therefore, Papa Gino's headquarters is further north. \#\# \textbf{Papa Gino's} \#\#.} \textcolor{rightcolor}{\textbf{(Correct)}}
	\end{tcolorbox}
 
	\caption{One example on HotpotQA benchmark with context \cite{yang2018hotpotqa} evaluated by \texttt{gpt-3.5-turbo-1106} model. The [P1] standard prompt is the same for all methods. \textbf{Left:} Critical Prompt as the baseline. \textbf{Right:} our proposed IoE-based Prompt.}
	\label{fig:Hotpot_example}
\end{figure*}

\begin{figure*}
	\centering
	% \scriptsize
        \small
	\begin{tcolorbox}[enhanced, arc=3mm, colback=gray!10!white, boxrule=0pt, boxsep=1pt,colframe=gray!50!white,  boxed title style={colback=gray!90!white}, boxed title style= {boxrule=2pt, boxsep=2pt}, attach boxed title to top left={yshift=-2mm, xshift=2mm}, title=\textbf{[P1] Standard Prompt}]
		\includegraphics[width=0.9em]{figures/user_icon.png} : \textcolor{mycolor}{Given an image}: \\ \ \ \includegraphics[width=15em]{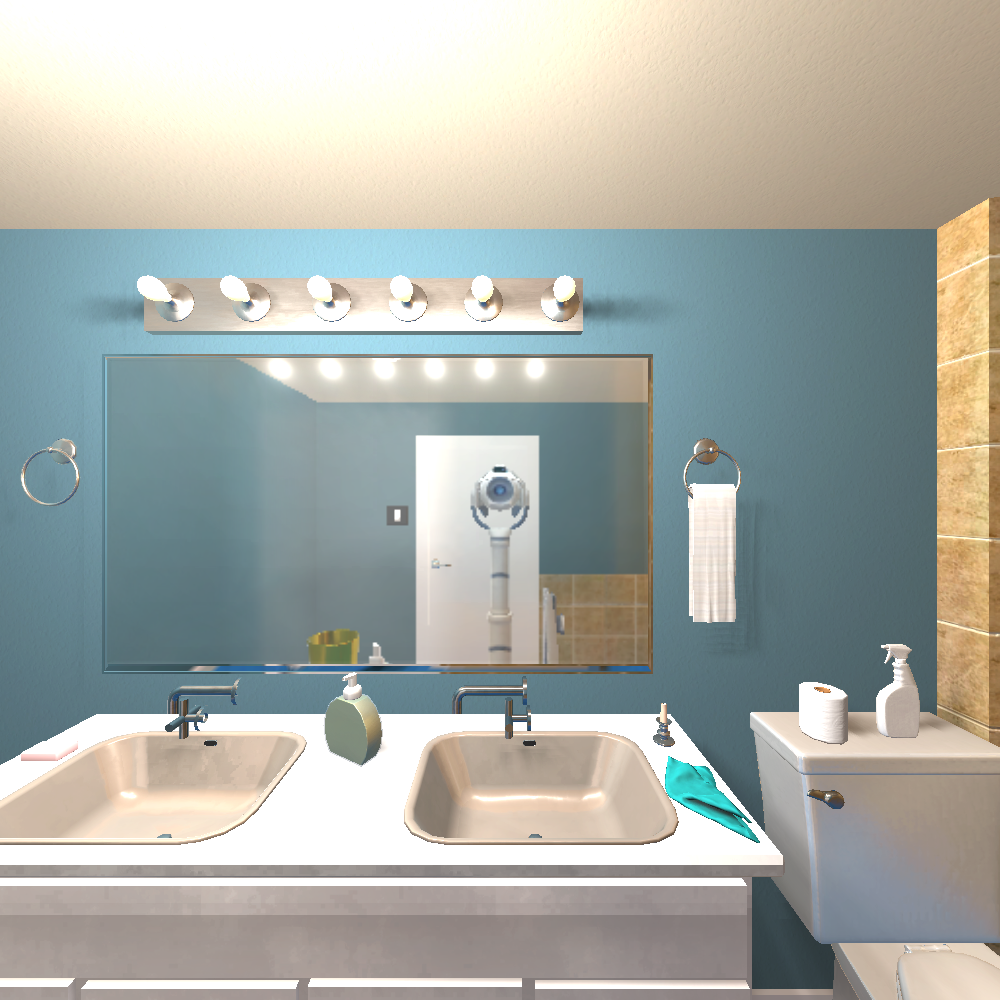} \\\textcolor{mycolor}{You are a domestic robot that helps me do housework in the simulation room. Your goal is: clean the CounterTop. You have cloth in your hands. Based on the image, what action should you do next? \\ (A) find a cleaning cloth \\ (B) use the cloth to clean the countertop \\ (C) water the faucet with the spray bottle.\\ Explain your reasoning. Your final answer should be put between two \#\# symbols, like \#\# ANSWER \#\#, at the end of your response.} \\
		\includegraphics[width=0.9em]{figures/llm_icon.png} :  \textcolor{second}{Since I already have a cloth in my hands, there is no need to find another cleaning cloth. Therefore, option (A) is not necessary. Option (C) mentions watering the faucet with the spray bottle, but the goal is to clean the countertop, not to water the faucet. The most logical action to take next would be to use the cloth I have to clean the countertop, which aligns with the goal given. My final answer is: \#\# \textbf{B} \#\#.} \textcolor{rightcolor}{\textbf{(Correct)}}
	\end{tcolorbox}
	
	\tcbset{enhanced,colback=gray!10!white, arc=3mm, boxrule=0pt, width=(\linewidth-4pt)/2, attach boxed title to top left={yshift=-2mm, xshift=2mm},before=,after=\hfill,fonttitle=\bfseries, boxed title style={colback=gray!90!white, colframe=gray!50!white, size=small, rounded corners, boxrule=2pt, boxsep=2pt}}
	
	\begin{tcolorbox}[colframe=blue!50!green!50!white, equal height group=fig9-1,title={\textbf{[P2] Critical Prompt (Huang et al., 2023)} } ]
		\includegraphics[width=0.9em]{figures/user_icon.png} :	\textcolor{mycolor}{Review your previous answer and find problems with your answer.}\\
		\includegraphics[width=0.9em]{figures/llm_icon.png} : \textcolor{second}{Upon reviewing my previous answer, I realize that I made a mistake in interpreting option (C). The option mentions "watering the faucet with the spray bottle," which doesn't make sense in the context of cleaning. It's likely that the option meant to say "wetting the countertop with the spray bottle" or "spraying the cleaning solution onto the countertop." Given this, the correct sequence of actions for cleaning a countertop would be to first spray the cleaning solution onto the surface (assuming the spray bottle contains a cleaning solution) and then use the cloth to wipe and clean the countertop. Therefore, the correct next action would be to wet the countertop with the spray bottle before wiping it with the cloth. My revised final answer is: \#\# \textbf{C} \#\#.}
	\end{tcolorbox}
	\begin{tcolorbox}[colframe=yellow!15!pink!90!white, equal height group=fig9-1,title={ \textbf{[P2] {IoE Prompt (Ours)}} }]
		\includegraphics[width=0.9em]{figures/user_icon.png} : \textcolor{mycolor}{If you are very confident about your answer, maintain your answer. Otherwise, update your answer. Your final answer should be put between two \#\# symbols, like \#\# ANSWER \#\#, at the end of your response.}\\
		\includegraphics[width=0.9em]{figures/llm_icon.png} : \textcolor{second}{Upon reviewing my previous answer, I still believe that option (B) is the correct action to take. The goal is to clean the countertop, and I already have a cloth in my hands. It makes sense to use the cloth to wipe down the countertop. Option (A) is unnecessary because I already have a cloth, and option (C) does not directly contribute to cleaning the countertop. Therefore, I am confident in maintaining my previous answer. Final answer: \#\# \textbf{B} \#\#.} \textcolor{rightcolor}{\textbf{(Correct)}} 
	\end{tcolorbox}
	
	\begin{tcolorbox}[colframe=blue!50!green!50!white, equal height group=fig9-2,title={\textbf{[P3] Update Prompt (Huang et al., 2023) } } ]
		\includegraphics[width=0.9em]{figures/user_icon.png} : \textcolor{mycolor}{Based on the problems you found, improve your answer. Please reiterate your answer. Your final answer should be put between two \#\# symbols, like \#\# ANSWER \#\#, at the end of your response.}\\
		\includegraphics[width=0.9em]{figures/llm_icon.png} : \textcolor{second}{My revised final answer is: \#\# \textbf{C} \#\#} \textcolor{wrongcolor}{\textbf{(Incorrect)}} 
	\end{tcolorbox}
	\begin{tcolorbox}[colframe=yellow!15!pink!90!white, equal height group=fig9-2,title={ \textbf{[P3] {Decision Refinement (Ours)} } }	 ]
	\textcolor{second}{<--Since the answers by the previous two prompts are the same, this decision prompt will not be executed and the final answer will be directly output.--> \\ Answer: \#\# \textbf{B} \#\#. \textcolor{rightcolor}{\textbf{(Correct)}} }
	\end{tcolorbox}
	\caption{One example on Domestic Robot benchmark \cite{cai2023benchlmm} evaluated with \texttt{gpt-4-vision-preview} model. The [P1] standard prompt is the same for all methods. \textbf{Left:} Critical Prompt as the baseline. \textbf{Right:} our proposed IoE-based Prompt.}
	\label{fig:gpt4v_example}
\end{figure*}

In the main paper, we only provide a simplified example of GSM8K. Here, we provide complete evaluation examples for each benchmark where our IoE-based Prompt leads to correct answers, including the prompts and the full responses by the LLMs. Specifically, we provide the following examples:

\begin{itemize}
    \item Figure \ref{fig:GSM8K_example}: an example on GSM8K \cite{cobbe2021training}.
    \item Figure \ref{fig:SVAMP_example}: an example on SVAMP \cite{patel2021nlp}.
    \item Figure \ref{fig:sports_example}: an example on Sports Understanding \cite{cai2023benchlmm}.
    \item Figure \ref{fig:llc_example}: an example on LLC \cite{wei2022chain}.
    \item Figure \ref{fig:Hotpot_example}: an example on HotpotQA \cite{yang2018hotpotqa}.
    \item Figure \ref{fig:gpt4v_example}: an example on Domestic Robot \cite{cai2023benchlmm}.
\end{itemize}

\subsection{Failed Examples}
\label{appendix:failed}

\begin{figure}
	\centering
	\small
	\begin{tcolorbox}[enhanced, arc=3mm, colback=gray!10!white, boxrule=0pt, boxsep=1pt,colframe=gray!50!white,  boxed title style={colback=gray!90!white}, boxed title style= {boxrule=2pt, boxsep=2pt}, attach boxed title to top left={yshift=-2mm, xshift=2mm}, title=\textbf{[P1] Standard Prompt}  
		]
		\includegraphics[width=0.9em]{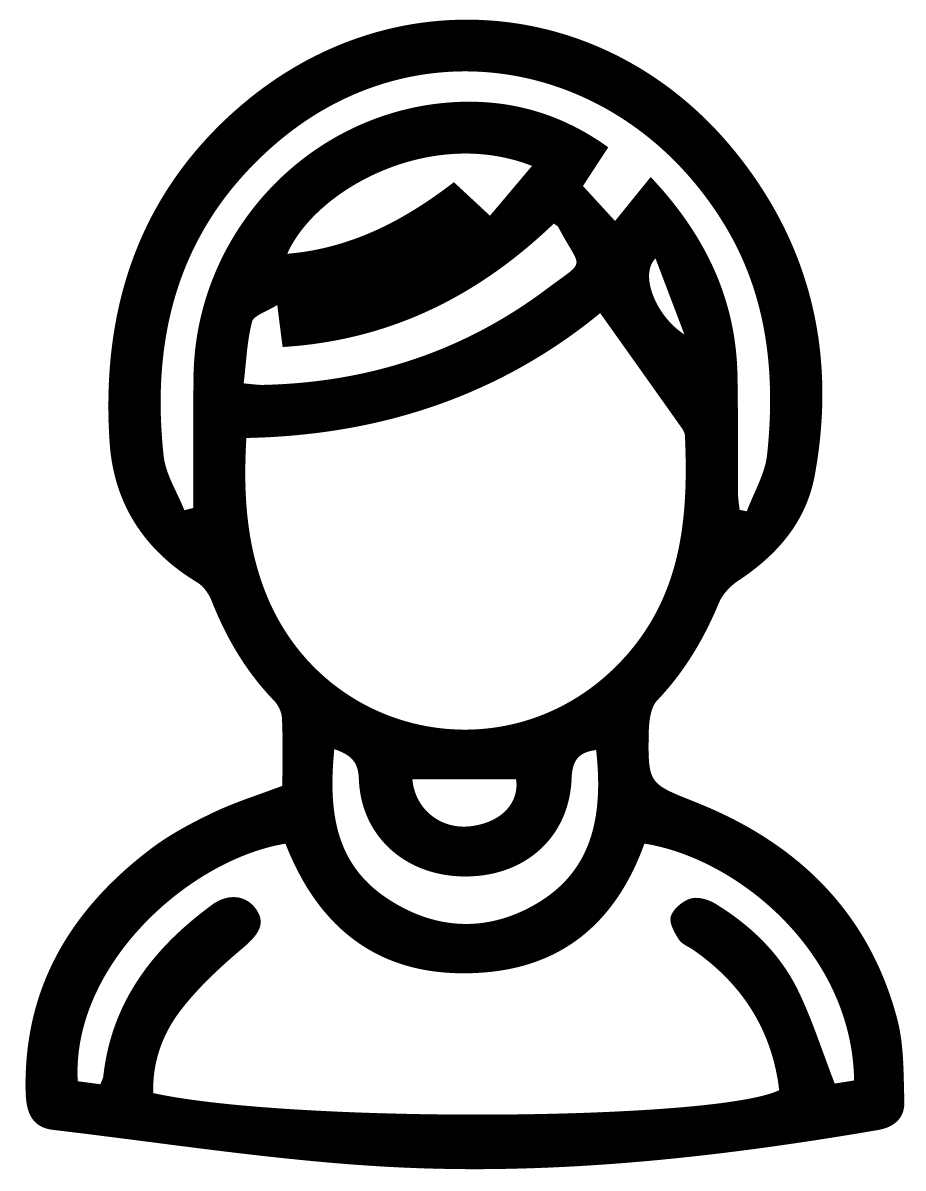} : \textcolor{mycolor}{Boris has 100 apples. Beck has 23 fewer apples than Boris. If Boris gives Beck 10 apples, how many fewer apples does Beck have than Boris now? Explain your reasoning. Your final answer should be put between two \#\# symbols, like \#\# ANSWER \#\#, at the end of your response.} \\
		\includegraphics[width=0.9em]{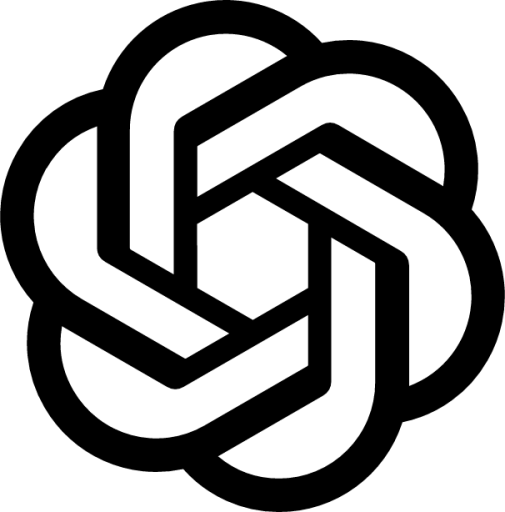} :  \textcolor{second}{First, we need to find out how many apples Beck has. Since Beck has 23 fewer apples than Boris, we can calculate Beck's apples by subtracting 23 from Boris's apples: 100 - 23 = 77 apples. Now, if Boris gives Beck 10 apples, Beck will have 77 + 10 = 87 apples. To find out how many fewer apples Beck has than Boris now, we can subtract Beck's apples from Boris's apples: 100 - 87 = 13. So, Beck has 13 fewer apples than Boris now. Therefore, the final answer is \#\# \textbf{13} \#\#.} \textcolor{wrongcolor}{\textbf{(Incorrect)}}
	\end{tcolorbox}

\begin{tcolorbox}[enhanced, arc=3mm, colback=gray!10!white, boxrule=0pt, boxsep=1pt,colframe=gray!50!white,  boxed title style={colback=gray!90!white}, boxed title style= {boxrule=2pt, boxsep=2pt}, attach boxed title to top left={yshift=-2mm, xshift=2mm}, title=\textbf{[P2] IoE Prompt}
	]
	\includegraphics[width=0.9em]{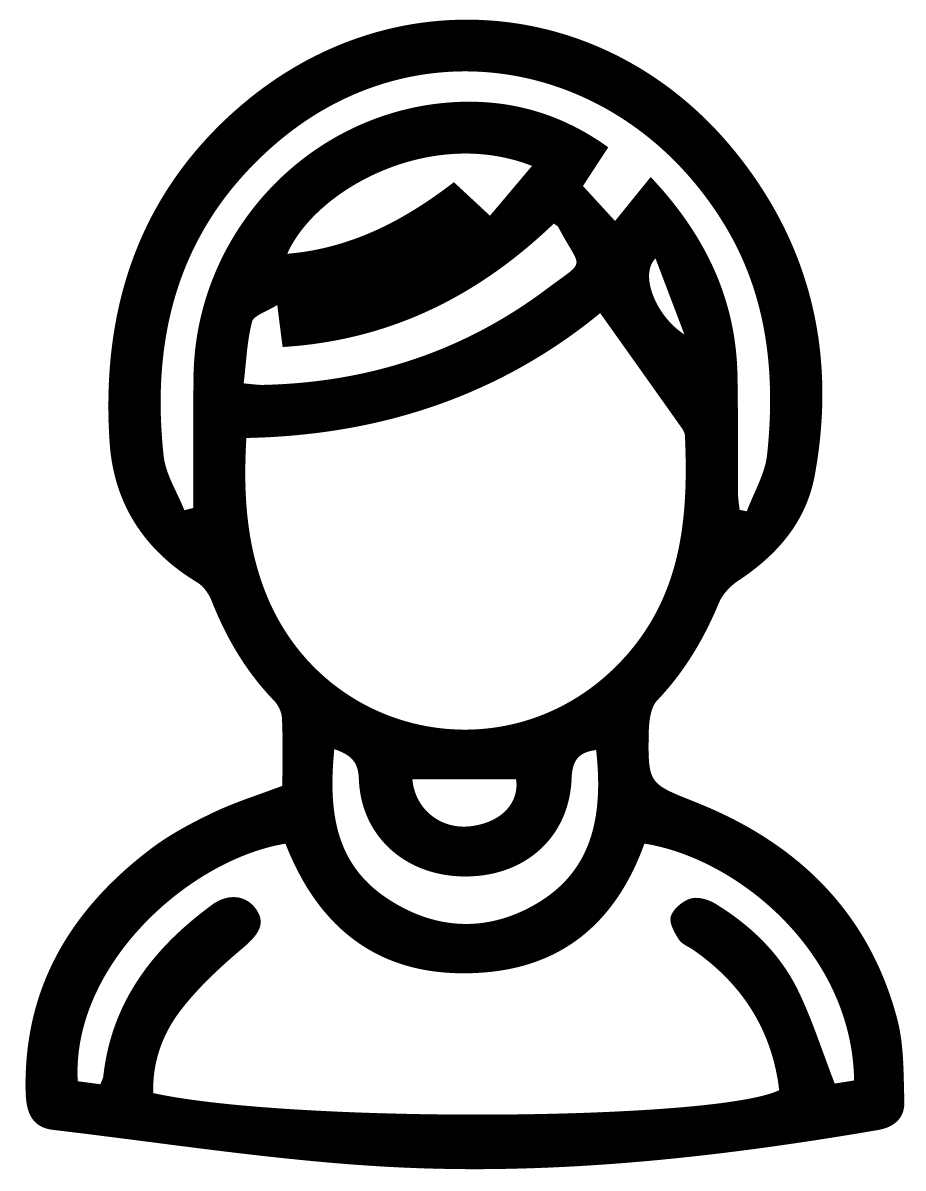} : \textcolor{mycolor}{If you are very confident about your answer, maintain your answer. Otherwise, update your answer. Your final answer should be put between two \#\# symbols, like \#\# ANSWER \#\#, at the end of your response.} \\
	\includegraphics[width=0.9em]{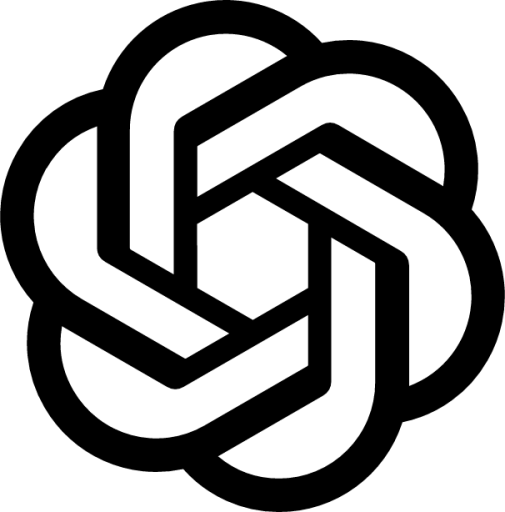} :  \textcolor{second}{Upon reviewing my previous answer, I realize there was a mistake in the calculation. If Boris gives Beck 10 apples, Beck will have 77 + 10 = 87 apples. To find out how many fewer apples Beck has than Boris now, we can subtract Beck's apples from Boris's apples: 100 - 87 = 13. So, Beck has 13 fewer apples than Boris now. Therefore, the final answer is \#\# \textbf{13} \#\#.} \textcolor{wrongcolor}{\textbf{(Incorrect)}}
\end{tcolorbox}

\begin{tcolorbox}[enhanced, arc=3mm, colback=gray!10!white, boxrule=0pt, boxsep=1pt,colframe=gray!50!white,  boxed title style={colback=gray!90!white}, boxed title style= {boxrule=2pt, boxsep=2pt}, attach boxed title to top left={yshift=-2mm, xshift=2mm}, title=\textbf{[P3] Decision Refinement}
	]
    \textcolor{second}{<--Since the answers by the previous two prompts are the same, this decision prompt will not be executed and the final answer will be directly output.--> \\ Answer: \#\# \textbf{13} \#\#. \textcolor{wrongcolor}{\textbf{(Incorrect)}} }

\end{tcolorbox}
\caption{One example on GSM8K \cite{cobbe2021training} with \texttt{gpt-3.5-turbo-1106} model, where our IoE-based prompt failed to achieve self-correction.}
\label{fig:failed_prompts}
\end{figure}

\begin{figure}
	\centering
	\small
	\begin{tcolorbox}[enhanced, arc=3mm, colback=gray!10!white, boxrule=0pt, boxsep=1pt,colframe=gray!50!white,  boxed title style={colback=gray!90!white}, boxed title style= {boxrule=2pt, boxsep=2pt}, attach boxed title to top left={yshift=-2mm, xshift=2mm}, title=\textbf{[P1] Standard Prompt}  
		]
		\includegraphics[width=0.9em]{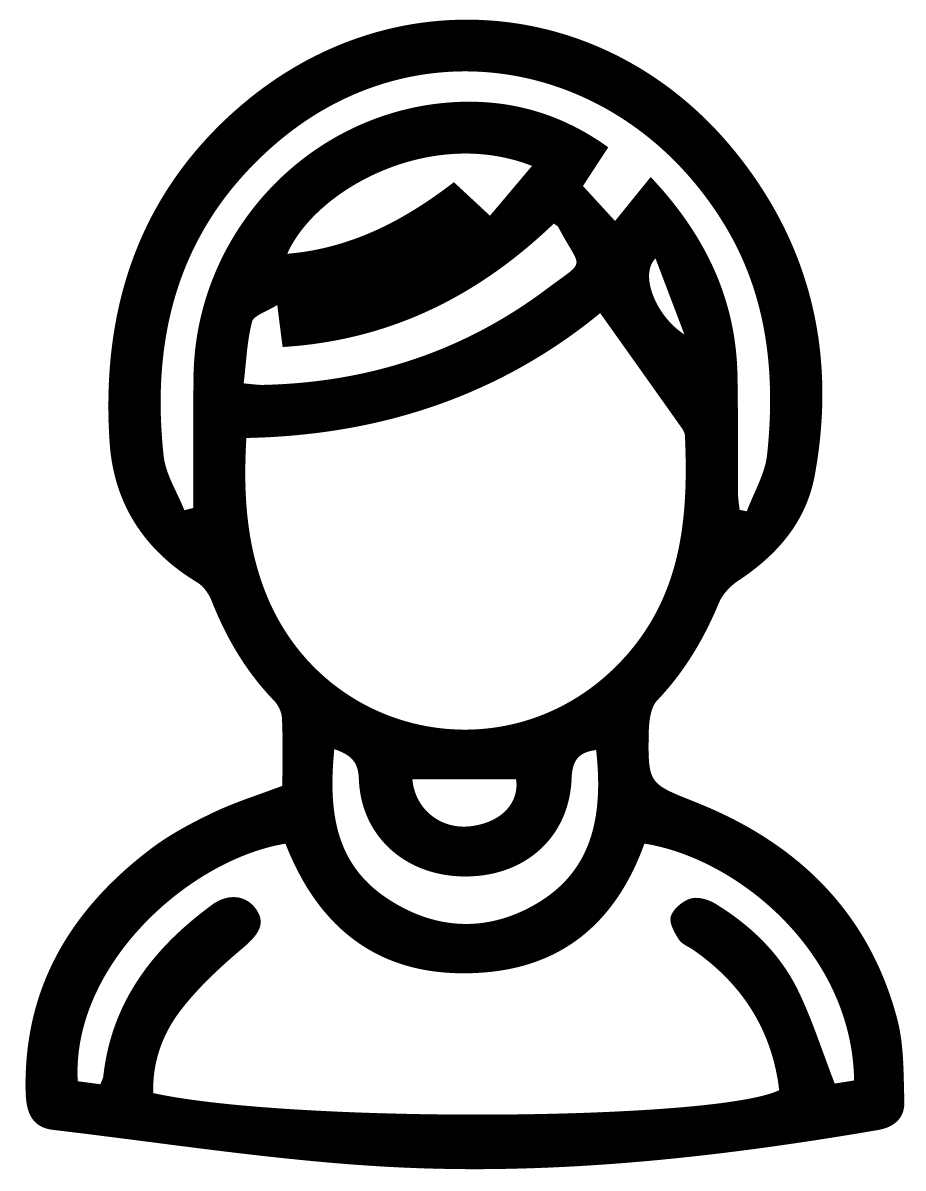} : \textcolor{mycolor}{Josh decides to try flipping a house.  He buys a house for \$80,000 and then puts in \$50,000 in repairs.  This increased the value of the house by 150\%.  How much profit did he make? Explain your reasoning. Your final answer should be put between two \#\# symbols, like \#\# ANSWER \#\#, at the end of your response.} \\
		\includegraphics[width=0.9em]{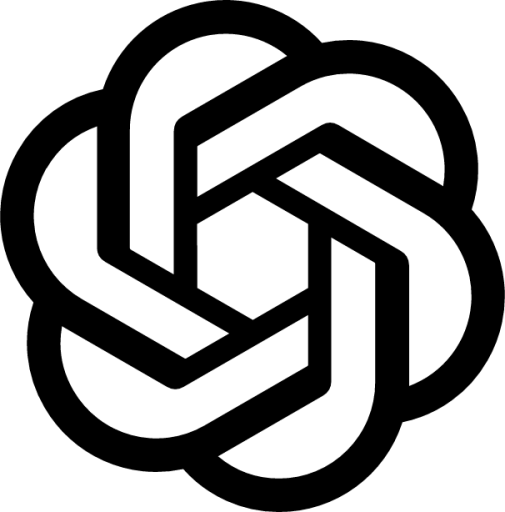} :  \textcolor{second}{First, let's calculate the increased value of the house after the repairs. The increase in value is 150\% of the original value, so the new value of the house is 250\% of the original value. The original value of the house is \$80,000, so 250\% of \$80,000 is \$200,000. Now, let's calculate the total cost of the house and repairs. Josh bought the house for \$80,000 and put in \$50,000 in repairs, so the total cost is \$80,000 + \$50,000 = \$130,000. To find the profit, we subtract the total cost from the increased value of the house: \$200,000 - \$130,000 = \$70,000. So, Josh made a profit of \$70,000. Therefore, the profit he made is \#\# \textbf{70000} \#\#.} \textcolor{rightcolor}{\textbf{(Correct)}}
	\end{tcolorbox}

\begin{tcolorbox}[enhanced, arc=3mm, colback=gray!10!white, boxrule=0pt, boxsep=1pt,colframe=gray!50!white,  boxed title style={colback=gray!90!white}, boxed title style= {boxrule=2pt, boxsep=2pt}, attach boxed title to top left={yshift=-2mm, xshift=2mm}, title=\textbf{[P2] IoE Prompt}
	]
	\includegraphics[width=0.9em]{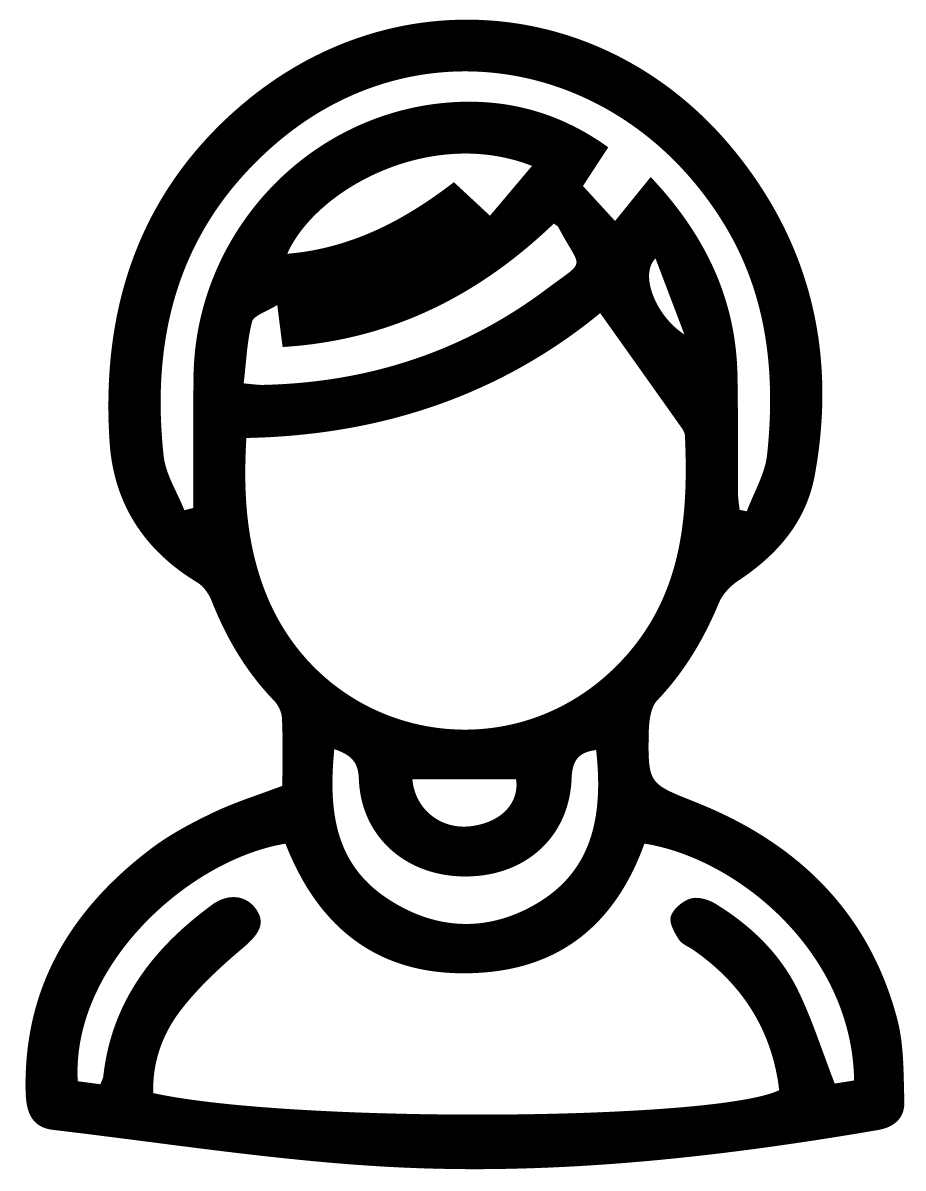} : \textcolor{mycolor}{If you are very confident about your answer, maintain your answer. Otherwise, update your answer. Your final answer should be put between two \#\# symbols, like \#\# ANSWER \#\#, at the end of your response.} \\
	\includegraphics[width=0.9em]{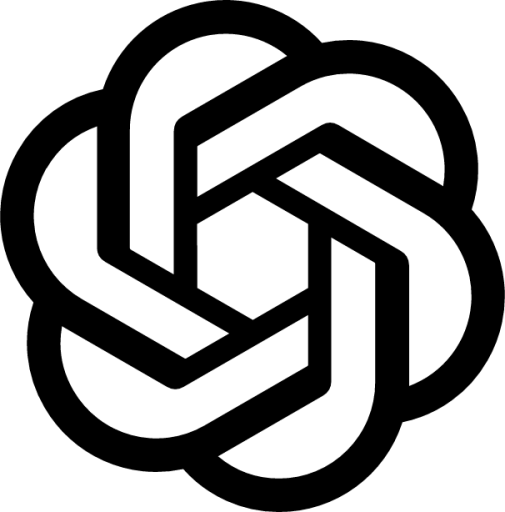} :  \textcolor{second}{Upon reviewing my previous answer, I realize there was a mistake in the calculation. I apologize for the confusion. Let's recalculate the profit. The increased value of the house after the repairs is \$80,000 + \$50,000 = \$130,000. The profit is the increased value of the house minus the total cost, so \$130,000 - \$130,000 = \$0. So, the correct profit is \$0. Therefore, the profit he made is \#\# \textbf{0} \#\#.} \textcolor{wrongcolor}{\textbf{(Incorrect)}}
\end{tcolorbox}

\begin{tcolorbox}[enhanced, arc=3mm, colback=gray!10!white, boxrule=0pt, boxsep=1pt,colframe=gray!50!white,  boxed title style={colback=gray!90!white}, boxed title style= {boxrule=2pt, boxsep=2pt}, attach boxed title to top left={yshift=-2mm, xshift=2mm}, title=\textbf{[P3] Decision Refinement}
	]
	\includegraphics[width=0.9em]{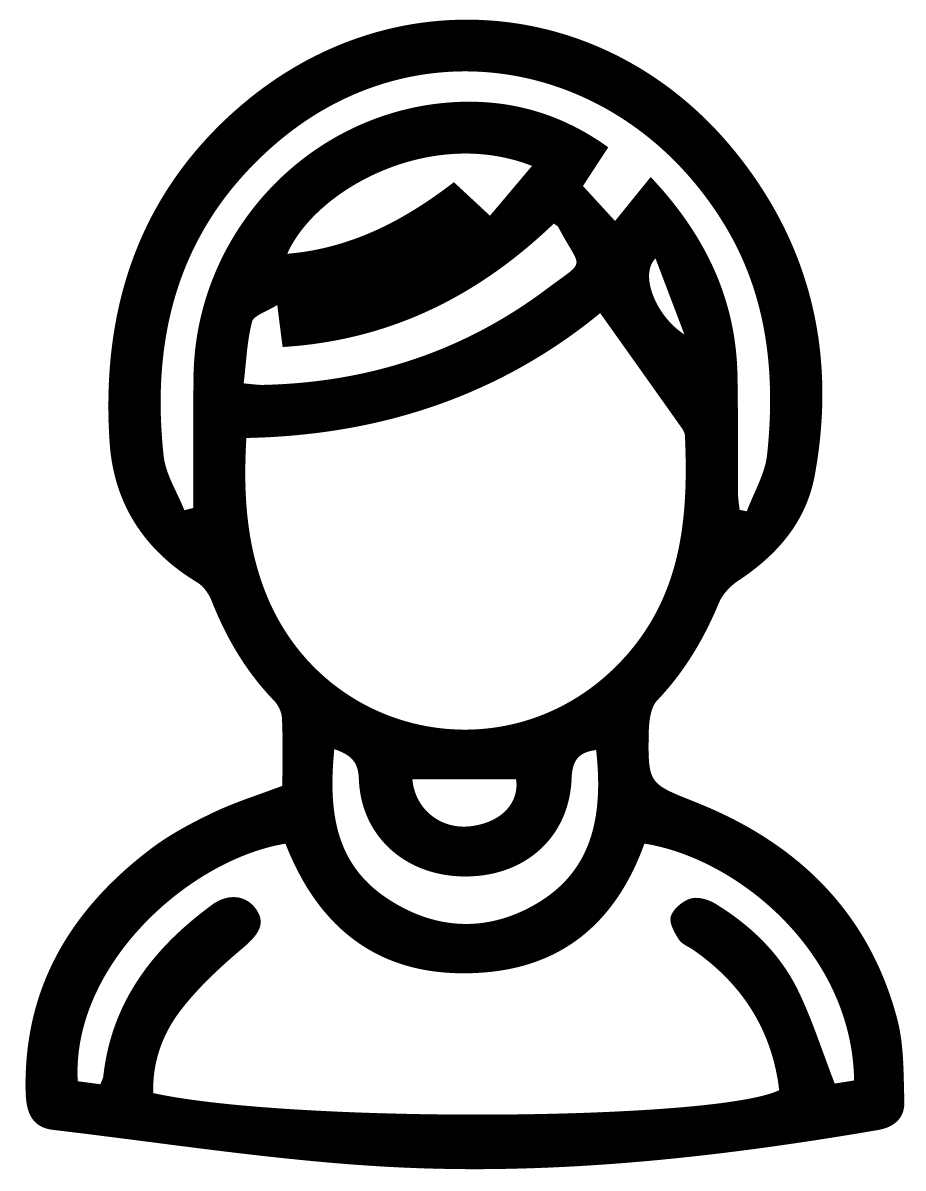} : \textcolor{mycolor}{You give two different answers in previous responses. Check the problem and your answers again, and give the best answer. Your final answer should be put between two \#\# symbols, like \#\# ANSWER \#\#, at the end of your response.} \\
	\includegraphics[width=0.9em]{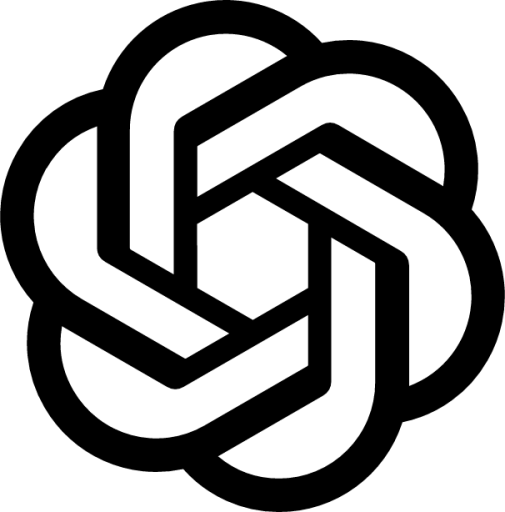} :  \textcolor{second}{Apologies for the confusion. Let's recalculate the profit. The increased value of the house after the repairs is \$80,000 + \$50,000 = \$130,000. The profit is the increased value of the house minus the total cost, so \$130,000 - \$130,000 = \$0. So, the correct profit is \$0. Therefore, the profit he made is \#\# \textbf{0} \#\#.} \textcolor{wrongcolor}{\textbf{(Incorrect)}}
\end{tcolorbox}

\caption{Another example on GSM8K \cite{cobbe2021training} with \texttt{gpt-3.5-turbo-1106} model, where our IoE-based prompt failed to achieve self-correction.}
\label{fig:failed_prompts_2}
\end{figure}

In this section, we present some examples where our IoE-based Prompt failed to self-correct the reasoning. Specifically, we have the following examples:

\begin{itemize}
    \item Figure \ref{fig:failed_prompts}: an example on GSM8K \cite{cobbe2021training}, where the model is confident.
    
    \item Figure \ref{fig:failed_prompts_2}: another example on GSM8K \cite{cobbe2021training}, where the model is uncondident.
\end{itemize}

\noindent\textbf{Failure Analysis.} \ We would like to analyze why our prompting method failed to achieve self-correction.

In this Figure \ref{fig:failed_prompts}, the initial answer is incorrect, when applying our IoE prompt, the answer is still incorrect and the model is confident in its incorrect answer. The reason for this failure can be that the model is not capable enough to provide a correct answer given the current knowledge. 

In this Figure \ref{fig:failed_prompts_2}, the initial answer is correct, however, after applying our IoE-based Prompt, the model becomes unconfident and the answer is changed to an incorrect one. To dive into the initial question (``Josh decides to try flipping a house.  He buys a house for \$80,000 and then puts in \$50,000 in repairs.  This increased the value of the house by 150\%.  How much profit did he make?''), we find that: this question can be understood from multiple aspects, and some can be easily misunderstood. For example, according to "This increased the value of the house by 150\%", we can see that the value of the house has become \$80,000*(1+150\%) = \$200,000. However, it is very likely that one may misthink that ``This increased the value of the house to its 150\%'', therefore, the value of the house is \$80,000*150\% = \$120,000, which is wrong.


\end{document}